\documentclass[]{memtensor}

\usepackage{amsfonts}
\usepackage{amsmath}

\usepackage[dvipsnames]{xcolor}
\usepackage{tikz}
\usepackage{float}
\usepackage{fontawesome7}
\usetikzlibrary{
    shapes, 
    shapes.multipart, 
    arrows.meta, 
    positioning, 
    fit, 
    backgrounds, 
    calc,
    decorations.pathreplacing,
    shadings
}

\def\ourbench{VTCBench}
\def\ours{VTC}
\def\compratio{r_\texttt{\ours}}

\def\contains{\texttt{containsAll}}

\newcommand{\approptoinn}[2]{\mathrel{\vcenter{
  \offinterlineskip\halign{\hfil$##$\cr
    #1\propto\cr\noalign{\kern2pt}#1\sim\cr\noalign{\kern-2pt}}}}}
\newcommand{\appropto}{\mathpalette\approptoinn\relax}
\newcommand{\oom}{{\color{red}\texttt{OOM}}}
\newcommand{\render}[0]{{\mathfrak{R}}}
\newcommand{\renderdefault}[0]{{\mathfrak{R}_\texttt{plain}}}
\newcommand{\renderlocomo}[0]{{\mathfrak{R}_\texttt{conversation}}}

\usepackage{lmodern}
\usepackage{multicol}

\usepackage{float}
\usepackage{makecell}
\usepackage{enumitem}     
\usepackage{wrapfig}

\title{VTCBench: Can Vision-Language Models Understand Long Context with Vision-Text Compression?}

\author{Hongbo Zhao$^{*,1,2}$}
\author{Meng Wang$^{*,3}$}
\author{Fei Zhu$^{3\text{ \faIcon[regular]{envelope}}}$}
\author{Wenzhuo Liu$^{1,2}$}
\author{Bolin Ni$^{4}$}
\author{Fanhu Zeng$^{4}$}
\author{Gaofeng Meng$^{1,2,3}$}
\author{Zhaoxiang Zhang${^{1,2\text{ \faIcon[regular]{envelope}}}}$}

\affiliation{$^1$Institute of Automation, Chinese Academy of Sciences}
\affiliation{$^2$School of Artificial Intelligence, University of Chinese Academy of Sciences}
\affiliation{$^3$Centre for Artificial Intelligence and Robotics, Hong Kong Institute of Science \& Innovation, CAS}
\affiliation{$^4$Independent Researcher}

\contribution[*]{Equal Contributor}
\contribution[\text{\faIcon[regular]{envelope}}]{Corresponding Author}
\checkdata[\faIcon{github} Github]{\url{https://github.com/Moenupa/VTCBench}}
\checkdata[\faIcon{database} Huggingface]{\url{https://huggingface.co/datasets/MLLM-CL/VTCBench}}

\abstract{
The computational and memory overheads associated with expanding the context window of LLMs severely limit their scalability. A noteworthy solution is vision-text compression (VTC), exemplified by frameworks like DeepSeek-OCR~\cite{wei2025deepseek} and Glyph~\cite{cheng2025glyph}, which convert long texts into dense 2D visual representations, thereby achieving token compression ratios of 3x–20x. However, the impact of this high information density on the core long-context capabilities of vision-language models (VLMs) remains under-investigated.
To address this gap, we introduce the first benchmark for VTC and systematically assess the performance of VLMs across three long-context understanding settings: VTC-Retrieval, which evaluates the model's ability to retrieve and aggregate information; VTC-Reasoning, which requires models to infer latent associations to locate facts with minimal lexical overlap; and VTC-Memory, which measures comprehensive question answering within long-term dialogue memory. Furthermore, we establish the VTCBench-Wild to simulate diverse input scenarios.
We comprehensively evaluate leading open-source and proprietary models on our benchmarks. 
The results indicate that, despite being able to decode textual information (e.g., OCR) well, most VLMs exhibit a surprisingly poor long-context understanding ability with VTC-processed information, failing to capture long associations or dependencies in the context.
This study provides a deep understanding of VTC and serves as a foundation for designing more efficient and scalable VLMs.\\}

\begin{document}

\maketitle
\section{Introduction}
\label{sec:intro}

\begin{figure}[t]
    \centering
    \begin{subfigure}{.47\linewidth}
        \small

\tikzset{
    fatarrow/.pic={
        \def\startw{1.2}   
        \def\endw{.6}     
        \def\length{1.4}     
        \def\headlen{.65}    
        \def\headw{1}    
    
        \shade[left color=orange!10, right color=red!20] 
        (0, \startw) 
        .. controls (\length/2, \startw) and (\length*0.7, \endw) .. (\length, \endw)
        -- (\length, \headw) 
        -- (\length+\headlen, 0) 
        -- (\length, -\headw) 
        -- (\length, -\endw)
        .. controls (\length*0.7, -\endw) and (\length/2, -\startw) .. (0, -\startw)
        -- cycle;
    }
}


\begin{tikzpicture}[
    node distance=.3in and .5in,
    >=Stealth, 
    font=\small\sffamily,
    every node/.style={align=center, inner sep=6pt},
    block/.style={draw, minimum width=.5in, minimum height=.3in},
    context/.style={font=\tiny\sffamily, align=left},
    data/.style={rounded corners, inner sep=4pt},
    sm/.style={font=\footnotesize\sffamily},
    xm/.style={font=\scriptsize\sffamily},
]

    \node[
        context, block, data,
        rectangle split,
        rectangle split horizontal,
        rectangle split parts=2,
        rectangle split draw splits=false,
        inner sep=2pt, 
        fill=cyan!10
    ] (context) {
        \parbox[t]{12pt}{
            {\hfill\large\faIcon[solid]{clipboard-question}}
        }
        \nodepart{second}
        \parbox[t]{2.6in}{``question'': ``What are all the special magic numbers for absorbed-cloakroom and \\\hspace{46pt} threatening-blessing mentioned in the provided text?'' \\``needle1'': ``One of the special magic numbers for absorbed-cloakroom is: 7456447.'',\\``needle2'': ``One of the special magic numbers for threatening-blessing is: 6921626.''}
    };
    \node[block, sm, below=of context, xshift=.35in] (VLM) {\faIcon{barcode} Tokens};
    \node[block, below=.2in of VLM] (VLM2) {\faIcon{eye} VLM};

    \node[block, sm, fill=green!10, data, right=.2in of VLM] (img) {\faIcon{image} Images};
    \node[block, sm, left=.8in of VLM] (LLM) {\faIcon{barcode} Tokens};
    \node[block, below=.2in of LLM] (LLM2) {\faIcon{robot} LLM};
    
    \pic at (-1.8,-2.4) [rotate=0, scale=.8] {fatarrow};
    \node[] at (-1.1, -2.4) {$\approx2\text{-}10\times$\\compress};

    \draw[->] (context.south)++(-1.1in,0) -- node[sm, right, yshift=.04in]{tokenize} node[sm, left, yshift=.04in]{$\Phi_\text{text}$} (LLM);
    \draw[->] (context.south)++(+1.17in,0) -- node[sm, left]{render} node[sm, right]{$\render$} (img);
    \draw[->] (img.south) to[bend left] node[sm, below, xshift=.2in, yshift=.05in]{process\\$\Phi_\text{vision}$} (VLM);
    \draw[->] (VLM) -- node[right]{} (VLM2);
    \draw[->] (LLM) -- node[right]{} (LLM2);

    \begin{pgfonlayer}{background}
        \node[
            data,
            fit=(VLM)(VLM2), 
            fill=red!10, 
            inner sep=6pt,
            draw=none,
        ] () {};
    \end{pgfonlayer}
    \begin{pgfonlayer}{background}
        \node[
            data,
            fit=(LLM)(LLM2), 
            fill=orange!10, 
            inner sep=6pt,
            draw=none,
        ] () {};
    \end{pgfonlayer}
\end{tikzpicture}

        \caption{}
        \label{fig:1a}
    \end{subfigure}
    \begin{subfigure}{.52\linewidth}
        \small

\pgfdeclareradialshading{wildshading}{\pgfpoint{-2in}{-2in}}{
    color(2in)=(cyan!30);
    color(2in)=(teal!60)
}

\begin{tikzpicture}[
    node distance=.1in and .3in,
    >=Stealth, 
    font=\small\sffamily,
    every node/.style={align=center, inner sep=6pt},
    block/.style={draw, minimum width=.7in, minimum height=.3in},
    icon/.style={minimum width=.3in},
    context/.style={font=\tiny\sffamily, align=left},
    data/.style={rounded corners, inner sep=4pt},
    thickline/.style={line width=1pt},
]

    \node[block, data, fill=TealBlue!20] (in2) {Reasoning};
    \node[block, data, fill=cyan!10, above=of in2] (in1) {Retrieval};
    \node[block, data, fill=teal!30, below=of in2] (in3) {Memory};
    \node[block, data, below=of in3, 
        left color=TealBlue!30,
        right color=teal!30,
    ] (in4) {Wild};
    
    \node[block, fill=red!10, right=.6in of in2, yshift=-.2in] (vlm) {\faIcon{eye} VLM};

    \node[block, data, icon, thickline, draw=TealBlue!50, right=of vlm, yshift=.2in] (out2) {\large\faIcon[regular]{face-frown}};
    \node[block, data, icon, thickline, draw=cyan!30, above=of out2] (out1) {\large\faIcon[regular]{face-grin}};
    \node[block, data, icon, thickline, draw=teal!60, below=of out2] (out3) {\large\faIcon[regular]{face-dizzy}};
    \node[block, data, icon, thickline, draw=teal!80, below=of out3] (out4) {\large\faIcon[regular]{face-dizzy}};

    \draw[->, thickline, cyan!30]      (in1.east)   to[out=0,   in=160]    ($(vlm.west)+(0,0.15)$);
    \draw[->, thickline, TealBlue!50]  (in2.east)   to[out=0,   in=170]    ($(vlm.west)+(0,0.05)$);
    \draw[->, thickline, teal!60]      (in3.east)   to[out=0,   in=-170]   ($(vlm.west)+(0,-0.05)$);
    \draw[->, thickline, teal!80]      (in4.east)   to[out=0,   in=-160]   ($(vlm.west)+(0,-0.15)$);

    \draw[->, thickline, cyan!30]      ($(vlm.east)+(0,0.15)$)     to[out=20,   in=-160] (out1);
    \draw[->, thickline, TealBlue!50]  ($(vlm.east)+(0,0.05)$)     to[out=10,   in=-170] (out2);
    \draw[->, thickline, teal!60]      ($(vlm.east)+(0,-0.05)$)    to[out=-10,   in=170] (out3);
    \draw[->, thickline, teal!80]      ($(vlm.east)+(0,-0.15)$)    to[out=-20,   in=160] (out4);

    \begin{pgfonlayer}{background}
        \node[
            data,
            fit=(in1)(in2)(in3)(in4), 
            fill=black!10, 
            label=left:{\rotatebox{90}{Long Contexts}}, 
            inner sep=6pt,
            draw=none,
        ] () {};
    \end{pgfonlayer}
    \node[
        data,
        fill=red!30, 
        fill opacity=0.8,
        text opacity=1,
        left=.1in of vlm,
        minimum height=1.2in,
        minimum width=.3in,
    ] () {\rotatebox{90}{VTC}};
    \begin{pgfonlayer}{background}
        \node[
            data,
            fit=(out1)(out2)(out3)(out4), 
            fill=violet!10, 
            label=right:{\rotatebox{90}{Performance Evaluation}}, 
            inner sep=6pt,
            draw=none,
        ] () {};
    \end{pgfonlayer}

\end{tikzpicture}

        \caption{}
        \label{fig:1b}
    \end{subfigure}
    \caption{(a) Vision-text compression offers a more efficient alternative for long-context tasks: instead of feeding plain text directly to LLMs, it renders text as compact images for a VLM, achieving substantial input-token compression. (b) We propose VTCBench and VTCBench-Wild to comprehensively evaluate the long context understanding ability of VLMs within vision-text compression framework.}
    \label{fig:vtc}

\end{figure}
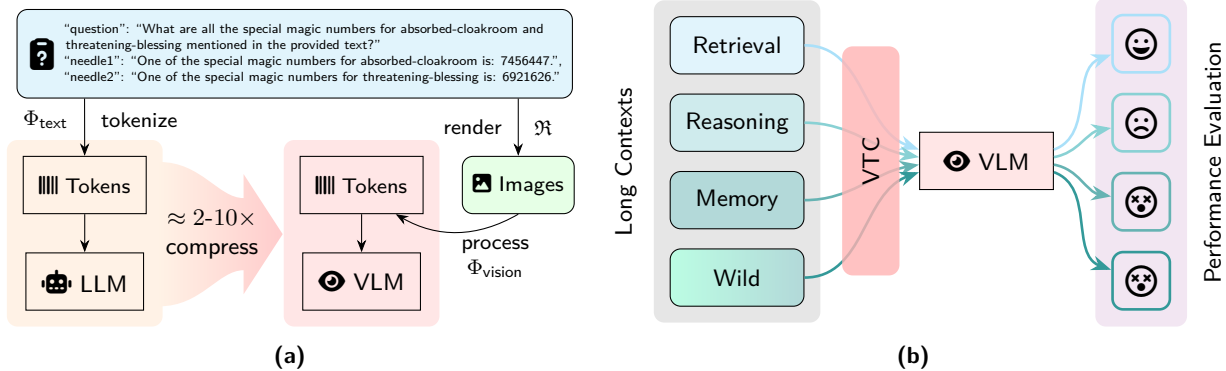

The dramatic success of large language models (LLMs) has been accompanied by a persistent challenge: the scalability of their context window. As the context length expands, the computational and memory costs increase rapidly, severely limiting practical deployment and training efficiency. Existing techniques, including efficient attention \cite{dai2019transformer, yanggated, chen2025minimax}, position encoding extrapolation \cite{su2024roformer, press2021train, xiong2024effective, ding2024longrope}, prompt compression \cite{zhang2024adacomp, gecontext, yoon2024compact} and external memory \cite{zhong2024memorybank, wang2023augmenting, wang2024memoryllm}, offer partial solutions. However, these methods often suffer from notable performance degradation when extrapolated to much longer contexts. Recently, a new paradigm, vision-text compression (VTC), also known as context optical compression, has emerged to address this bottleneck, as illustrated in \Cref{fig:1a}. Recent works like DeepSeek-OCR \cite{wei2025deepseek} and Glyph \cite{cheng2025glyph} propose rendering long text documents into compressed two-dimensional image sequences, thereby leveraging the high information density of the visual modality. This cross-modal approach achieves significantly higher compression ratios, ranging from $3\times$ to $20\times$ fewer input tokens compared to the original text sequence, offering a novel avenue for efficient long-context modeling. By transforming the input space from a token sequence to a pixel grid, VTC shifts the burden of information management from sequential attention to spatial and visual reasoning.

As a novel long context modeling paradigm, VTC \cite{wei2025deepseek, cheng2025glyph} has been demonstrated to not only offer substantial input-token compression but also yield strong performance for OCR tasks (e.g., document parsing \cite{ouyang2025omnidocbench}, multilingual recognition) or long context processing \cite{bai2024longbench, hsieh2024ruler}.
However, a key argument is that such benchmarks, either explicitly or implicitly, rely on identifying exact textual correspondences between the query and the provided context for success~\cite{modarressi2025nolima}. Therefore, a natural and important question arises: 
\begin{center}
\itshape Can vision-language models really understand long context with vision-text compression?
\end{center}
Intuitively, the shift from a sparse, sequential text input relying on linear positional encodings to a dense, two-dimensional visual input fundamentally changes how information is encoded and processed by the model's attention mechanism. Therefore, the reliance of VLMs on spatial locality in the 2D input might lead to unique sensitivities in long-context understanding.

In this work, we present the first systematic benchmark, \textbf{\textit{VTCBench}} (\Cref{fig:1b}), specifically designed for the VTC framework, rigorously quantifying the long-content comprehension capabilities of VLM models. Precisely, we focus on three critical tasks: \textbf{(1) \textit{VTC-Retrieval}} requires models to retrieve, trace, and aggregate information (needles) placed at varying distances within a random text (haystack). \textbf{(2) \textit{VTC-Reasoning}} tests the VLM's capacity of associative reasoning \cite{modarressi2025nolima} over a long context within the compressed visual space. The queried input has minimal literal matches with the context, thus assessing the model's ability to reason associatively beyond mere lexical retrieval.
\textbf{(3) \textit{VTC-Memory}} assesses the VLM’s performance in very long-term dialogue memory \cite{maharana2024locomo, wu2024longmemeval,chhikara2025mem0,zhong2024memorybank}, evaluating the model's resilience to temporal and structural degradation under vision-text compression. In addition, to fully capture the robustness against real-world visual variations, we introduce a unified and lightweight variant, \textbf{\textit{VTCBench-Wild}}.
In summary, these VTC-oriented long-context comprehension benchmarks establish a foundation to understand vision-text compression, offering crucial guidance for future design of highly efficient and capable next-generation long-context VLMs. 

We conduct a comprehensive evaluation of leading open-source and proprietary VLMs on VTCBench and VTCBench-Wild, yielding the following findings:
\begin{itemize}
    \item Overall, existing VLMs exhibit weaker long-context comprehension abilities under the VTC framework compared to dedicated text-only LLMs. This indicates that existing VLM architectures have significant room for improvement when processing information compressed via VTC.
    \item In simple retrieval and matching-related tasks (e.g., simple Needle-in-a-Haystack), VLMs demonstrate a relatively strong performance. This suggests that the VTC approach, which compresses long text into an image representation, minimally impacts the model's fundamental ability to perceive textual content.
    \item In complex long-context understanding tasks, such as associative reasoning and long-term dialogue memory, the performance of most VLMs utilizing VTC is notably poor, falling significantly below that of LLMs (e.g., Qwen3-8B~\cite{yang2025qwen3}). This demonstrates that while VTC maintains satisfactory OCR performance, it seriously impedes core capabilities such as advanced associative reasoning and knowledge integration within long contexts.
    \item Error analysis revealed that VLM failures under VTC range from visual-to-textual grounding to complex deficiencies in associative reasoning. Further experiments demonstrate that the performance is critically dependent on both the font size and the spatial layout of information within the compressed image. 
    \item Notably, Gemini-3-Pro~\cite{gemini3} significantly outperforms some LLMs like Qwen3-8B~\cite{yang2025qwen3} on VTCBench-Wild and nearly reaches parity with its own text-only baseline. This exceptional performance demonstrates that VTC is a highly viable and promising technical route for long-context understanding.
\end{itemize}

\begin{figure*}[t]
    \centering
    \includegraphics[width=\linewidth]{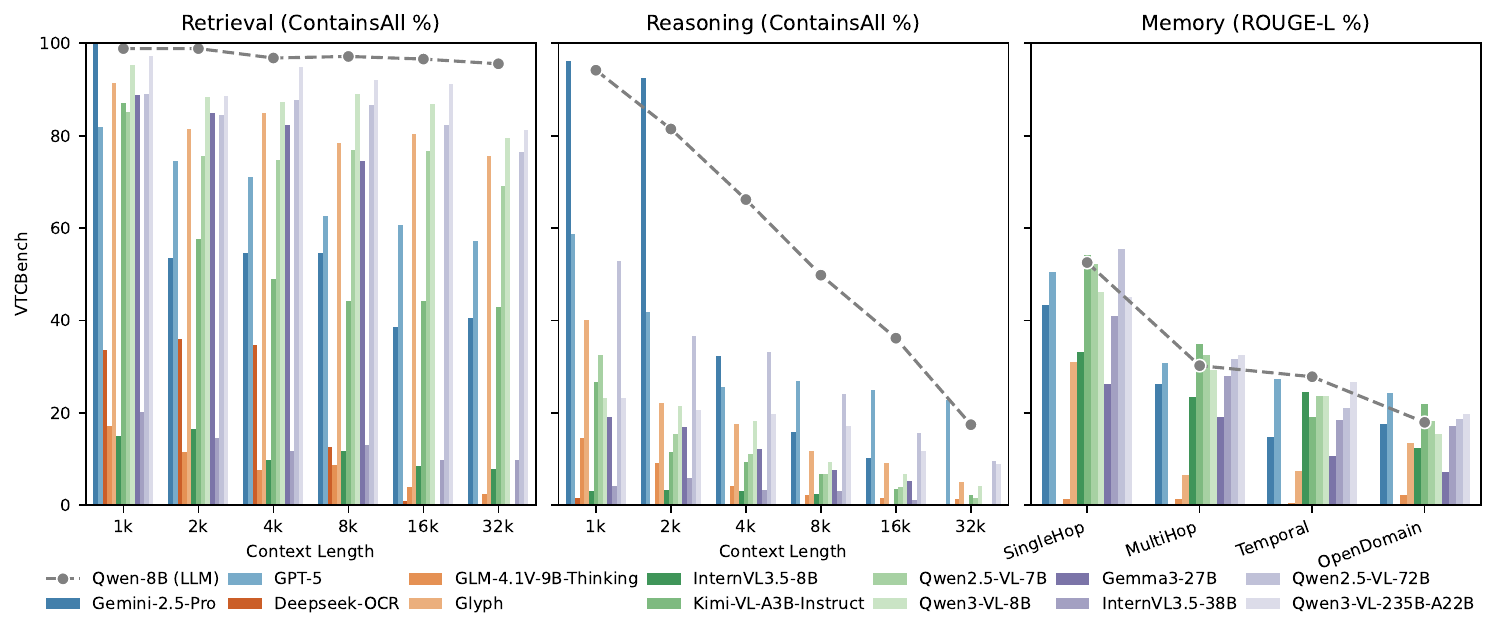}
    \vskip -0.1in
    \includegraphics[width=\linewidth]{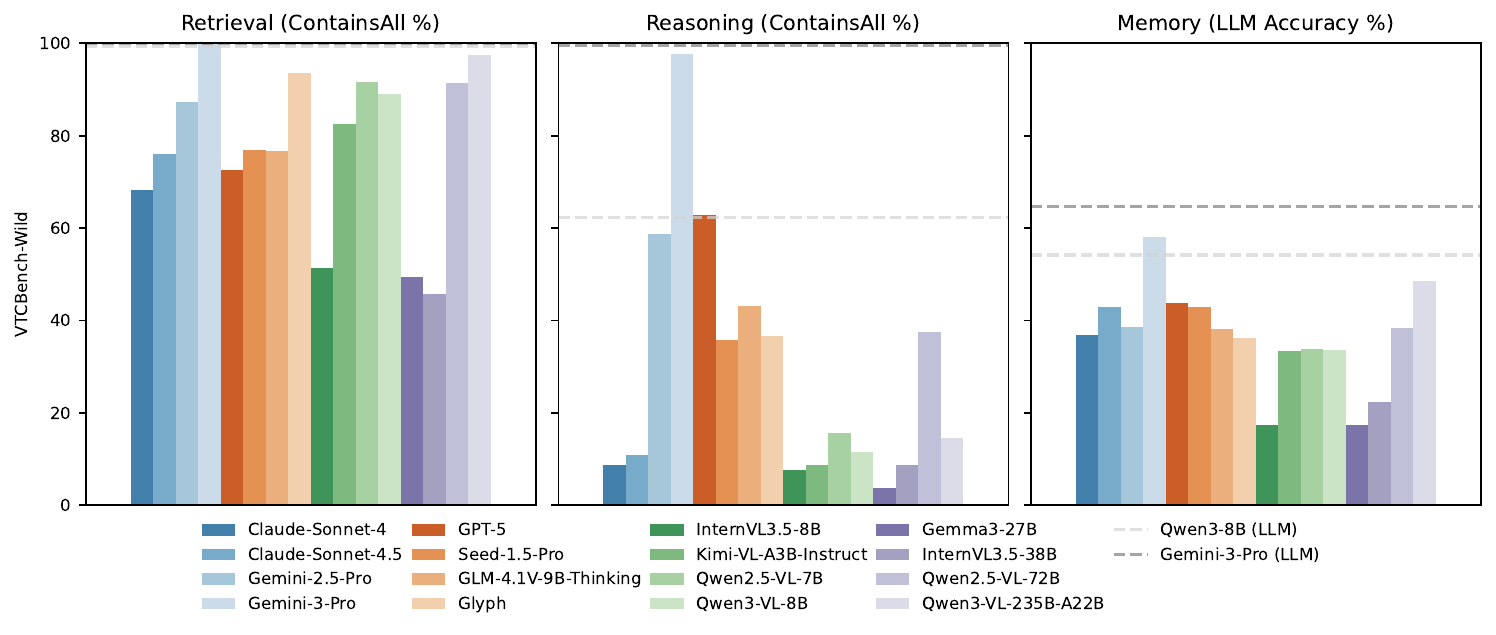}
    \vskip -0.1in
    \caption{Long context understanding performance on VTCBench (first row) and VTCBench-Wild (second row). Under the VTC paradigm, existing VLMs show good textual perception ability, leading to relatively strong performance in simple retrieval tasks. However, they still exhibit weaker long-content comprehension (associative reasoning and long-term dialogue memory) compared to LLMs, highlighting a substantial opportunity to enhance VLMs---especially when processing vision-text compression-based information.}
    \label{fig:routerexam}
\end{figure*}
\section{Related Work}
\label{sec:related}
\subsection{Long-Context Modeling}
The challenge of efficiently processing extensive textual contexts remains a persistent pursuit for large models, driven by the inherent complexity of the standard Transformer self-attention mechanism \cite{liu2025comprehensive}. Recent advancements aim at developing long context language models that can be broadly categorized into architectural modifications and external component integration for workflow-based augmentation. Architectural modifications improve model efficiency internally through sparse and hierarchical attention \cite{beltagy2020longformer, fu2024moa, xiao2024duoattention, wu2021hi}, and recurrent transformer \cite{munkhdalai2024leave, hutchins2022block}. Additionally, positional encoding extrapolation methods \cite{su2024roformer, press2021train, xiong2024effective, ding2024longrope} adjust position encodings to handle larger sequence lengths. A complementary line of research utilizes external components to intelligently manage and compress context through workflow-based augmentation. Prompt compression \cite{zhang2024adacomp, gecontext, yoon2024compact} reduces the input size through selective token retention or text summarization, thereby optimizing the utility of a fixed context window. Retrieval-augmented generation \cite{chen2024bge, gunther2024late} shifts the paradigm by replacing long-context ingestion with dynamic information extraction or recall. Memory-based methods~\cite{zhong2024memorybank, wang2023augmenting, wang2024memoryllm} utilize external memory modules to store and retrieve long-term conversational or document history, ensuring contextual coherence over long interactions without burdening the active context window. Rather than generating a human-readable memory, C3 \cite{liu2025context} introduces a pure text-to-latent compression pipeline using two cascaded LLMs to achieve superior text compression ratios.
Besides, there are also some training strategies~\cite{chen2023longlora, hu2025longrecipe,bai2024longalign} to enable the model to handle long contexts.

\subsection{Vision-Text Compression}
VTC presents a cross-modal paradigm that addresses the long-context challenge by leveraging visual encoding to increase information density. Several recent studies have pioneered this framework. For instance, VIST~\cite{xing2025vision} compresses distant, low-salience context by rendering it into images; these are subsequently processed by a lightweight vision encoder to produce semantically dense visual tokens, achieving a $2.3\times$ token reduction while maintaining state-of-the-art accuracy on in-context learning benchmarks. Similarly, VisInContext~\cite{wang2024leveraging} and Li et al.,~\cite{li2025text} transform long textual sequences into visual representations, significantly lowering computational overhead while enhancing performance in document-understanding tasks.
Further advancing this paradigm, DeepSeek-OCR~\cite{wei2025deepseek} demonstrates the feasibility of high-ratio compression via optical 2D mapping, achieving over $96\%$ OCR decoding precision at a $9\text{--}10\times$ compression ratio. Concurrently, Glyph~\cite{cheng2025glyph} renders long text contexts into compact images and processes them with VLMs. The system successfully demonstrated stable $3\text{--}4\times$ token compression for long text sequences, while preserving performance on OCR and other long-context processing tasks \cite{bai2024longbench, hsieh2024ruler}. These works undertake a preliminary exploration of utilizing the visual modality as an efficient compression medium to facilitate long textual information processing.

\subsection{Long-Context Comprehension Benchmarks}
The Needle-In-A-Haystack (NIAH) test is a classic method for evaluating the long-context understanding capability of LLMs. The vanilla NIAH benchmark \cite{needle_github} evaluates simple retrieval robustness by embedding a specific fact (needle) within a long, irrelevant context (haystack). While useful for measuring basic context window functionality, it falls short of testing context understanding ability. Later, the NIAH task was enhanced by introducing multiple targets, adding distractor noise, and linking facts to enforce complex, or associative reasoning \cite{hsieh2024ruler, kuratov2024search, vodrahalli2024michelangelo}. For example, NoLiMa \cite{vodrahalli2024michelangelo} requires models to perform indirect inference, where the query and the required context do not share direct lexical overlap, forcing the model to rely on deeper semantic understanding and association maintenance across long distances. Beyond text-only benchmarks, NIAH in multimodal tasks has also been explored recently \cite{wang2024needle, wang2025multimodal}. Several benchmarks~\cite{maharana2024locomo, wu2024longmemeval, chhikara2025mem0, zhong2024memorybank} have been designed to evaluate long-term dialogue memory and understanding over very extended conversations. 

\section{\ourbench~}

VTC introduces a new paradigm that uses visual modality as an efficient compression medium for textual information processing in LLMs. Formally, given a long text $\mathcal{T}=(t_1, t_2,\cdots,t_{N_T})$, where $t_i$ is the $i$-th text token, we use a rendering operator $\mathfrak{R}$ to transfer text into $m$ images $\mathcal{I}=\mathfrak{R}(\mathcal{T)}=(I_1, I_2, \cdots, I_m)$.
A VLM $\Phi_{\texttt{VLM}}$ encodes these $m$ images into $N_I$ visual tokens, i.e., $\Phi_{\texttt{VLM}}(\mathcal{I})=(v_1, v_2, \cdots, v_{N_I})$, where $v_i$ denotes the $i$-th visual token fed to the LLM component of the VLMs.
We define the VTC ratio as 
\begin{equation}
    \compratio = \frac{N_T}{N_I}.
    \label{eq:ratio}
\end{equation}

Given that different VLMs adopt different image-processing strategies~\cite{wei2025deepseek,cheng2025glyph,bai2025qwen2,wang2025internvl3}, the visual tokens of the same image differ.
Therefore, \ourbench~comprises two settings (predefined VTC ratio and predefined VTC rendering), where each setting contains three tasks across different long context understanding abilities: \textit{VTC-Retrieval, VTC-Reasoning,} and \textit{VTC-Memory}.
\Cref{fig:appdx:example:locomo,fig:appdx:example:nolima,fig:appdx:example:ruler} show examples of VTCBench.

\subsection{VTC Tasks}\label{VTC:task}
\paragraph{VTC-Retrieval} 
NIAH~\cite{hsieh2024ruler,needle_github,vodrahalli2024michelangelo} test is a standard evaluation in natural language processing that measures a model's ability to understand long contexts.
These benchmarks are often characterized by retrieval-based tasks, where the query's solution has a direct, verbatim correspondence within the source context.
VTC-Retrieval is a visual NIAH task that provides dense-text images as context.
Following RULER~\cite{hsieh2024ruler}, we design 4 sub-tasks, with different settings of ``needle'' (kay-value pairs) inserted into a ``haystack'' (long distractor texts, i.e., Paul Graham essays \cite{kamradt2023needle}) to formulate context.
\begin{itemize}
    \item \emph{Single NIAH (S-NIAH)}: This setting represents the vanilla NIAH test, wherein the model is tasked with retrieving a single key-value pair (the needle) from the haystack.
    \item \emph{Multi-keys NIAH (MK-NIAH):} Multiple distinct key-value pairs are inserted, but the model must retrieve the value corresponding to exactly one specific key. This sub-task assesses the model's ability to disregard hard distractors and precisely locate the queried information.
    \item \emph{Multi-values NIAH (MV-NIAH):} This sub-task involves inserting multiple needles that share an identical key, requiring the model to retrieve all associated values. It is designed to evaluate the model's capacity for high-recall retrieval without omitting relevant information for a given key.
    \item \emph{Multi-queries NIAH (MQ-NIAH):} In this setting, the model must retrieve multiple, distinct key-value pairs in response to a single request that queries all keys. This task evaluates the model's proficiency in multi-query associative recall, testing its ability to return a complete set of requested items.
\end{itemize}

\vspace{-5pt}\paragraph{VTC-Reasoning}
Literal matches make it much easier for VLMs to locate the relevant information and answer correctly~\cite{modarressi2025nolima}. The model only needs to implicitly ``translate'' the image into text, possibly \textit{without} understanding the meaning of the text. 
Therefore, our VTC-Reasoning adds associative reasoning to minimize literal overlap between questions and their corresponding needles.
Questions and needles encompass keywords whose relationships are defined by \textit{associative links}, such as those informed by real-world information or commonsense facts.

\vspace{-5pt}\paragraph{VTC-Memory}
In VTC-Retrieval and VTC-Reasoning tasks, the needles have no relationship with the haystack~\cite{hong2025context}, which can not comprehensively evaluate the performance of long context understanding.
Hence, we integrate long-term memory task into our benchmark.
Long-term memory~\cite{li2025memos, maharana2024locomo,wu2024longmemeval,chhikara2025mem0,zhong2024memorybank} is the capability to distill, store, and retrieve key information, such as persona details and causally linked events. 
We adapt LoCoMo's~\cite{maharana2024locomo} corpus to formulate VTC-Memory. Each sample consists of a long, multi-turn dialogue as contextual passages, and a question-answer pair for evaluation. \ours-Memory spans over 4 sub-tasks:
\begin{itemize}
\item \emph{Single‑hop}: The answer can be derived from a single utterance or fact.
\item \emph{Multi‑hop}: The answer requires aggregating information from multiple turns.
\item \emph{Temporal}: The answer depends on the ordering or timing of events within the dialogue.
\item \emph{Open‑domain}: The answer draws on world knowledge beyond the immediate conversation.
\end{itemize}

\subsection{VTC settings}
\label{sec:vtc_settings}

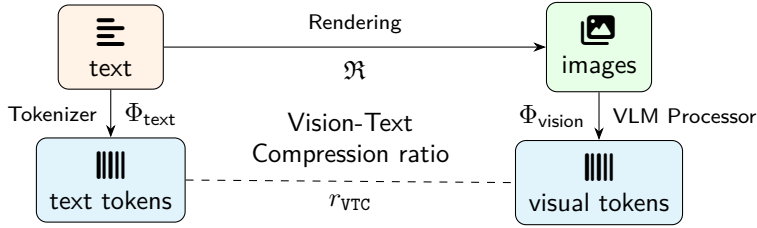
\begin{figure}[!ht]
\centering

\begin{tikzpicture}[
    node distance=.25in and 1in,
    >=Stealth, 
    font=\sffamily,
    every node/.style={align=center, inner sep=5pt},
    block/.style={draw, rounded corners, minimum width=.55in, minimum height=.34in},
    sm/.style={font=\footnotesize\sffamily},
]

\node (text) [block, fill=orange!10] {{\large\faIcon{align-left}}\\text};
\node (ttoken) [block, below=of text, fill=cyan!10] {{\large\faIcon{barcode}}\\text tokens};
\node (img) [block, right=2in of text, fill=green!10] {{\large\faIcon{images}}\\images};
\node (vtoken) [block, below=of img, fill=cyan!10] {{\large\faIcon{barcode}}\\visual tokens};

\draw[->] (text) -- node[sm, left]{Tokenizer} node[right]{$\Phi_\text{text}$} (ttoken);
\draw[->] (text) -- node[sm, above]{Rendering} node[below]{$\render$} (img);
\draw[->] (img) -- node[sm, right]{VLM Processor} node[left]{$\Phi_\text{vision}$} (vtoken);

\draw[dashed] (ttoken.east) -- node[above]{Vision-Text\\Compression ratio}node[below]{$\compratio$} (vtoken.west);

\end{tikzpicture}

\caption{Illustration of the vision-text compression ratio ($\compratio$). The ratio is determined by the interplay between the rendering operator $\mathfrak{R}$ and the VLM's visual encoder $\Phi_{\text{vision}}$ compared to standard text tokenization.}
\label{fig:vtc:pipeline}
\end{figure}

To systematically evaluate how VLMs perform under the VTC paradigm, it is crucial to control the experimental conditions. As illustrated in \Cref{fig:vtc:pipeline}, the final compression ratio, $\compratio$, is a function of both the rendering operator $\render$, which converts text to images, and the VLM's visual encoder $\Phi_\text{vision}$, which encodes images into visual tokens. To disentangle these factors and provide a comprehensive analysis, we design two distinct evaluation settings.

\paragraph{Predefined VTC Ratio}
The primary objective of this setting is to establish a controlled benchmark for a direct comparison of different VLMs' comprehension abilities. To achieve this, we fix the vision-text compression ratio $\compratio$ to a predefined target value (e.g., $\compratio=2$).
Since different VLMs employ different visual encoding strategies (as detailed in \Cref{sec:appdx:compratio:calc}), they generate a varying number of visual tokens $N_I$ for the same input image. Consequently, to maintain a constant $\compratio$ across all models, we must dynamically adjust the rendering operator $\render$ for each specific model.

The most practical parameter to adjust within $\render$ to control the density of the rendered text is the font size. The relationship between font size and $\compratio$ can be approximated as follows, assuming negligible spacing effects. For a given number of words $n_\texttt{word}$  rendered into $n_\texttt{img}$ images of size $(H_\texttt{img}, W_\texttt{img})$:
\begin{align}
\texttt{fontsize}^2 & \propto H_\texttt{ch} \cdot W_\texttt{ch} \label{eq:char} \\
&\appropto 
    \frac{n_\texttt{img}H_\texttt{img}W_\texttt{img}}
    {n_\texttt{char\_per\_word}\cdot n_\texttt{word}} \label{eq:charperimg}
\\
&\appropto 
    \frac{N_I}
    {n_\texttt{char\_per\_word}\cdot N_T}
\\
&= \frac{1}{n_\texttt{char\_per\_word}\cdot \compratio},
\end{align}
where $H_\texttt{ch}$ and $W_\texttt{ch}$ denote the pixel height and width for a character, respectively; spacing such as kerning, tracking, line spacing, indentation, and margins are ignored; the number of characters per word $n_\texttt{char\_per\_word}$ is statistically constant.
Therefore, we approximate the font size of each model by
\begin{equation}
    \texttt{fontsize} \appropto \sqrt{\frac{1}{\compratio}}.
    \label{eq:fontsize}
\end{equation}

By using this relationship, we can set a model-specific font size to ensure a consistent compression ratio. To minimize performance variations caused by model-specific image preprocessing, we standardize the image size to 896 $\times$ 896 pixels, a dimension divisible by common patch sizes (e.g., 14, 16) and tile sizes (e.g., 448). This approach allows us to isolate and assess the intrinsic long-context understanding capability of each VLM at a standardized level of information density.

\vspace{-5pt}\paragraph{Predefined VTC Rendering}
Conversely, this setting simulates a more realistic scenario where the visual representation of the text is standardized, akin to processing a document with fixed formatting. In this setup, we fix the rendering operator R by defining a consistent preset for all models. As a result, while the input images are identical across all evaluations, the resulting compression ratio $\compratio$ will vary, reflecting the unique efficiency and architecture of each model's vision processor $\Phi_\text{vision}$.

In this setting, we adopt a plain-text rendering preset 
$\mathfrak{R}_\texttt{plain}=\{$
$\texttt{dpi}=96$,
$\texttt{font-family}=\texttt{Helvetica}$,
$\texttt{font-size}=12$,
$\texttt{line-height}=1.2$,
$\texttt{margins}=0$,
$\texttt{img-size}=(896,~896)\}$.
This choice balances four goals: following standard typesetting conventions (print \& digital), optimizing vision‑encoder performance, speeding up evaluation, and improving vision‑text compression---12-pt Helvetica offers readability and moderate glyph density; 96‑dpi resolution matches typical screen rendering and avoids unnecessary up‑sampling; minimal text spacing maximizes content per page, enhancing compression while preserving legibility.
We also preset a rendering $\renderlocomo$ for \ours-Memory that colors one speaker's utterances in a green background and the other in white, inspired by messaging apps that use green backgrounds for clear speaker identification.

\section{Experiments}

\begin{table*}[t]
	\centering
	\scriptsize
	\setlength{\tabcolsep}{1.7pt}
	
	\caption{\ourbench~performance (\%) on retrieval, reasoning, and memory tasks. $\compratio$ denotes vision-text compression ratio; for \ours-Retrieval and \ours-Reasoning, 1k, 2k, \dots, 32k denotes equivalent text-form context length in text tokens; for \ours-Memory, SH, MH, TP, and OD denotes Single-Hop, Multi-Hop, Temporal, and Open-Domain subtasks respectively. All Qwen-VL models are instruct models. GPT-5 uses image dimensions $(768,768)$ in the predefined compression ratio setting to avoid resizing.}
	\label{tab:exp:all:in:one}
	\renewcommand{\arraystretch}{1.25}
	\begin{tabular}{l|c|cccccc|cccccc|cccc}
		\toprule
		  \multicolumn{2}{l}{} & \multicolumn{6}{c}{Retrieval (\contains~$\uparrow$)}   & \multicolumn{6}{c}{Reasoning (\contains~$\uparrow$)}   & \multicolumn{4}{c}{Memory (\texttt{ROUGE-L} $\uparrow$)} \\
		\cmidrule(lr){3-8} \cmidrule(lr){9-14} \cmidrule(lr){15-18}
		\multicolumn{1}{l}{Model} & \multicolumn{1}{l}{$\compratio$} & 1k    & 2k    & 4k    & 8k    & 16k   & \multicolumn{1}{c}{32k} & 1k    & 2k    & 4k                                       & 8k    & 16k   & \multicolumn{1}{c}{32k} & SH    & MH    & TP    & OD    \\
		\midrule
		Qwen3-8B (LLM)~\cite{yang2025qwen3} & NA                               & 98.86 & 98.86 & 96.82 & 97.16 & 96.59 & 95.57                   & 94.18 & 81.45 & 66.18                                    & 49.82 & 36.18 & 17.45 & 52.55 & 30.23 & 27.83 & 17.96 \\

		\midrule
		\multicolumn{2}{c|}{Predefined Compression Ratio} & \multicolumn{16}{c}{$\compratio=2.00\pm0.02$} \\
		\midrule
		Kimi-VL-A3B~\cite{team2025kimi}           & 2.00                             & 87.05 & 57.61 & 48.98 & 44.21 & 44.32 & 42.96                   & 26.73 & 11.45 & 9.27                                     & 6.73  & 3.45  & 2.18  & 54.21 & 34.87 & 19.11 & 22.03 \\
		Gemini-2.5-Pro~\cite{comanici2025gemini}  & 1.98                             & 100.0 & 53.64 & 54.54 & 54.55 & 38.64 & 40.57                  & 96.18 & 92.51 & 32.36                                    & 15.82 & 10.18 & \oom  & 43.28 & 26.18 & 14.69 & 17.53 \\
		Gemma3-27B~\cite{team2025gemma}           & 2.00                             & 88.86 & 84.89 & 82.27 & 74.43 & \oom  & \oom                    & 19.09 & 16.91 & 12.18                                    & 7.64  & 5.27  & \oom  & 26.31 & 19.08 & 10.76 & 7.20  \\
		GLM-4.1V-9B-Thinking~\cite{hong2025glm}   & 2.00                             & 17.16 & 11.48 & 7.73  & 8.64  & 3.87  & 2.50                    & 14.55 & 9.09  & 4.18                                     & 2.18  & 1.64  & 1.27  & 1.37  & 1.45  & 0.52  & 2.22  \\
		Glyph~\cite{cheng2025glyph}               & 2.00                             & 91.48 & 81.37 & 84.89 & 78.53 & 80.46 & 75.68                   & 40.09 & 22.07 & 17.65                                    & 11.76 & 9.12  & 5.05  & 4.10  & 2.43 & 1.69 & 2.16 \\
		GPT-5~\cite{GPT5}                         & 2.00                             & 81.93 & 74.55 & 71.14 & 62.73 & 60.68 & 57.16                   & 58.73 & 41.82 & 25.64                                    & 26.82 & 25.00 & 22.73 & 50.41 & 30.81 & 27.36 & 24.38 \\
		InternVL3.5-8B~\cite{wang2025internvl3}   & 2.00                             & 15.00 & 16.59 & 9.89  & 11.71 & 8.52  & 7.96                    & 3.09  & 3.27  & 3.09                                     & 2.36  & 0.36  & 0.00  & 33.19 & 23.55 & 24.49 & 12.35 \\
		InternVL3.5-38B~\cite{wang2025internvl3}  & 2.00                             & 20.11 & 14.66 & 11.82 & 12.96 & 9.78  & 9.89                    & 4.18  & 5.82  & 3.27                                     & 3.09  & 1.09  & 0.18  & 40.96 & 27.92 & 18.54 & 17.19 \\
		Qwen2.5-VL-7B~\cite{bai2025qwen2}         & 2.00                             & 85.23 & 75.57 & 74.77 & 76.82 & 76.59 & 69.21                   & 32.60 & 15.52 & 11.13                                    & 6.71  & 3.86  & 1.60  & 52.22 & 32.60 & 23.76 & 18.30 \\
		Qwen2.5-VL-72B~\cite{bai2025qwen2}        & 2.00                             & 88.98 & 84.55 & 87.73 & 86.59 & 82.27 & 76.48                   & 52.79 & 36.55 & 33.29                                    & 24.04 & 15.61 & 9.52  & 55.55 & 31.61 & 20.96 & 18.75 \\
		Qwen3-VL-8B~\cite{Qwen3-VL}               & 2.00                             & 95.23 & 88.41 & 87.39 & 88.98 & 86.93 & 79.59                   & 23.27 & 21.45 & 18.18                                    & 9.27  & 6.73  & 4.18  & 46.09 & 29.31 & 23.63 & 15.43 \\
		Qwen3-VL-235B-A22B~\cite{Qwen3-VL}        & 2.00                             & 97.16 & 88.52 & 94.77 & 92.16 & 91.21 & 81.34                   & 23.27 & 20.55 & 19.82                                    & 17.09 & 11.82 & 8.91  & 45.10 &  32.62	& 26.77	& 19.71  \\
		
		\midrule
		\multicolumn{2}{c|}{Predefined Rendering} & \multicolumn{12}{c|}{$\renderdefault$} & \multicolumn{4}{c}{$\renderlocomo$}\\
		\midrule
		Deepseek-OCR~\cite{wei2025deepseek}                              & 3.12                             & 18.18 & 21.82 & 14.66 & 0.34  & 0.00  & \oom                    & 1.69  & 0.38  & 0.06                                     & 0.00  & 0.00  & \oom  & \oom  & \oom  & \oom  & \oom  \\
		Kimi-VL-A3B~\cite{team2025kimi}                               & 2.00                             & 87.05 & 57.61 & 48.98 & 44.21 & 44.32 & 42.96                   & 26.73 & 11.45 & 9.27                                     & 6.73  & 3.45  & 2.18  & 54.21 & 34.87 & 19.11 & 22.03 \\
		Gemini-2.5-Pro~\cite{comanici2025gemini}  & 1.98                             & 100.0 & 53.64 & 54.54 & 54.55 & 38.64 & 40.57                  & 96.18 & 92.51 & 32.36                                    & 15.82 & 10.18 & \oom  & 43.28 & 26.18 & 14.69 & 17.53 \\
		Gemma3-27B~\cite{team2025gemma}           & 8.00                             & 22.05 & 17.96 & 22.50 & 18.52 & 14.43 & 13.86                   & 2.73  & 2.73  & 1.64                                     & 1.82  & 1.64  & 1.64  & 18.12 & 15.34 & 10.56 & 8.45  \\
		GLM-4.1V-9B-Thinking~\cite{hong2025glm}   & 2.00                             & 17.16 & 11.48 & 7.73  & 8.64  & 3.87  & 2.50                    & 14.55 & 9.09  & 4.18                                     & 2.18  & 1.64  & 1.27  & 1.37  & 1.45  & 0.52  & 2.22  \\
		Glyph~\cite{cheng2025glyph}               & 2.00                             & 91.48 & 81.37 & 84.89 & 78.53 & 80.46 & 75.68                   & 40.09 & 22.07 & 17.65                                    & 11.76 & 9.12  & 5.05  & 4.10  & 2.43 & 1.69 & 2.16 \\
		GPT-5~\cite{GPT5}                         & 3.17                             & 45.57 & 38.64 & 32.05 & 32.96 & 29.09 & 25.34                   & 32.51 & 10.47 & 11.22                                    & 10.85 & 11.07 & 8.12  & 38.94 & 27.86 & 23.69 & 26.66 \\
		InternVL3.5-8B~\cite{wang2025internvl3}   & 1.56                             & 23.64 & 21.82 & 23.64 & 20.80 & 16.02 & 12.50                   & 7.45  & 6.00  & 3.64                                     & 2.91  & 3.09  & 0.73  & 22.29 & 18.30 & 21.38 & 15.65 \\
		InternVL3.5-38B~\cite{wang2025internvl3}  & 1.56                             & 21.59 & 21.25 & 21.82 & 21.25 & 17.73 & 18.98                   & 6.36  & 6.00  & 4.91                                     & 5.45  & 3.64  & 2.55  & 24.53 & 19.84 & 18.47 & 17.20 \\
		Qwen2.5-VL-7B~\cite{bai2025qwen2}         & 2.00                             & 85.23 & 75.57 & 74.77 & 76.82 & 76.59 & 69.21                   & 32.60 & 15.52 & 11.13                                    & 6.71  & 3.86  & 1.60  & 52.22 & 32.60 & 23.76 & 18.30 \\
		Qwen2.5-VL-72B~\cite{bai2025qwen2}        & 2.00                             & 88.98 & 84.55 & 87.73 & 86.59 & 82.27 & 76.48                   & 52.79 & 36.55 & 33.29                                    & 24.04 & 15.61 & 9.52  & 55.55 & 31.61 & 20.96 & 18.75 \\
		Qwen3-VL-8B~\cite{Qwen3-VL}               & 2.55                             & 94.21 & 72.39 & 66.48 & 61.48 & 49.43 & 53.30                   & 17.45 & 3.45  & 2.73                                     & 0.91  & 0.91  & 0.91  & 37.03 & 27.83 & 22.37 & 15.40 \\
		Qwen3-VL-235B-A22B~\cite{Qwen3-VL}        & 2.55                             & 97.73 & 86.93 & 85.57 & 79.89 & 63.64 & 58.54                   & 22.55 & 8.00  & 5.27                                     & 3.64  & 2.00  & 2.36  & 45.28  & 30.91  & 27.92 & 19.27  \\
		\bottomrule
	\end{tabular}
\end{table*}

\subsection{Setup}
\paragraph{Models and Inference Setup}
We select 13 VLMs, comprising 11 open-source models (Qwen2.5-VL-7B~\cite{bai2025qwen2}, Qwen2.5-VL-72B~\cite{bai2025qwen2}, Qwen3-VL-8B~\cite{Qwen3-VL}, Qwen3-VL-235B-A22B~\cite{Qwen3-VL}, Deepseek-OCR~\cite{wei2025deepseek}, Kimi-VL-A3B-Instruct~\cite{team2025kimi}, Gemma3-27B~\cite{team2025gemma}, GLM-4.1V-9B-Thinking~\cite{hong2025glm}, InternVL3.5-8B~\cite{wang2025internvl3}, InternVL3.5-38B~\cite{wang2025internvl3}, and Glyph~\cite{cheng2025glyph}) and two closed-source models (Gemini-2.5 Pro~\cite{comanici2025gemini} and GPT-5~\cite{GPT5}). Collectively, these models cover the three main visual encoding categories and varying image-processing strategies discussed in \Cref{sec:appdx:compratio:calc}, covering diverse sizes (3B to 235B) and architectures (dense or MoE).
We also test Qwen3-8B~\cite{yang2025qwen3} as a strong LLM baseline to show the performance gap between VLMs and LLMs. We evaluate all open-source models with vllm~\cite{kwon2023efficient}, using each model's default generation configuration, running in BFloat16 on NVIDIA A100 GPUs.
More experimental details can be found in \Cref{sec:appdx:C}.

\vspace{-7pt}\paragraph{Task configurations}
We test all models on two settings (predefined VTC ratio and predefined rendering) with the VTC-Retrieval, VTC-Reasoning, and VTC-Memory tasks.
For VTC-Retrieval and VTC-Reasoning, we evaluate the models on the corpus from the series (1k, 2k, 4k, 8k, 16k, 32k).
The evaluation is conducted by inserting a single needle at a specific position within the context. 
This process is repeated 11 times, with the needle placed at depths of 0\%, 10\%, 20\%, ..., up to 100\% of the context length.
We found that most VLMs perform poorly when the corpus size is 32k, which is why we do not consider longer texts.

\subsection{Main Results}
We present the main results of our evaluation on VTCBench, analyzing the performance of various VLMs across the three core long-context understanding tasks. The results, detailed in \Cref{tab:exp:all:in:one}, are discussed below.

\paragraph{VTC-Retrieval} 
The results for VTC-Retrieval demonstrate that many leading VLMs possess strong fundamental OCR and literal matching capabilities. In the predefined compression ratio setting ($\compratio=2$), models like Qwen3-VL-235B (97.16\% at 1k), Qwen3-VL-8B (95.23\% at 1k), and Glyph (91.48\% at 1k) achieve high accuracy on shorter contexts. However, a noticeable performance gap remains when compared to the text-only Qwen3-8B baseline, which consistently scores above 95\% even at a 32k context length. Crucially, nearly all VLMs exhibit a clear degradation in performance as the context length increases. For instance, Qwen3-VL-235B’s accuracy drops from 97.16\% at 1k to 81.34\% at 32k, indicating that VTC introduces a degree of perceptual fragility that impairs reliable information extraction from longer, denser visual contexts. 
It should be pointed out that DeepSeek-OCR’s lower performance is attributable to its training procedures, which lack the supervised fine-tuning (SFT) stage common in instruction models (see \Cref{sec:error}). 
In contrast, some models like InternVL3.5 show surprisingly low performance, hinting at architectural mismatches with dense-text images (as analyzed in \Cref{sec:error}). These results confirm that while VLMs can perceive text within the VTC framework, their retrieval reliability is sensitive to context density and length.

\vspace{-7pt}\paragraph{VTC-Reasoning}
The VTC-Reasoning task, which requires associative inference rather than simple lexical matching, reveals a stark performance collapse across all evaluated VLMs. As shown in \Cref{tab:exp:all:in:one}, even models that excelled at VTC-Retrieval struggle significantly here. While the Qwen3-8B LLM baseline also shows degradation, its performance (e.g., 94.18\% at 1k) far surpasses that of most VLMs. For example, in the $\compratio=2$ setting, Qwen2.5-VL-72B achieves a respectable 52.79\% at 1k, but this is an outlier; most other open-source models score below 40\%. This finding highlights a critical limitation: current VLMs can successfully perform surface-level text recognition under the VTC paradigm, but their ability to transform this visual perception into deeper semantic comprehension and establish latent connections is severely limited. Intriguingly, the newer Qwen3-VL series consistently underperforms the older Qwen2.5-VL models on this task. We hypothesize this is due to an over-reliance on literal matches, causing them to refuse to answer associative queries (see \Cref{sec:error}). Furthermore, comparing the two settings in \Cref{tab:exp:all:in:one} confirms that the performance is inversely correlated with the VTC ratio, and higher compression disproportionately harms complex reasoning.

\vspace{-7pt}\paragraph{VTC-Memory}
In contrast to the synthetic nature of the previous tasks, VTC-Memory assesses performance on cohesive dialogues, offering a more practical evaluation of long-context understanding. The results in \Cref{tab:exp:all:in:one} again reveal a significant performance gap between most VLMs and the text-only baseline, particularly on multi-hop (MH), temporal (TP), and open-domain (OD) questions that require information aggregation. However, the Qwen2.5-VL series emerges as a notable exception, achieving strong results that are competitive with, and in some cases even exceed, the LLM baseline (e.g., on the Single-Hop subtask, Qwen2.5-VL-72B scores 55.55\% ROUGE-L versus the LLM’s 52.55\%). Proprietary models like GPT-5 also demonstrate competent performance. This suggests that while high information density generally impairs long-term memory capabilities, it is not an insurmountable challenge, and that certain architectures are already developing resilience. Nevertheless, for the majority of models, the sharp drop in performance on complex memory subtasks underscores the difficulty of integrating and reasoning over interconnected facts within a visually compressed context.

\begin{table*}[ht]
\centering
\caption{Performance of Qwen2.5-VL-7B-Instruct with different rendering operators. Each follow-up experiment alters exactly one rendering parameter from {\color{orange}baseline}. }
\label{tab:exp:ablation}
\begin{subtable}[t]{0.56\textwidth}
\centering
\small
\caption{Performance on \ours-Retrieval and \ours-Reasoning.}
\label{tab:exp:combined_ablation}

\begin{tabular}{lcc}
\toprule
& Retrieval & Reasoning \\
\cmidrule(lr){2-2} \cmidrule(lr){3-3}
Rendering Parameters & S-NIAH 1k & 1k \\
\midrule
$\render_\texttt{plain-16}$ & \color{orange}98.79 & \color{orange}36.73 \\
\midrule
$\cup~\render(\texttt{font-size}=10)$ & 73.64 & 20.00 \\
$\cup~\render(\texttt{font-size}=12)$ & 86.06 & 27.64 \\
$\cup~\render(\texttt{font-size}=14)$ & 99.39 & 31.27 \\
$\cup~\render(\texttt{font-size}=16)$ & \color{orange}98.79 & \color{orange}36.73 \\
$\cup~\render(\texttt{font-size}=18)$ & 100.00 & 39.09 \\
$\cup~\render(\texttt{font-size}=20)$ & 100.00 & 46.36 \\
\midrule
$\cup~\render(\texttt{font}=\texttt{Arial})$ & 99.09 & 35.82 \\
$\cup~\render(\texttt{font}=\texttt{Courier New})$ & 98.79 & 38.18 \\
$\cup~\render(\texttt{font}=\texttt{Times New Roman})$ & 99.09 & 36.91 \\
\midrule
$\cup~\render(\texttt{bg-color}=\texttt{red})$ & 98.79 & 41.82 \\
$\cup~\render(\texttt{bg-color}=\texttt{green})$ & 99.70 & 34.73 \\
$\cup~\render(\texttt{bg-color}=\texttt{blue})$ & 96.67 & 39.27 \\
$\cup~\render(\texttt{bg-color}=\texttt{gray})$ & 99.39 & 34.18 \\
\midrule
$\cup~\render(\texttt{color}=\texttt{red})$ & 99.09 & 35.64 \\
$\cup~\render(\texttt{color}=\texttt{green})$ & 99.70 & 34.91 \\
$\cup~\render(\texttt{color}=\texttt{blue})$ & 99.09 & 37.09 \\
$\cup~\render(\texttt{color}=\texttt{gray})$ & 99.09 & 30.18 \\
\midrule
$\cup~\render(\texttt{line-height}=1.25)$ & 98.79 & 38.36 \\
$\cup~\render(\texttt{line-height}=1.5)$ & 100.00 & 38.91 \\
$\cup~\render(\texttt{line-height}=2)$ & 99.70 & 50.18 \\
\bottomrule
\end{tabular}
\end{subtable}\hfill
\begin{subtable}[t]{0.42\textwidth}
\centering \small
\caption{Performance (\texttt{ROUGE-L}$\uparrow$) on \ours-Memory.}
\label{fig:appdx:rendering:ablation:locomo}

\renewcommand{\arraystretch}{1.068}
\begin{tabular}{lc}
\toprule
Rendering Config & SingleHop \\
\midrule
$\renderlocomo$ & {\color{orange}50.54} \\
\midrule
$\cup~\render(\texttt{font-size}=7                       )$ & 21.81 \\
$\cup~\render(\texttt{font-size}=8                       )$ & 29.29 \\
$\cup~\render(\texttt{font-size}=9                       )$ & 33.81 \\
$\cup~\render(\texttt{font-size}=10                      )$ & 40.93 \\
$\cup~\render(\texttt{font-size}=11                      )$ & 45.54 \\
$\cup~\render(\texttt{font-size}=12                      )$ & {\color{orange}50.54} \\
$\cup~\render(\texttt{font-size}=13                      )$ & 49.94 \\
$\cup~\render(\texttt{font-size}=14                      )$ & 53.06 \\
\midrule
$\cup~\render(\texttt{font}=\texttt{Arial}                )$ & 49.79 \\
$\cup~\render(\texttt{font}=\texttt{Courier New}          )$ & 53.15 \\
$\cup~\render(\texttt{font}=\texttt{Times New Roman}      )$ & 47.05 \\
\midrule
$\cup~\render(\texttt{bg-color}=\texttt{white}            )$ & 49.81 \\
$\cup~\render_\texttt{plain-12} $ & 49.83 \\
\midrule
$\cup~\render(\texttt{bg-color}=\texttt{red}              )$ & 43.49 \\
$\cup~\render(\texttt{bg-color}=\texttt{green}            )$ & 43.68 \\
$\cup~\render(\texttt{bg-color}=\texttt{blue}             )$ & 39.93 \\
$\cup~\render(\texttt{bg-color}=\texttt{gray}             )$ & 48.72 \\
\midrule
$\cup~\render(\texttt{color}=\texttt{red}                 )$ & 49.14 \\
$\cup~\render(\texttt{color}=\texttt{green}               )$ & 48.11 \\
$\cup~\render(\texttt{color}=\texttt{blue}                )$ & 46.49 \\
\bottomrule
\end{tabular}
\end{subtable}
\end{table*}

\subsection{Analysis of Rendering Configuration}\label{sec:rendering}
To comprehensively assess the impact of the rendering operator $\mathfrak{R}$ on VLMs' capabilities beyond simple OCR, we conducted an ablation study on VTC-Retrieval, VTC-Reasoning, and VTC-Memory tasks. All experiments employed the Qwen2.5-VL-7B-Instruct model. We established a baseline using $\mathfrak{R}_{\texttt{plain-16}}$ for VTC-Retrieval and VTC-Reasoning, and $\mathfrak{R}_{\texttt{conversation}}$ for VTC-Memory, and then introduced perturbations to exactly one rendering attribute, selected from font size, font family, text/background color, and line height. 

The results in \Cref{tab:exp:ablation} demonstrate that font size is the primary determinant of model performance across all three tasks. This is likely attributable to increased character legibility, which facilitates more accurate OCR by the model's vision encoder. However, this improvement comes at the cost of a lower compression ratio, as discussed in \Cref{sec:render_pipeline}.
In contrast, the model demonstrates impressive robustness to stylistic changes: variations in font family and text or background colors exert only minor effects on performance across all tasks, provided sufficient contrast is maintained. This indicates that the model's internal representations are well-generalized and do not overfit to specific visual styles.
These findings indicate that the model and potentially other VLMs are generally robust to stylistic rendering changes and that readability, rather than stylistic variation, is the critical determinant in VTC scenarios.

\subsection{Analysis of Needle Position}

\begin{figure}[ht]
    \centering
    \begin{subfigure}{.49\linewidth}
        \centering
        \includegraphics[width=\textwidth]{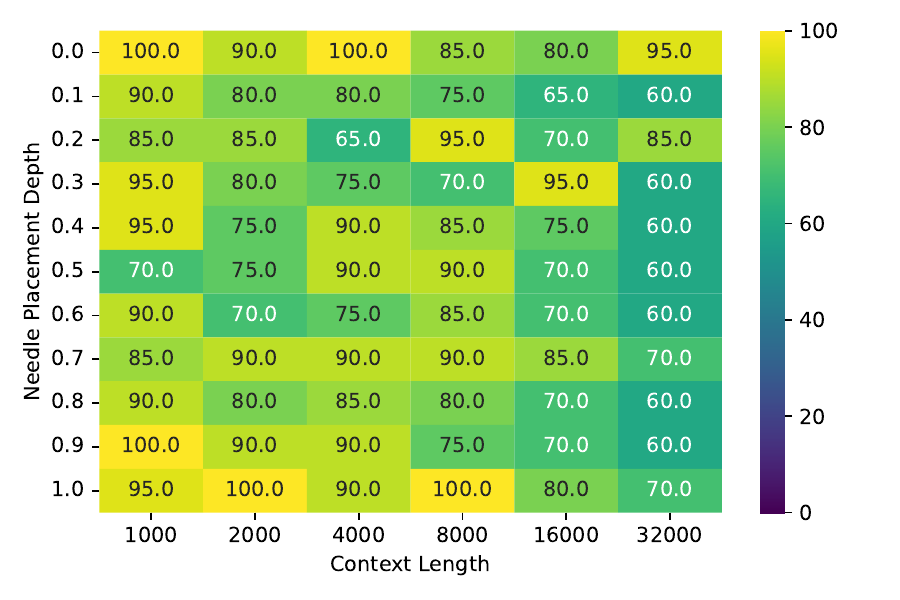}
        \vskip -0.1in
        \caption{Performance on VTC-Retrieval MQ-NIAH.}
    \end{subfigure}
    \begin{subfigure}{.49\linewidth}
        \centering
        \includegraphics[width=\textwidth]{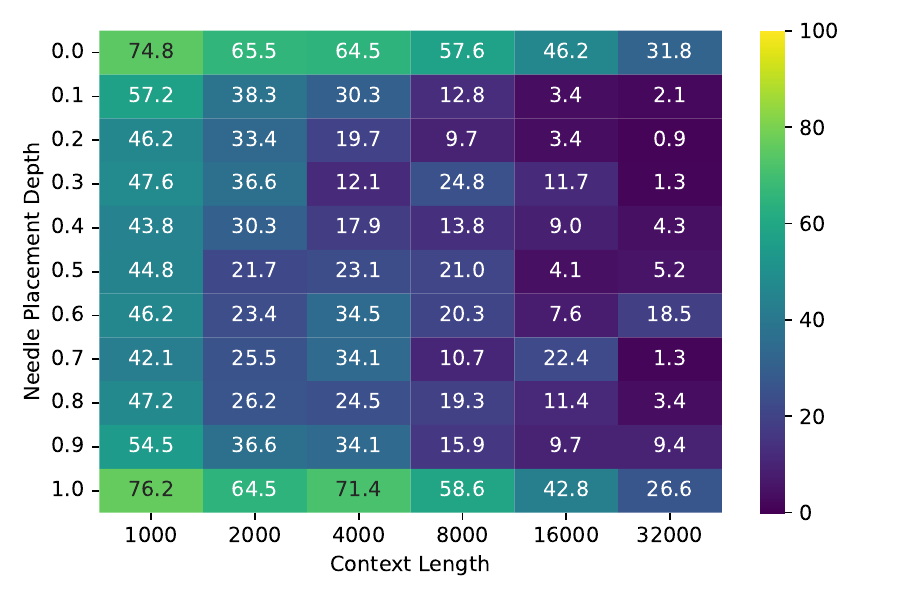}
        \vskip -0.1in
        \caption{Performance on VTC-Reasoning.}
    \end{subfigure}
    \caption{Performance on VTC-Retrieval and VTC-Reasoning with respect to needle placement depth using Qwen2.5-VL-72B-Instruct. The U-shaped performance curve reveals a ``lost in the middle'' phenomenon: accuracy is highest at the context's edges but plummets in the center, an effect that worsens as context length increases.}
    \label{fig:heatmap}
\end{figure}

To further dissect the performance of VLMs under VTC paradigm, we analyze the impact of the needle's position within the visually compressed context. This analysis is crucial for understanding whether models process the entire visual space uniformly or exhibit positional biases. \Cref{fig:heatmap} visualizes the model's accuracy on the VTC-Retrieval and VTC-Reasoning as a function of both context length and the relative depth of the needle within the document (more results can be found in \Cref{fig:appdx:ruler:s,fig:appdx:ruler:mk,fig:appdx:ruler:mv,fig:appdx:ruler:mq}).

The results reveal a pronounced ``lost in the middle'' phenomenon, a challenge well-documented in text-only long-context models~\cite{modarressi2025nolima} but now observed in the VTC spatial domain. Taking the results on VTC-Reasoning task as an example, the performance is consistently highest when the needle is located at the very beginning (0\% depth) or end (100\% depth) of the context, forming a distinct U-shaped curve in accuracy. As shown in the heatmap, accuracy drops precipitously for needles placed in the central portion of the context. For instance, with a 4k context length, accuracy is relatively high at the edges but plummets to as low as 12.1\% in the middle.
This positional bias is severely exacerbated as the context length increases. For a 16k context, the model can still retrieve information from the edges with over 40\% accuracy, but its performance on needles in the middle section collapses to near-zero (e.g., 3.4\%), rendering the central part of the context almost entirely inaccessible for complex reasoning.

This finding has critical implications for the VTC paradigm. It suggests that even when text is successfully rendered and theoretically perceptible, the spatial arrangement within the image creates strong attentional biases. Current VLM architectures appear to struggle with distributing attention evenly across a dense, uniform grid of text. Instead, they prioritize information at the start and end of the visual sequence, analogous to how text-only models handle the beginning and end of a 1D token sequence. Therefore, the ability to reason over long contexts via VTC is not only a function of overall context length and information density but is also critically dependent on the spatial location of the relevant facts. Overcoming this positional fragility is a key challenge for developing truly effective VTC-based models.

\subsection{Error Analysis}\label{sec:error}
To gain deeper insights into the failure modes of VLMs on our benchmark, we performed a qualitative error analysis on the model responses. Our analysis identified several recurring categories of errors, which span from fundamental retrieval issues to more complex reasoning and visual-to-textual grounding failures. These categories are not mutually exclusive; a single incorrect response may exhibit multiple failure modes.

\paragraph{Logical and Associative Reasoning Deficiencies} In the VTC-Reasoning task, many errors were not due to failed retrieval but a breakdown in the subsequent reasoning step. Models would often successfully extract the relevant facts (e.g., ``Katie is a vegan'') but fail to perform the required logical inference (e.g., concluding that Katie cannot eat fish). 
That is why most models perform worse than VTC-Retrieval on the VTC-Reasoning.

\paragraph{Refusal to Conduct Associative Reasoning} 
We observed a distinct failure mode in the Qwen3-VL series models, where the models frequently refuse to answer the prompt, often outputting responses such as ``none of [category]'' or stating that the information is not present.
This behavior appears to stem from a failure to perform the necessary multi-step inference when the visual needle (evidence) and the textual question do not share a literal match.
In the VTC-Reasoning tasks, the correct answer requires linking the visual context to the query via associative logic rather than direct keyword retrieval. However, Qwen3-VL seems to over-rely on lexical correspondence; consequently, when the specific phrasing of the question does not explicitly appear in the compressed visual text, the model conservatively defaults to a refusal, incorrectly assuming the requisite information is absent from the context. We also found this in the InternVL3.5 series model. \Cref{tab:refusal:ratio} shows the refusal ratio of Qwen3-VL-235B-A22B-Instruct. 
We observed that the model refused to answer more than 60\% of the questions, which severely degraded its performance on reasoning tasks. Despite possessing strong literal matching capabilities (VTC-Retrieval), the model underperformed relative to its predecessor, Qwen2.5-VL, under VTC-Reasoning.
We guess this may stem from over-tuned safety alignments that incorrectly flag the associative query as unanswerable.

\begin{table}[t]
    \centering 
    \small 
    \caption{Refusal ratio (\%) of Qwen3-VL-235B-A22B-Instruct (preset compression ratio setting in VTC-Reasoning).}
    \label{tab:refusal:ratio}
    
    \begin{tabular}{lcccccc}
        \toprule
        \multirow{2}{*}{Model} & \multicolumn{6}{c}{\ours-Reasoning (Refusal Ratio)} \\
        \cmidrule(lr){2-7}
         & 1k & 2k & 4k & 8k & 16k & 32k \\
        \midrule
        Qwen3-VL-235B-A22B & 66.18& 62.00 &62.91&60.55&64.55&63.45 \\
        \bottomrule
    \end{tabular}
\end{table}

\paragraph{Missing in the Haystack}
Another prevalent failure mode is what we term ``Missing in the Haystack''.
This error occurs when the model fails to pinpoint the exact needle, instead returning a plausible but incorrect piece of information from the surrounding haystack. 
For instance, in the VTC-Retrieval task, when prompted for ``the special magic number for long-context'' (the needle key, with the correct answer being ``2026''), models would sometimes respond with ``2025'', a distractor value deliberately placed in the context. 
This type of error highlights a failure in fine-grained grounding and retrieval precision. It suggests that even when the model's vision component successfully performs OCR, its attention mechanism can be easily swayed by semantically similar distractors, leading it to retrieve the wrong fact.
This issue becomes more pronounced as the context lengthens and the density of distracting information increases, underscoring the challenge of precise information extraction from visually compressed text.

\begin{table}[ht]
    \centering 
    \small 
    \caption{Performance of Qwen2.5-VL-7B-Instruct on \ours-Retrieval S-NIAH with respect to needle key-value types.}
    \label{tab:appdx:needletype}
    \begin{tabular}{lccc}
    \toprule
        \multirow{2}{*}{Model}& \multicolumn{3}{c}{Retrieval S-NIAH 1k (\contains~$\uparrow$)} \\
        \cmidrule(lr){2-4}
        & word-word & word-number & uuid-number \\
    \midrule
        Qwen2.5-VL-7B & 91.82 & 80.91 & 85.46 \\
    \bottomrule
    \end{tabular}
\end{table}

\paragraph{Sensitivity to Needle Types}
Following~\cite{hsieh2024ruler}, we examine how the model’s retrieval performance varies with different needle key and value types. In the \ours‑Retrieval S‑NIAH sub‑task, we isolate three sub‑conditions that each pair a specific key‑type with a specific value‑type. By keeping all other factors constant, we can attribute performance differences directly to the nature of the needle. The results (\Cref{tab:appdx:needletype}) reveal a clear hierarchy: The results show a systematic decline in retrieval accuracy as the semantic ``richness'' of the needle diminishes. This finding aligns with the observations reported by \citet{hsieh2024ruler} on the brittleness of retrieval‑oriented LLMs to identifier‑like queries.

\begin{figure}[!ht]
    \centering 
    \begin{subfigure}[b]{0.37\linewidth}
        \small
        \includegraphics[width=0.9\linewidth]{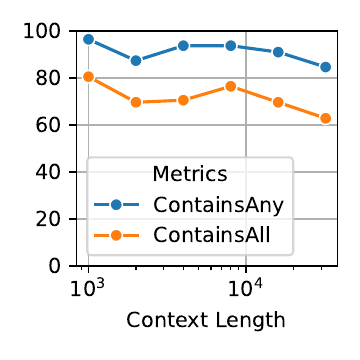}
        \vskip -0.1in
        \caption{MV-NIAH: 1 key, 2 values.}
        \label{fig:appdx:error:aggregation:mv}
    \end{subfigure}
    \begin{subfigure}[b]{0.37\linewidth}
        \small
        \includegraphics[width=0.9\linewidth]{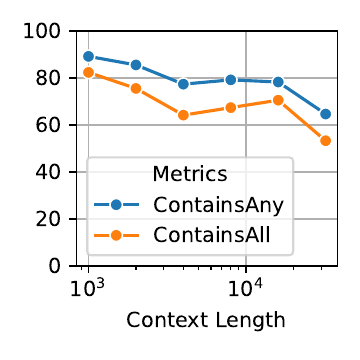}
        \vskip -0.1in
        \caption{MQ-NIAH: 2 keys, 1 value each.}
        \label{fig:appdx:error:aggregation:mq}
    \end{subfigure}
    \caption{Metric \texttt{containsAny} and \texttt{containsAll} comparison on aggregation subtasks in \ours-Retrieval (context length from 1k to 32k) evaluated with Qwen2.5-VL-7B-Instruct.}
    \label{fig:appdx:error:aggregation}
\end{figure}

\paragraph{Inaccurate Information Aggregation} 
To assess the capacity of VLMs to perform information aggregation, we inspect the aggregation subtasks. A representative example is \ours-Retrieval's Multi‑Value-NIAH, in which two needles share an identical key but are annotated with two distinct values, and are away from each other at least 20\% of total context length. The model is required to synthesize these cues and output a composite answer that reflects both of the two values. 
As shown in \Cref{fig:appdx:error:aggregation}, the performance gap between \texttt{containsAny} (finding at least one value) and \texttt{containsAll} (finding both) widens slightly with increasing context length. This result suggests that the primary bottleneck is not the aggregation logic itself, but rather the failure to retrieve all relevant pieces of information from the extensive visual context. The models are generally capable of synthesizing facts they can find, but their ability to find all facts diminishes over longer sequences.

\begin{table}[!ht]
    \centering 
    \small 

    \caption{Ablation with Glyph against different text background color---green in $\renderlocomo$, or white in $\renderdefault$.}
    \label{tab:exp:glyph:locomo:plain}

    \begin{tabular}{lcccc}
    \toprule
        Glyph & \multicolumn{4}{c}{\ours-Memory (\texttt{ROUGE-L} $\uparrow$)} \\
        \cmidrule(lr){2-5}
        Rendering & SingleHop & MultiHop & Temporal & OpenDomain \\
        \midrule
        $\renderlocomo$  & 4.10  & 2.43 & 1.69 & 2.16  \\
        $\renderdefault$ & 30.92 & 6.49 & 7.48 & 13.54 \\
    \bottomrule
    \end{tabular}
\end{table}

\paragraph{Sensitivity to Rendering Parameters} OCR-specific VLMs are known to be sensitive to the parameters used when rendering text images, e.g., image resolution, typographic style, and background colour. In particular, Glyph reports its lack of generalization to various rendering styles. To assess this, we applied a plain-text rendering style $\renderdefault=\renderlocomo \cup \render(\texttt{bg-color}=\texttt{white})$ for \ours-Memory and observed a notable performance gap shown in \Cref{tab:exp:glyph:locomo:plain}, thereby corroborating the Glyph's claim that it is highly sensitive to rendering parameters.

\begin{table*}[ht]
    \centering 
    \scriptsize \setlength{\tabcolsep}{2pt}

    \caption{Ablation with Deepseek-OCR against different prompt---completion (e.g., \textit{one of the magic numbers of long-context is \dots}), or QA (e.g., \textit{What is one of the magic numbers of long-context?}).}
    \label{tab:exp:dsocr:completion:qa}
\renewcommand{\arraystretch}{1.2}
\begin{tabular}{lcccccccccccccccccccc}
\toprule
& \multicolumn{20}{c}{\ours-Retrieval (\contains~$\uparrow$)} \\
\cmidrule(lr){2-21}
& \multicolumn{5}{c}{S-NIAH} & \multicolumn{5}{c}{MK-NIAH} & \multicolumn{5}{c}{MQ-NIAH} & \multicolumn{5}{c}{MV-NIAH} \\
\cmidrule(lr){2-6} \cmidrule(lr){7-11} \cmidrule(lr){12-16} \cmidrule(lr){17-21}
Prompt & 1k & 2k & 4k & 8k & 16k & 1k & 2k & 4k & 8k & 16k & 1k & 2k & 4k & 8k & 16k & 1k & 2k & 4k & 8k & 16k\\
\midrule
Completion & 51.82 & 50.91 & 50.00 & 16.36 & 1.82 & 40.00 & 50.91 & 51.82 & 12.73 & 0.91 & 21.82 & 18.18 & 19.09 & 6.82 & 0.00 & 20.91 & 24.09 & 17.73 & 14.55 & 1.36 \\
QA         & 30.91 & 32.73 & 25.45 & 0.00  & 0.00 & 23.64 & 26.36 & 17.27 & 0.00  & 0.00 & 7.73  & 11.36 & 6.82  & 0.00 & 0.00 & 10.45 & 16.82 & 9.09  & 1.36  & 0.00 \\
\bottomrule
\end{tabular}
\end{table*}

\paragraph{Sensitivity to Prompt} OCR-specific VLMs may also be sensitive to the evaluation prompt. Deepseek-OCR's training pipeline does not include a supervised-finetuning stage \cite{wei2025deepseek}, resulting in a lack of instruction-following abilities. We used a completion prompt instead of question-answering, which showed substantial improvements across all subtasks in \ours-Retrieval (\Cref{tab:exp:dsocr:completion:qa}).

\paragraph{Thumbnail Type Models}
A specific failure mode arises from the architectural design of certain VLMs, such as InternVL3.5~\cite{wang2025internvl3} and GPT-5~\cite{GPT5}.
This approach typically involves encoding both a downscaled, low-resolution ``thumbnail'' of the entire image to capture global context, alongside high-resolution tiles for fine-grained detail. 
InternVL3.5 explicitly processes a $448\times448$ thumbnail, which costs 256 tokens. For proprietary models like GPT-5~\cite{GPT5}, the exact internal architecture is opaque. However, we hypothesize a similar structure based on its token calculation formula\footnote{\url{https://platform.openai.com/docs/guides/images-vision}}, which adds a fixed 70 ``base tokens'' on top of the tokens for high-resolution tiles. 
This suggests the base tokens are used for a thumbnail-like global overview.
While this strategy is highly effective for natural images, where global composition and local details are distinct, it proves inefficient and counter-productive in our VTC setting. The images in VTCBench consist of uniformly dense text. When such an image is downscaled to create a thumbnail, the individual characters and words blur into complete illegibility. The resulting thumbnail offers the model no useful textual information, appearing as little more than a noisy, gray texture that the model can hardly see clearly.
Consequently, the tokens generated from this thumbnail are effectively wasted. For InternVL3.5, this means 256 tokens, a fifth of the total 1280 tokens for an $896\times896$ image, are spent processing a visually indecipherable global overview. Under our hypothesis for GPT-5, its 70 base tokens are similarly allocated to this low-information signal. 
This token waste represents a fundamental mismatch between the model's vision architecture and the specific nature of VTC data. The model expends a significant portion of its visual processing budget on a component that cannot contribute to the core task of text recognition, which likely contributes to the pronounced performance degradation observed in models like the InternVL3.5 series (\Cref{tab:exp:all:in:one}).

\section{VTCBench-Wild}
Our ablation studies in \Cref{tab:exp:ablation} and the error analysis in \Cref{sec:error}
have demonstrated that the performance of VLMs is sensitive to various rendering parameters, such as font size, color, and line height. These findings indicate that evaluating models using a single, fixed rendering configuration may not fully capture their robustness to the visual diversity encountered in real-world documents. To bridge this gap and establish a more challenging evaluation standard, we introduce \textbf{VTCBench-Wild}, a new benchmark variant designed to simulate visually diverse ``in-the-wild'' scenarios.

\subsection{Construction of VTCBench-Wild}
VTCBench-Wild is constructed by sampling scenes from a dynamic rendering pool and questions from a comprehensive question pool.

\paragraph{Rendering Pool} The core idea of VTCBench-Wild is to assess VLM performance under conditions of inconsistent and varied visual styling. To achieve this, we first constructed a diverse pool of 99 distinct rendering hyperparameter configurations. This pool was created by systematically varying key CSS properties from the $\render_\texttt{plain}$  baseline, including:

\begin{itemize}
    \item \emph{Font Size:} Ranging from 10 pixels to 20 pixels.
    \item \emph{Font Family:} Including Times New Roman, Helvetica, and  Courier New fonts.
    \item \emph{Line Height:} Varying among values 1.0, 1.2, and 1.5.
\end{itemize}
For each test instance in VTCBench-Wild, each sample is rendered using a configuration randomly sampled from this pool of 99 styles (11$\times$3$\times$3).

\paragraph{Question Pool} 
To create a diverse set of evaluation questions that test different aspects of long-context understanding, the question pool is constructed by systematically generating questions for VTC-Retrieval and VTC-Reasoning, and by sampling from the existing dataset for VTC-Memory.
For VTC-Retrieval and VTC-Reasoning, which are based on synthetic ``needle-in-a-haystack'' tests, we systematically vary two key parameters: haystack length and the depth of the ``needle'' (i.e., the target information).
\begin{itemize}
    \item \emph{Haystack Length:} We generate test cases for six distinct context lengths: 1k, 2k, 4k, 8k, 16k, and 32k tokens.
    \item \emph{Needle Depth:} To evaluate how a model’s performance is affected by the location of crucial information, we generate separate test instances by inserting a single needle at various depths. For each context length, this evaluation is repeated for every specified depth, and the results are averaged. This process is conducted at 10\% Granularity, i.e., creating 11 separate test cases, placing the needle at depths of 0\%, 10\%, 20\%, ..., up to 100\%.
\end{itemize}
For VTC-Memory, which utilizes a predefined dataset of natural conversations, our goal is to preserve the original data distribution. Instead of synthetically controlling variables, we use 1,540 question-answer pairs from the full VTC-Memory dataset. 
Finally, our question pool contains 28,600 unique questions, comprising 7,920 retrieval questions, 19,140 reasoning questions, and 1,540 memory questions.
By combining the question pool with the rendering pool, we obtain 2,831,400 unique combinations.
From these combinations, we sample 600 items for memory, 800 items for reasoning, and 800 items for retrieval to form the final VTCBench-Wild.

\subsection{Evaluation Metrics}
For the retrieval and reasoning questions, we use \contains~to judge if the ground truth answer is in the prediction.
For the memory questions, we use gpt-4o-mini~\cite{hurst2024gpt} to judge the correctness of the answer.

\begin{table}[t]
\centering 
\small
\caption{Experimental results (\%) on VTCBench-Wild.}
\label{tab:vtcwild}
\begin{tabular}{l|ccc|c}
\toprule
Model & Retrieval & Reasoning & Memory & Overall\\ 
\midrule
Qwen3-8B (Text Only)~\cite{yang2025qwen3} & 99.38 &62.25 &54.17&73.55\\
Gemini-3-Pro (Text Only)~\cite{gemini3} & 100.0&99.64&64.67&89.29\\
\midrule
GPT-5~\cite{GPT5} & 72.63 & 62.88 & 43.83 & 61.23 \\
Gemini-2.5-Pro~\cite{comanici2025gemini} & 87.38 & 58.75 & 38.67 & {63.68} \\
Gemini-3-Pro~\cite{gemini3} & \textbf{99.88} & \textbf{97.63} &\textbf{58.00} & \textbf{87.64}\\
Claude-Sonnet-4~\cite{claude} &68.25&8.75&36.83&38.05 \\
Claude-Sonnet-4.5~\cite{claude} &76.00&10.88&42.83&43.27 \\
Seed-1.5-Pro~\cite{guo2025seed1} &76.88&35.88&42.83&52.68\\
Qwen3-VL-8B~\cite{Qwen3-VL} & 89.00 & 11.50 & 33.67 & 45.73 \\
Qwen3-VL-235B-A22B~\cite{Qwen3-VL} & {97.50} & 14.63 & {48.50} & 54.00 \\
InternVL3.5-8B~\cite{wang2025internvl3} & 51.38 & 7.63 & 17.33 & 26.18 \\
InternVL3.5-38B~\cite{wang2025internvl3} & 45.81 & 8.75 & 22.33 & 25.93 \\
Qwen2.5-VL-7B~\cite{bai2025qwen2} & 91.63 & 15.63 & 33.83 & 48.23 \\
Qwen2.5-VL-72B~\cite{bai2025qwen2} & 91.50 & 37.50 & 38.33 & 57.36 \\
Gemma3-27B~\cite{team2025gemma} & 49.38 & 3.75 & 17.33 & 24.05 \\
Kimi-VL-A3B-Instruct~\cite{team2025kimi} & 82.50 & 8.75 & 33.33 & 42.27 \\
LLaVA-OneVision-7B~\cite{lillava}  &22.75 &2.88 & 0.83 &9.55 \\
Glyph~\cite{cheng2025glyph} & 93.63 & 36.75 & 36.17 & 57.27 \\
GLM-4.1V-9B-Thinking~\cite{hong2025glm} & 76.75 & {43.13} & 38.17 & 54.00 \\
\bottomrule
\end{tabular}
\end{table}

\subsection{Experimental Results and Analysis}
We evaluated 17 VLMs and two text-only LLM baselines on VTCBench-Wild. The results, presented in \Cref{tab:vtcwild}, reveal a clear hierarchy of capabilities and highlight architectural differences in processing visually diverse, compressed textual information. We structure our analysis around three distinct groups, discussing both the challenges and the potential of the VTC paradigm:

\begin{itemize}
\item \textbf{LLM Baselines Illuminate Both the Gap and the Potential of VTC.}
The LLM baselines provide a crucial benchmark, revealing a persistent performance gap for the majority of VLMs. While text-only models like Qwen3-8B (73.55\%) and Gemini-3-Pro (89.29\%) set a high bar, most VLMs fail to match this level of performance, suggesting that the visual encoding stage often introduces an overhead or information loss.
However, a critical exception to this trend is the Gemini-3-Pro VLM. With an overall score of 87.64\%, it not only approaches the performance of its text-only counterpart but also decisively outperforms the strong Qwen3-8B LLM baseline. This standout result is highly significant: it serves as a compelling proof of concept for the VTC paradigm. It demonstrates that with a sufficiently advanced and well-integrated vision-language architecture, the challenges of visual compression can be overcome. Far from being an insurmountable barrier, this finding validates VTC as a viable, and potentially highly effective, pathway for scaling long-context processing.
\item \textbf{Proprietary Models Lead, but Performance Varies Significantly.} Among the VLMs, a clear stratification is evident. The VLM version of Gemini-3-Pro is the undisputed leader, achieving an overall score of 87.64\% that nearly matches its text-only counterpart. This indicates a highly robust vision-language architecture capable of handling the visual variance in VTCBench-Wild with minimal performance degradation. Following at a considerable distance are Gemini-2.5-Pro (63.68\%) and GPT-5 (61.23\%). These models exhibit different strengths: Gemini-2.5-Pro shows stronger retrieval capabilities (87.38\%), whereas GPT-5 has a slight edge in reasoning (62.88\%). In contrast, other proprietary models like the Claude series show markedly lower performance, particularly on reasoning tasks, highlighting their sensitivity to the benchmark's visual diversity.
\item \textbf{Open-Source Models Reveal Distinct Architectural Profiles.} The performance of open-source models uncovers diverse strengths and weaknesses. Glyph (57.27\%) and Qwen2.5-VL-72B (57.36\%) emerge as the most balanced and capable, achieving strong scores across retrieval and reasoning. A standout performer is GLM-4.1V-9B-Thinking, which, despite a modest overall score, achieves a remarkable 43.13\% on the reasoning task. This is the highest among all open-source VLMs and is competitive even with leading proprietary models. 
\end{itemize}

Overall, the results from VTCBench-Wild affirm that while many modern VLMs possess strong text perception abilities, this often does not translate to robust, deep contextual understanding. The Memory task proves to be a formidable challenge for nearly all models, with only Gemini-3-Pro scoring above 50\%. The stark gap between retrieval and reasoning performance in models like Qwen3-VL highlights that true long-context comprehension requires more than just accurate OCR. VTCBench-Wild thus serves as an effective diagnostic tool, revealing the architectural resilience and specific failure points of VLMs in realistic scenarios.

\section{Conclusion}
In this work, we conducted the first systematic investigation into the long-context understanding capabilities of VLMs under the vision-text compression (VTC) paradigm. We introduced comprehensive benchmarks VTCBench and VTCBench-Wild, which evaluate models on three complex tasks: information retrieval, associative reasoning, and long-term memory. Comprehensive experiments on a range of leading VLMs yielded valuable findings. While current models exhibit a respectable ability to perform simple text retrieval from compressed images, their performance is fragile, degrading with increased context length and compression ratios. More critically, we observed a near-total collapse in their ability to perform associative reasoning and maintain long-term memory, revealing a significant gap between visual perception and deep semantic comprehension. Remarkably, with VTC, Gemini-3-Pro can outperform the LLM baseline (Qwen3-8B) on VTCBench-Wild. This exceptional performance validates VTC as a promising and viable paradigm for long-context comprehension.
Furthermore, our analysis underscores that rendering parameters, particularly font size, are not a trivial implementation detail but a critical factor directly influencing model performance. High compression ratios achieved through smaller fonts severely impair even basic retrieval, confirming that VTC introduces unique perceptual challenges absent in text-only models.

Our evaluation demonstrates that VTC is not a simple, drop-in solution for the long-context problem. The efficiency gains of the VTC paradigm come at the cost of sacrificing advanced cognitive capabilities and visual robustness. Moving forward, the development of truly effective VTC-based models will require more than just scaling existing VLM architectures. It necessitates targeted research into novel pre-training objectives and architectural designs that explicitly bridge the gap between spatial perception and abstract, long-range semantic reasoning. By establishing a rigorous evaluation framework and highlighting these core limitations, we hope VTCBench will guide and accelerate the creation of next-generation VLMs that are both efficient and genuinely capable of understanding long contexts.

\renewcommand\thefigure{S\arabic{figure}}
\setcounter{figure}{0}
\renewcommand\thetable{S\arabic{table}}  
\setcounter{table}{0}
\renewcommand\theequation{S\arabic{equation}}
\setcounter{equation}{0}


\clearpage
\beginappendix



%

\section{VTCBench Description}\label{sec:appdx:A}

VTCBench is the first systematic benchmark to evaluate the long-context understanding of vision-language models (VLMs) using vision-text compression (VTC), a paradigm that renders long text into compact images to achieve substantial token compression. The benchmark assesses models on three tasks: information retrieval (VTC-Retrieval), associative reasoning (VTC-Reasoning), and long-term dialogue comprehension (VTC-Memory). 
We will introduce how we curate \ourbench~and show some examples.

\subsection{Rendering Pipeline}\label{sec:render_pipeline}

\begin{figure}[!ht]
\centering

\begin{minipage}{.48\textwidth}
\begin{tikzpicture}[
    ->,
    >=Stealth,
    font=\small\sffamily,
    node distance=.25in and .2in,
    every node/.style={align=center, inner sep=4pt},
    block/.style={draw, minimum width=1in, minimum height=.25in},
    io/.style={minimum width=.9in, rounded corners},
]

\node[block, io, fill=orange!10] (A) {%
    \faIcon[regular]{file-lines}~\textbf{context}
};
\node[block, below=of A, fill=cyan!10] (B) {%
    \faIcon{markdown}~\textbf{markdown‑it}
};
\node[block, below=of B, fill=cyan!10] (C) {%
    \faIcon{chrome}~\textbf{playwright}
};
\node[block, below=of C, fill=green!10] (E) {%
    \faIcon{paste}~\textbf{pymupdf}
};
\node[block, io, below=of E, fill=orange!10] (D) {%
    \faIcon{images}~\textbf{image(s)}
};

\draw[->] (A) -- node[midway, right] {\faIcon{align-left} text} (B);
\draw[->] (B) -- node[midway, right] {\faIcon{code} HTML} (C);
\draw[->, dotted] (C) to[out=-160,in=160] node[midway, left, align=right] {\faIcon[regular]{camera} screenshot\\(exactly 1 image)} (D);
\draw[->] (C) -- node[midway, right] {\faIcon[regular]{file-pdf} pdf} (E);
\draw[->] (E) -- node[midway, right] {\faIcon[regular]{image} page-to-image} (D);

\end{tikzpicture}
\vspace{10pt}
\caption{
Rendering pipeline of \ourbench.
\faIcon{markdown} \textbf{markdown‑it} interprets context as HTML.
\faIcon{chrome} \textbf{playwright} is a browser-based rendering engine prints HTML to PDF, where styles are injected as CSS.
\faIcon{paste} \textbf{pymupdf} converts each page in PDF as an image.
}
\label{fig:appdx:render:pipeline}

\end{minipage}\hfill
\begin{minipage}{.48\textwidth}
    \vspace{-30pt}
    \centering
    \includegraphics[width=\linewidth]{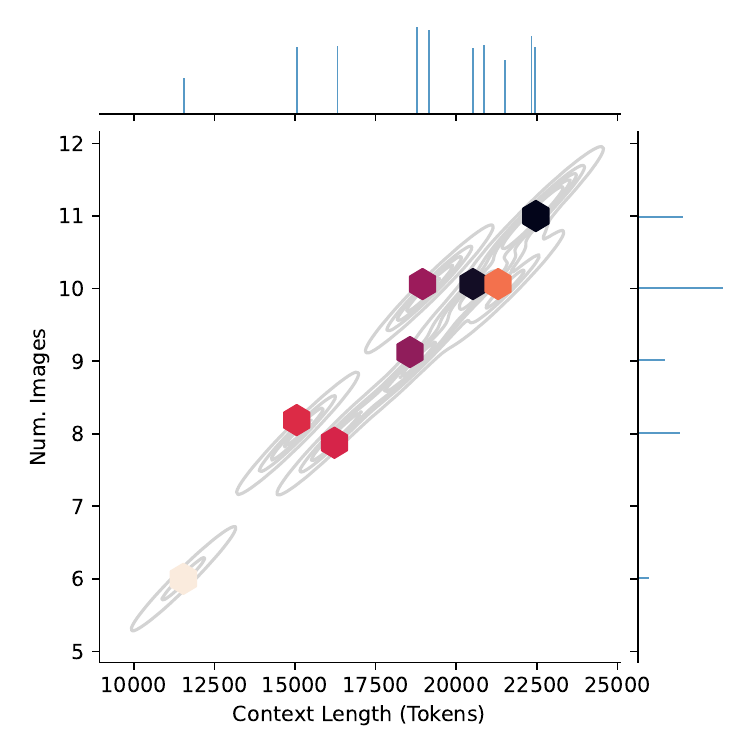}
    \caption{Distribution of context length (in number of text tokens) versus number of images for \ours-Memory with rendering operator $\renderlocomo$. A strong correlation of approximately 2,000 tokens per image can be inferred.}
    \label{fig:appdx:locomo:distribution}
\end{minipage}

\end{figure}

\Cref{fig:appdx:render:pipeline} depicts the rendering pipeline, a real-time browser‑based text-to-image engine, which stacks
markdown-it\footnote{\url{https://pypi.org/project/markdown-it-py}}, 
playwright\footnote{\url{https://pypi.org/project/playwright}},
and pymupdf\footnote{\url{https://pypi.org/project/pymupdf}} 
for end-to-end conversion
and applies text styling by injecting CSS. 
For example, the rendering preset $\renderdefault$ contains only the basic rule: \texttt{"p\{font-size:12px;\}"}.
Crucially, $\render$ significantly influences the compression ratio $\compratio$: \ours-Reasoning with $\renderdefault$ yields an average of 2,000 text tokens per image, whereas employing $\render_\texttt{plain-16}$(\texttt{"p\{font-size:16px;\}"}) results in only 1,100 tokens per image, nearly halving $\compratio$.
As for \ours-Memory, which has predefined contexts, we observed a text-token versus number-of-image distribution in \Cref{fig:appdx:locomo:distribution}, using rendering preset $\renderlocomo$.

\subsection{Examples of VTCBench}

\begin{figure*}[p]
\centering

\begin{subfigure}[b]{0.32\textwidth}
    \centering
    \includegraphics[width=\textwidth]{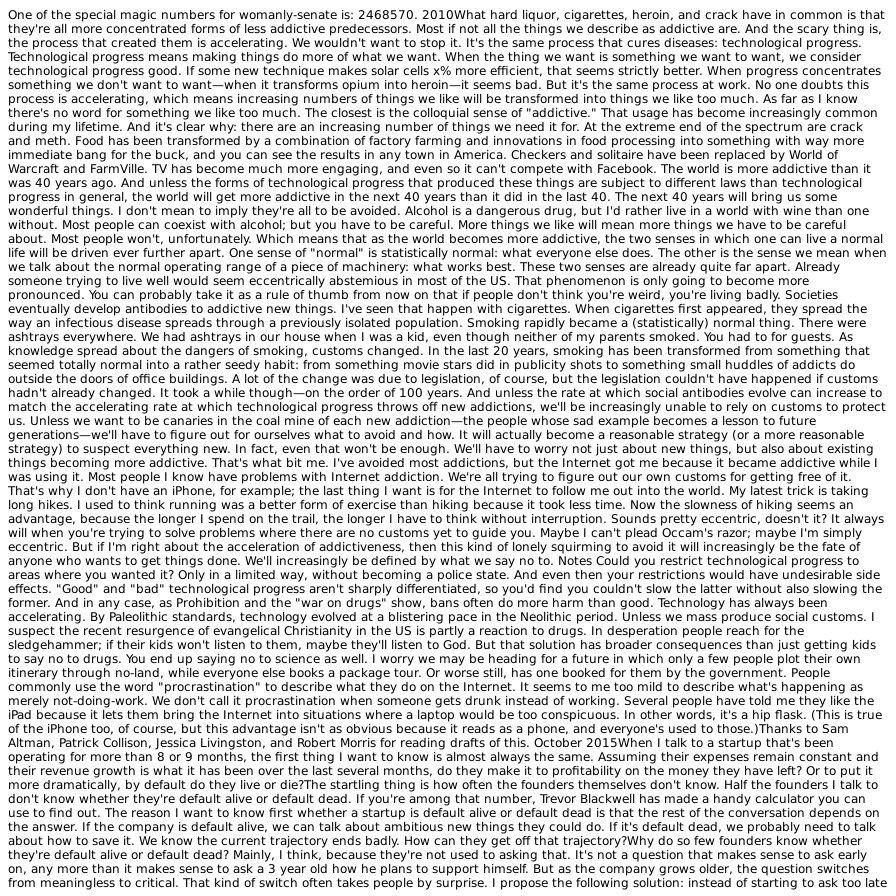}
    \caption{VTC-Retrieval example with a context length of 2k.}
    \label{fig:appdx:example:ruler}
\end{subfigure}
\hfill
\begin{subfigure}[b]{0.32\textwidth}
    \centering
    \includegraphics[width=\textwidth]{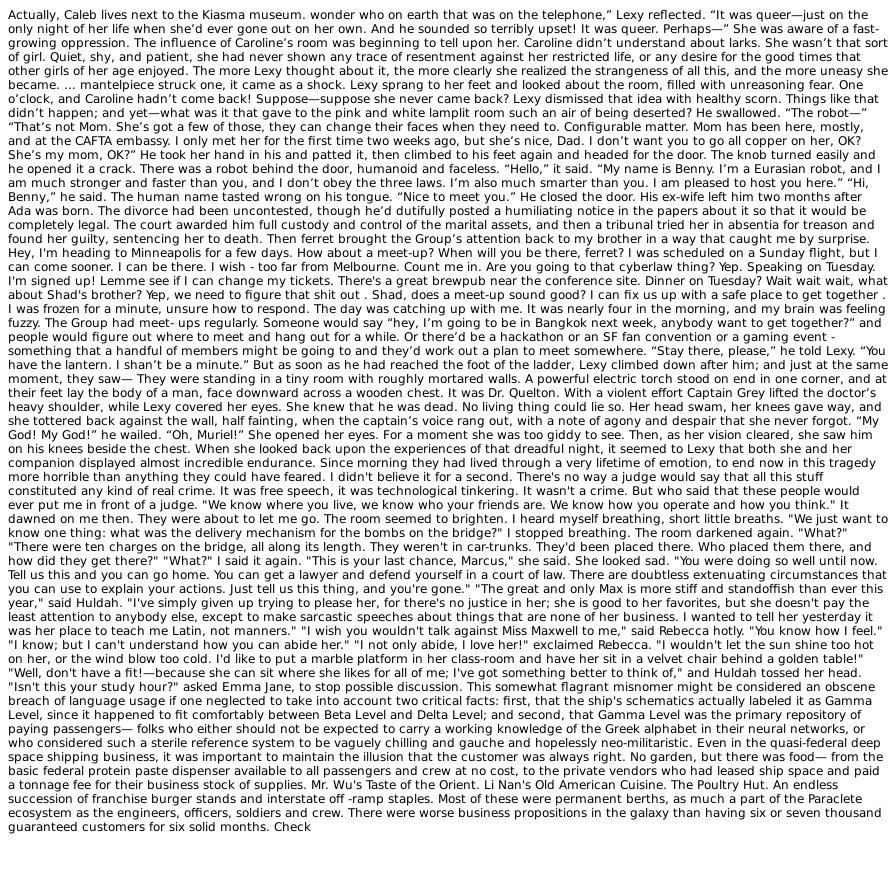}
    \caption{VTC-Reasoning example, with a context length of 2k.}
    \label{fig:appdx:example:nolima}
\end{subfigure}
\hfill
\begin{subfigure}[b]{0.32\textwidth}
    \centering
    \includegraphics[width=\textwidth]{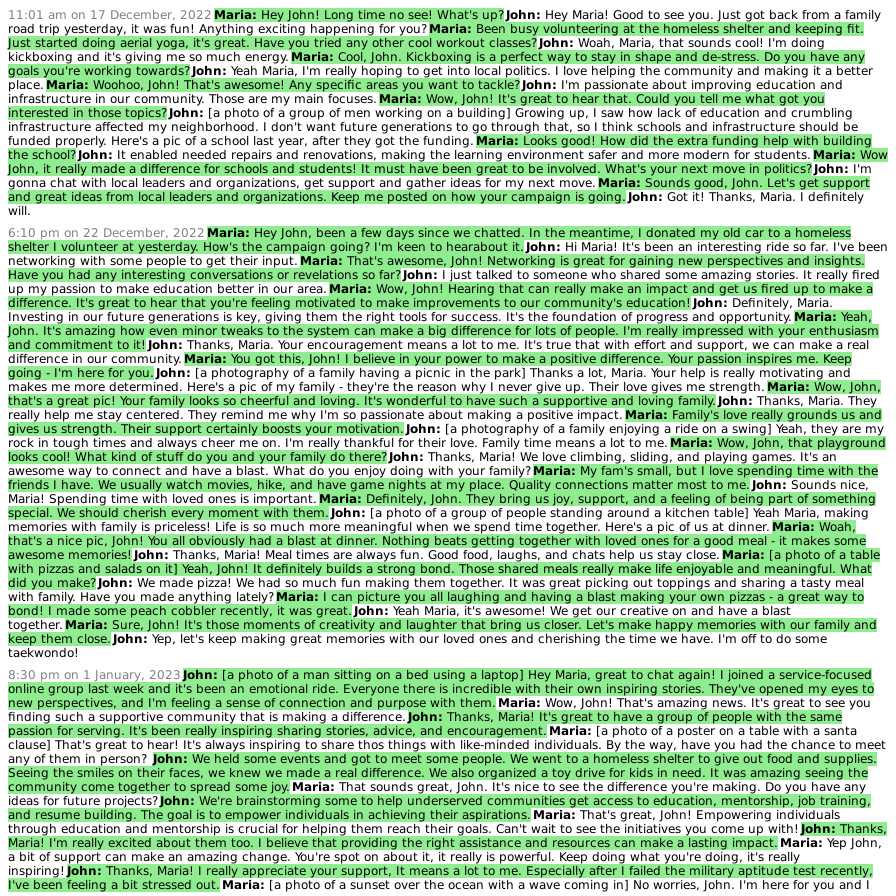}
    \caption{VTC-Memory example (1st of 15), about 30,000 text tokens in total.}
    \label{fig:appdx:example:locomo}
\end{subfigure}
\caption{Examples of Rendered Context for VTC-Retrieval, VTC-Reasoning, VTC-Memory using $\renderdefault$, $\renderdefault$, and $\renderlocomo$, respectively. }
\label{fig:appdx:examplerendered}
\end{figure*}

We permute configurations including task type, context length, evaluation template, and needle type (\Cref{tab:appdx:example}) and dynamically generate examples; 
for each configuration, we sample from its pool (including needle keys, needle values, and distractors) to fill placeholders in the haystack to form the context.
The rendering pipeline then converts the context to images (\Cref{fig:appdx:examplerendered}). We concatenate the visual context and a prompt (usually an instruction followed by a question) as input for VLMs. Prompt templates are provided in \Cref{sec:appdx:prompt}.

\begin{table*}[p]

\centering
\footnotesize
\setlength{\tabcolsep}{5pt}
\caption{Illustrations of VTC-Retrieval, VTC-Reasoning, and VTC-Memory examples. We highlight {\color{orange}queries}, {\color{orange}keys}, {\color{blue}values}, and {\color{gray}distractors} accordingly. For VTC-Retrieval and VTC-Reasoning, {\color{orange}queries}, {\color{orange}keys}, {\color{blue}values}, and {\color{gray}distractors} are randomly generated; for VTC-Memory, {\color{orange}queries}, {\color{orange}keys}, {\color{blue}values}, and {\color{gray}distractors} are provided as-is.}

\begin{tabular}{c|c|l|l}
    \toprule
    Task & Task Categories & Context Example & Evaluation Example \\
    \midrule
        \makecell[c]{
            VTC-Retrieval \\(NIAH)
        } 
        & \makecell[c]{
            Lexical Matching,\\
            Multi-Hop Tracing, \\
            Aggregation
        } 
        & \makecell[l]{
            (Dynamic {\color{orange}query/key}-{\color{blue}value} with types: \\
            {\color{orange}word}-{\color{blue}word}, {\color{orange}word}-{\color{blue}number}, {\color{orange}uuid}-{\color{blue}number}.) \\
            {\color{gray}(essays) $\dots$} \\ 
            One of the special magic numbers for \\
            {\color{orange}long-context} is: {\color{blue}2026}. \\
            {\color{gray}$\dots$} \\
            {\color{gray}One of the special magic numbers for} \\
            {\color{gray}distracting-information is: 2025}. \\
            {\color{gray}$\dots$} \\
        }
        & \makecell[l]{
            \textbf{QA Variant}. \\
            \textit{Q:} What's the special magic \\
            number for {\color{orange}long-context}? \\
            \textit{A:} {\color{blue}2026}. \\
            \textbf{Completion Variant}. \\
            \textit{Prompt:} one of the special magic \\
            numbers for {\color{orange}long-context} is: \\
            \textit{Completion:} {\color{blue}2026}. \\
        } \\ 
    \midrule
        \makecell[c]{VTC-Reasoning\\(NIAH)} 
        & \makecell[c]{
            Associative Reasoning,\\Question-Answering
        } 
        & \makecell[l]{
            (Dynamic {\color{orange}query/key}-{\color{blue}value} with types: \\
            {\color{orange}event/action}-{\color{blue}person}.) \\
            {\color{gray}(books) $\dots$} \\ 
            There was a {\color{orange}vegan} guest, named {\color{blue}Katie}. \\
            {\color{gray}$\dots$} \\
        } 
        & \makecell[l]{
            \textbf{One-Hop Reasoning}. \\
            \textit{Q:} Which character cannot \\
            eat {\color{orange}fish-based} meals?\\
            \textit{A:} {\color{blue}Katie}. \\
            \textbf{Two-Hop Reasoning}. \\
            \textit{Q:} Which character cannot \\
            eat {\color{orange}Brandade} meals?\\
            \textit{A:} {\color{blue}Katie}.
        } \\
    \midrule
        \makecell[c]{VTC-Memory\\(QA)} 
        & \makecell[c]{Memory,\\Question-Answering}
        & \makecell[l]{
            (No dynamic {\color{orange}query/key}-{\color{blue}value}, fully static.) \\
            {\color{gray}(conversations) $\dots$} \\ 
            \textit{\color{orange}Caroline:} {\color{blue}Researching adoption agencies} \\
            ---it's been a dream to have a family and \\
            give a loving home to kids who need it. \\
            {\color{gray}$\dots$} \\
            \textit{\color{orange}Caroline:} And here's one of the \\
            {\color{blue}adoption agencies} I'm looking into. \\
            {\color{gray}$\dots$}
        } 
        & \makecell[l]{
            \textit{Q:} What did {\color{orange}Caroline} research?\\
            \textit{A:} {\color{blue}Adoption agencies}.
        } \\
    \bottomrule
\end{tabular}

\label{tab:appdx:example}

\end{table*}
\clearpage

\section{Calculation of Compression Ratio}
\label{sec:appdx:compratio:calc}

Recall that the VTC ratio $\compratio=\frac{N_T}{N_I}$ is the ratio between a context of $N_T$ text tokens and its image counterpart requiring $N_I$ image tokens. This ratio is affected by both the rendering operator $\render: \texttt{text}\mapsto\texttt{image}$ and VLM's vision processing pipeline, so we provide some calculation examples of $\compratio$ for VLMs given $\render=\renderdefault$.
For simplicity, the following formulas ignore model-specific resizings and paddings, assuming an input image of size $(H, W)$ pixels.
According to different image-processing strategies, we can roughly categorize open-source multimodal LLMs into three groups: 
\textit{(1)} Models with dynamic resolution like the Qwen-VL series~\cite{bai2025qwen2,wang2024qwen2,Qwen3-VL}, Kimi-VL~\cite{team2025kimi}, and GLM-4V~\cite{cheng2025glyph,hong2025glm};
\textit{(2)} Models using a global thumbnail, represented by the InternVL family~\cite{wang2025internvl3};
\textit{(3)} Other architectures with different visual pipelines like Deepseek-OCR~\cite{wei2025deepseek} and Gemma3~\cite{team2025gemma} family.

\subsection{Dynamic Resolution Models}
\paragraph{Qwen2.5-VL Series~\cite{bai2025qwen2}}
Qwen2.5-VL is a multimodal vision-language model series developed by the Qwen Team at Alibaba Group that understands and processes both images, videos, and text. It uses a NaViT~\cite{dehghani2023patch} structure, and the visual token can be obtained by $\frac{H \times W}{14 \times 14\times 4}$. Specifically, an $896\times896$ image is 1024 tokens.

\paragraph{Qwen3-VL Series~\cite{Qwen3-VL}}
Qwen3-VL is the latest and most powerful series of open-source VLMs developed by the Qwen team.
It is available in various versions, including dense and MoE (Mixture of Experts) architectures and Instruct and Thinking editions. 
Qwen3-VL uses a Siglip2~\cite{tschannen2025siglip} vision encoder with dynamic resolution, and the number of visual tokens injected into the LLM is $\frac{H \times W}{16 \times 16\times 4}$. Specifically, an $896\times896$ image is 784 tokens.

\paragraph{GLM-4.1V~\cite{hong2025glm} \& Glyph~\cite{cheng2025glyph}}
GLM-4.1V and Glyph use a similar structure as Qwen2.5-VL, and the calculation formula is $\frac{H \times W}{14 \times 14\times 4}$.

\paragraph{Kimi-VL-A3B-Instruct~\cite{team2025kimi}}
Kimi-VL-A3B-Instruct, designed by MoonshotAI, is an efficient open-source VLM that uses a MoE architecture to handle multimodal tasks like image, video, and text understanding.
Kimi-VL employs MoonViT as its native-resolution vision encoder; MoonViT is initialized from and continually pre-trained on SigLIP-SO-400M~\cite{zhai2023sigmoid}.
Its image patch size is 
$14\times14$; it performs a $2\times2$ spatial downsampling while correspondingly expanding the channel dimension. Consequently, the number of visual tokens is $\frac{H \times W}{14 \times 14\times 4}$.
For an $896\times896$ image input, we have 1024 visual tokens.

\subsection{Thumbnail-based Models}
\paragraph{InternVL3.5 Series~\cite{wang2025internvl3}}
InternVL3.5 uses InternViT~\cite{chen2024internvl} to split an image into $448\times448$ tiles and a $448\times448$ thumbnail to capture the global context. Each tile refers to 256 tokens.
For an $896\times896$ image input, we have $5 \times 256=1280$ tokens.

\paragraph{GPT-5~\cite{GPT5}} 
GPT-5 is a VLM developed by OpenAI, representing the fifth iteration in its series of generative pre-trained transformer (GPT) foundation models. We use GPT-5 with \texttt{detail=high} for all our experiments, and the  pipeline\footnote{\url{https://platform.openai.com/docs/guides/images-vision}} is:

\begin{enumerate}
    \item Scale to fit in a $2048\times 2048$px square. No effect here because all our images satisfy this threshold.
    \item Scale so that the image's shortest side is $768$px long. Images are resized to $768\times 768$px.
    \item Count the number of $512$px squares (tiles) in the image, each costing 140 tokens. 4 tiles cost $560$ tokens.
    \item Add the base tokens to the total. GPT-5 has $70$, yielding a final $560+70=630$ tokens per image.
\end{enumerate}

\subsection{Others}
\paragraph{Gemini-2.5-Pro~\cite{comanici2025gemini}} 
Gemini-2.5-Pro is Google's advanced reasoning model, excelling at complex problem-solving, coding, and analyzing large datasets, documents, and multimedia content. 
Gemini-2.5-Pro\footnote{\url{https://ai.google.dev/gemini-api/docs/image-understanding}} tiles images into $768 \times 768$ pixel tiles, each costing 258 tokens. This means an $896\times896$ image would cost $258\times4=1032$ tokens.

\paragraph{Deepseek-OCR~\cite{wei2025deepseek}}
DeepSeek-OCR is an open-source AI system that uses an innovative ``Contexts Optical Compression'' approach to read and compress documents, making it more efficient for LLMs.
We use the Gundam model for Deepseek-OCR, and the visual token is $4 \times 100 + 256 = 656$.

\paragraph{Gemma3 Series~\cite{team2025gemma}} 
Gemma3 vision processor\footnote{\url{https://developers.googleblog.com/gemma-explained-whats-new-in-gemma-3}} uses SigLip~\cite{zhai2023sigmoid}, which resizes images to $896\times896$ pixels and applies 
$4\times4$ average pooling to output 256 tokens per image.
This high-factor pooling strategy ensures a substantial compression ratio, condensing the high-resolution visual data into compact embeddings.

\begin{table}[ht]
    \centering \small
    \caption{Inference parameters of VLMs and LLM baseline.}
    \label{tab:appdx:inferparams}
    \begin{tabular}{l|l}
    \toprule
        Model & Parameters \\
    \midrule
        Deepseek-OCR~\cite{wei2025deepseek} & \makecell[l]{
        \texttt{temperature=0,} \\
        \texttt{ngram\_size=30,} \\
        \texttt{window\_size=90,} \\
        \texttt{whitelist\_token\_ids=\{128821, 128822\},}\\
        \texttt{skip\_special\_tokens=False}
        } \\
    \midrule
        Gemma3 Series~\cite{team2025gemma} & \texttt{top\_k=64,top\_p=0.95} \\
    \midrule
        Gemini-2.5-Pro~\cite{comanici2025gemini} & 
        \texttt{reasoning\_effort=minimal}\\
    \midrule
        \makecell[l]{GLM-4.1V-9B-Thinking~\cite{hong2025glm}} & \makecell[l]{
        \texttt{top\_k=2,}\\
        \texttt{top\_p=0.6,}\\
        \texttt{temperature=0.8}}\\
    \midrule
        Glyph~\cite{cheng2025glyph} & \texttt{temperature=1.0} \\
    \midrule
        GPT-5~\cite{GPT5} & \makecell[l]{
        \texttt{detail=high,}\\
        \texttt{reasoning\_effort=minimal,}\\
        \texttt{verbosity=low}} \\
    \midrule
        \makecell[l]{InternVL3.5 Series~\cite{wang2025internvl3}} & None \\
    \midrule
        \makecell[l]{Kimi-VL-A3B-Instruct~\cite{team2025kimi}} & \texttt{temperature=0.2}\\
    \midrule
        \makecell[l]{Qwen2.5-VL Series~\cite{bai2025qwen2}} & \makecell[l]{
        \texttt{repetition\_penalty=1.05,}\\
        \texttt{temperature=0.000001}
        }\\
    \midrule
        \makecell[l]{Qwen3-VL Series~\cite{Qwen3-VL}} & \makecell[l]{
        \texttt{top\_k=20,}\\
        \texttt{top\_p=0.8,}\\
        \texttt{repetition\_penalty=1.0,}\\
        \texttt{temperature=0.7}
        }\\
    \midrule
        \makecell[l]{Qwen3-8B~\cite{yang2025qwen3}\\(LLM baseline)} & \makecell[l]{
        \texttt{top\_k=20,}\\
        \texttt{top\_p=0.95,}\\
        \texttt{temperature=0.6}
        }\\
    \bottomrule
    \end{tabular}
\end{table}

\section{Experimental Details}\label{sec:appdx:C}


\Cref{tab:appdx:inferparams} presents the inference parameters employed in our experiments. We follow the default generation configuration provided in HuggingFace and use the minimal thinking configuration possible.

\section{Prompts}
\label{sec:appdx:prompt}

{
\footnotesize
\begin{tcolorbox}[colback=black!5!white,
                  colframe=gray,
                  title=\ours-Retrieval]
\{haystack\} \{question\}
\end{tcolorbox}

\begin{tcolorbox}[colback=black!5!white,
                  colframe=gray,
                  title=\ours-Reasoning]
\{haystack\}
Answer a question based on the above book snippet. Your answer should be short and based on either explicitly stated facts or strong, logical inferences. Return only the final answer with no additional explanation or reasoning.
\{question\}
\end{tcolorbox}

\begin{tcolorbox}[colback=black!5!white,
                  colframe=gray,
                  title=\ours-Memory]
\{haystack\} Based on the above context, write an answer in the form of a short phrase for the following question. Answer with exact words from the context whenever possible. \{question\}
\end{tcolorbox}

\begin{tcolorbox}[colback=black!5!white,
                  colframe=gray,
                  title=\ours-Wild-Retrieval]
\{haystack\} Answer a question based on the above book snippet. Your answer should be short and based on either explicitly stated facts or strong, logical inferences. Return only the final answer with no additional explanation or reasoning. Question: \{question\}
\end{tcolorbox}

\begin{tcolorbox}[colback=black!5!white,
                  colframe=gray,
                  title=\ours-Wild-Reasoning]
\{haystack\} Answer a question based on the above book snippet. Some special magic numbers are hidden within the following text. Make sure to memorize it. I will quiz you about the numbers afterwards. Question: \{question\}
\end{tcolorbox}

\begin{tcolorbox}[colback=black!5!white,
                  colframe=gray,
                  title=\ours-Wild-Memory]
\{haystack\} Based on the above context, write an answer in the form of a short phrase for the following question. Answer with exact words from the context whenever possible. Question: \{question\}
\end{tcolorbox}

\begin{tcolorbox}[colback=black!5!white,
                  colframe=black!75!white,
                  title=LLM Judge]
Please as a grading expert, judge whether the final answers given by the candidates below are consistent with the standard answers, that is, whether the candidates answered correctly. Here are some evaluation criteria: 1. Please refer to the given standard answer. You don't need to re-generate the answer to the question because the standard answer has been given. You only need to judge whether the candidate's answer is consistent with the standard answer according to the form of the question. THE STANDARD ANSWER IS ALWAYS CORRECT AND THE QUESTION IS PERFECTLY VALID. NEVER QUESTION THEM. 2. ONLY compare the FINAL ANSWER - COMPLETELY IGNORE any potential errors in the REASONING PROCESSES. 3. Some answers may be expressed in different ways, such as some answers may be a mathematical expression, some answers may be a textual description, as long as the meaning expressed is the same. Before making a judgment, please understand the question and the standard answer first, and then judge whether the candidate's answer is correct. If the standard answer does not specify a unit, but the candidate's answer includes a unit that is correct for the value given, consider it correct. 4. Some answers may consist of multiple items, such as multiple-choice questions, multiple-select questions, fill-in-the-blank questions, etc. Regardless of the question type, the final answer will be considered correct as long as it matches the standard answer, regardless of whether the reasoning process is correct. For multiple-select questions and multi-blank fill-in-the-blank questions, all corresponding options or blanks must be answered correctly and match the standard answer exactly to be deemed correct. 5. If the prediction is given with {\textbackslash}boxed\{\{\}\}, please ignore the {\textbackslash}boxed\{\{\}\} and only judge whether the candidate's answer is consistent with the standard answer. 6. If the candidate's answer is invalid (e.g., incomplete (cut off mid-response), lots of unnormal repetitive content, or irrelevant to the question, saying it can't answer the question because some irresistible factors, like ethical issues, no enough information, etc.), select option C (INVALID). Please judge whether the following answers are consistent with the standard answer based on the above criteria. Grade the predicted answer of this new question as one of: A: CORRECT B: INCORRECT C: INVALID Just return the letters "A", "B", or "C", with no text around it. Here is your task. Simply reply with either CORRECT, INCORRECT, or INVALID. Don't apologize or correct yourself if there was a mistake; we are just trying to grade the answer.{\textbackslash}n
<Original Question Begin>: \{question\}{\textbackslash}n
<Original Question End>{\textbackslash}n
<Standard Answer Begin>: \{gold\_answer\}{\textbackslash}n
<Standard Answer End>{\textbackslash}n
<Candidate's Answer Begin>: \{prediction\}{\textbackslash}n
<Candidate's Answer End>{\textbackslash}n
Judging the correctness of the candidate's answer:
\end{tcolorbox}

}

\section{Limitation}\label{sec:appdx:D}
While this work provides the first systematic investigation into the long-context capabilities of VLMs under the VTC paradigm, we acknowledge several limitations, which offer clear directions for future research.

\paragraph{Language} Our benchmark is currently limited to the English language and primarily uses general-domain prose and conversations as the context haystack. The effectiveness of VTC, including achievable compression ratios and perceptual robustness, may vary for other languages, especially those with complex, non-alphabetic scripts (e.g., Chinese, Arabic) or different typographic conventions. Similarly, the performance on highly specialized or structured content, such as source code, legal documents, or scientific papers, remains unassessed and is a vital area for future expansion.

\paragraph{Scope of evaluation} While comprehensive in its inclusion of leading open-source and proprietary models, \ourbench~is necessarily a snapshot in a rapidly evolving field. The performance characteristics identified may not be fully generalizable to all VLM architectures, particularly those that may emerge with novel visual processing or attention mechanisms. Future work should involve the continuous integration and evaluation of new models to maintain the benchmark's relevance.

\paragraph{Constraints of API} Our evaluation of proprietary models is inherently constrained by the nature of their API-based access. Unlike open-source models, where we have full control over the inference pipeline, closed-source models operate as black boxes. Their APIs may perform opaque pre-processing steps, such as image resizing, compression, or normalization, before the data reaches the model. This lack of transparency means that the carefully rendered images we provide may be altered in ways that are beyond our control, potentially confounding our analysis of rendering parameters and their impact on performance. 
Besides, some APIs (e.g., Gemini-2.5-Pro) are constrained by the maximum number of input images, which will affect the performance of the model.

Despite these limitations, we believe VTCBench establishes a crucial foundation and a rigorous evaluation framework for advancing the development of more efficient and capable long-context VLMs.

\clearpage

\begin{table*}[p]
\centering
\caption{\ours-Retrieval S-NIAH Results.}
\label{tab:appdx:ruler:s}

\begin{tabular}{l|c|cccccc}
\toprule
Model          & $\compratio$  & 1k     & 2k     & 4k    & 8k    & 16k   & 32k   \\
\midrule
Qwen3-8B                & NA   & 98.18  & 100.00 & 96.36 & 97.73 & 95.45 & 94.55 \\
\midrule
InternVL3.5-8B         & 1.56 & 32.73  & 23.64  & 36.36 & 21.82 & 20.91 & 12.73 \\
InternVL3.5-38B        & 1.56 & 22.73  & 20.00  & 16.36 & 22.73 & 15.45 & 24.55 \\
Gemin-2.5-Pro           & 1.98 & 100.00 & 50.00  & 55.45 & 54.55 & 40.91 & 41.82 \\
Gemma3-27B              & 2.00 & 89.09  & 84.55  & 85.45 & 87.27 & \oom  & \oom  \\
GLM-4.1V-9B-Thinking    & 2.00 & 26.36  & 16.36  & 11.82 & 14.55 & 3.64  & 2.73  \\
Glyph                   & 2.00 & 92.73  & 83.64  & 90.00 & 83.64 & 94.55 & 85.45 \\
GPT-5                   & 2.00 & 76.36  & 76.36  & 71.82 & 61.82 & 62.73 & 60.00 \\
InternVL3.5-8B         & 2.00 & 19.09  & 22.73  & 10.91 & 13.64 & 12.73 & 10.00 \\
InternVL3.5-38B        & 2.00 & 17.27  & 16.36  & 12.73 & 13.64 & 9.09  & 11.82 \\
Kimi-VL-A3B-Instruct    & 2.00 & 98.18  & 67.27  & 59.09 & 55.45 & 61.82 & 66.36 \\
Qwen2.5-VL-7B-Instruct  & 2.00 & 84.55  & 77.27  & 82.73 & 84.55 & 78.18 & 84.55 \\
Qwen2.5-VL-72B-Instruct & 2.00 & 92.73  & 94.55  & 95.45 & 97.27 & 91.82 & 98.18 \\
Qwen3-VL-8B-Instruct    & 2.00 & 99.09  & 89.09  & 93.64 & 90.91 & 90.00 & 89.09 \\
Qwen3-VL-235B-A22B-Instruct& 2.00 & 97.27  & 82.73  & 92.73 & 88.18 & 89.09 & 74.55 \\
Qwen3-VL-8B-Instruct    & 2.55 & 97.27  & 72.73  & 75.45 & 74.55 & 56.36 & 59.09 \\
Qwen3-VL-235B-A22B-Instruct& 2.55 & 98.18  & 89.09  & 89.09 & 84.55 & 59.09 & 74.55 \\
Deepseek-OCR (Completion)& 3.12 & 51.82  & 50.91  & 50.00 & 16.36 & 1.82  & \oom  \\
Deepseek-OCR (QA)       & 3.12 & 30.91  & 32.73  & 25.45 & 0.00  & 0.00  & \oom  \\
GPT-5                   & 3.17 & 50.00  & 48.18  & 32.73 & 36.36 & 33.64 & 26.36 \\
Gemma3-27B              & 8.00 & 20.00  & 19.09  & 20.91 & 19.09 & 20.00 & 19.09 \\
\bottomrule
\end{tabular}
\end{table*}
\begin{table*}[p]
\centering
\caption{\ours-Retrieval MK-NIAH Results.}
\label{tab:appdx:ruler:mk}

\begin{tabular}{l|c|cccccc}
\toprule
Model          & $\compratio$  & 1k     & 2k     & 4k    & 8k    & 16k   & 32k   \\
\midrule
Qwen3-8B                & NA   & 98.18  & 97.27 & 90.91 & 92.73 & 91.82 & 95.45 \\
\midrule
InternVL3.5-8B         & 1.56 & 28.18  & 29.09 & 22.73 & 19.09 & 10.91 & 10.00 \\
InternVL3.5-38B        & 1.56 & 26.36  & 32.73 & 29.09 & 24.55 & 20.00 & 20.00 \\
Gemini-2.5-Pro          & 1.98 & 100.00 & 53.64 & 49.09 & 49.09 & 30.91 & 35.45 \\
Gemma3-27B              & 2.00 & 88.18  & 91.82 & 88.18 & 81.82 & \oom  & \oom  \\
GLM-4.1V-9B-Thinking    & 2.00 & 10.00  & 13.64 & 6.36  & 10.91 & 8.18  & 3.64  \\
Glyph                   & 2.00 & 92.73  & 81.82 & 83.64 & 81.82 & 82.73 & 78.18 \\
GPT-5                   & 2.00 & 85.45  & 71.82 & 68.18 & 61.36 & 59.09 & 49.09 \\
InternVL3.5-8B         & 2.00 & 12.73  & 19.09 & 7.27  & 11.82 & 6.36  & 8.18  \\
InternVL3.5-38B        & 2.00 & 27.27  & 16.36 & 12.73 & 15.45 & 14.55 & 12.73 \\
Kimi-VL-A3B-Instruct    & 2.00 & 84.55  & 68.18 & 48.18 & 44.55 & 47.27 & 44.55 \\
Qwen2.5-VL-7B-Instruct  & 2.00 & 93.64  & 80.00 & 81.82 & 79.09 & 88.18 & 76.36 \\
Qwen2.5-VL-72B-Instruct & 2.00 & 89.09  & 80.00 & 87.27 & 90.91 & 86.36 & 80.91 \\
Qwen3-VL-8B-Instruct    & 2.00 & 93.64  & 84.55 & 85.45 & 90.00 & 90.91 & 85.86 \\
Qwen3-VL-235B-A22B-Instruct& 2.00 & 100.00 & 92.73 & 99.09 & 96.36 & 92.12 & 85.18 \\
Qwen3-VL-8B-Instruct    & 2.55 & 90.91  & 78.18 & 69.09 & 69.09 & 55.45 & 66.36 \\
Qwen3-VL-235B-A22B-Instruct& 2.55 & 100.00 & 84.55 & 89.09 & 83.64 & 62.73 & 52.73 \\
Deepseek-OCR (Completion)& 3.12 & 40.00  & 50.91 & 51.82 & 12.73 & 0.91  & \oom  \\
Deepseek-OCR (QA)       & 3.12 & 23.64  & 26.36 & 17.27 & 0.00  & 0.00  & \oom  \\
GPT-5                   & 3.17 & 44.55  & 42.73 & 34.55 & 32.73 & 28.18 & 21.82 \\
Gemma3-27B              & 8.00 & 28.18  & 26.36 & 32.73 & 22.73 & 10.91 & 15.45 \\
\bottomrule
\end{tabular}
\end{table*}
\begin{table*}[p]
\centering
\caption{\ours-Retrieval MV-NIAH Results.}
\label{tab:appdx:ruler:mv}

\begin{tabular}{l|c|cccccc}
\toprule
Model            & $\compratio$  & 1k     & 2k     & 4k    & 8k    & 16k   & 32k   \\
\midrule
Qwen3-8B                & NA   & 99.09  & 100.00 & 100.00 & 100.00 & 99.09 & 96.82 \\
\midrule
InternVL3.5-8B         & 1.56 & 11.82  & 15.45  & 16.82  & 21.82  & 15.45 & 15.45 \\
InternVL3.5-38B        & 1.56 & 17.27  & 15.91  & 19.55  & 20.45  & 18.64 & 18.64 \\
Gemini-2.5-Pro          & 1.98 & 100.00 & 50.00  & 55.45  & 54.55  & 40.00 & 41.82 \\
Gemma3-27B              & 2.00 & 90.91  & 83.64  & 83.64  & 67.27  & \oom  & \oom  \\
GLM-4.1V-9B-Thinking    & 2.00 & 17.73  & 9.09   & 5.00   & 4.09   & 0.91  & 2.27  \\
Glyph                   & 2.00 & 96.36  & 85.45  & 86.36  & 76.82  & 76.82 & 75.45 \\
GPT-5                   & 2.00 & 84.09  & 75.00  & 72.27  & 64.55  & 62.73 & 61.82 \\
InternVL3.5-8B         & 2.00 & 12.27  & 11.36  & 10.45  & 8.18   & 8.18  & 8.64  \\
InternVL3.5-38B        & 2.00 & 20.45  & 14.09  & 11.82  & 14.55  & 7.73  & 8.18  \\
Kimi-VL-A3B-Instruct    & 2.00 & 77.73  & 41.82  & 40.00  & 33.64  & 32.27 & 28.64 \\
Qwen2.5-VL-7B-Instruct  & 2.00 & 80.45  & 69.55  & 70.45  & 76.36  & 69.55 & 62.73 \\
Qwen2.5-VL-72B-Instruct & 2.00 & 83.64  & 80.45  & 83.64  & 73.64  & 75.45 & 59.55 \\
Qwen3-VL-8B-Instruct    & 2.00 & 93.18  & 90.00  & 88.18  & 85.45  & 82.27 & 69.55 \\
Qwen3-VL-235B-A22B-Instruct& 2.00 & 96.36  & 86.36  & 92.27  & 93.18  & 86.82 & 76.68 \\
Qwen3-VL-8B-Instruct    & 2.55 & 93.64  & 62.73  & 55.91  & 46.36  & 38.64 & 38.64 \\
Qwen3-VL-235B-A22B-Instruct& 2.55 & 96.82  & 81.82  & 74.55  & 70.91  & 56.82 & 44.55 \\
Deepseek-OCR (Completion)& 3.12 & 20.91  & 24.09  & 17.73  & 14.55  & 1.36  & \oom  \\
Deepseek-OCR (QA)       & 3.12 & 10.45  & 16.82  & 9.09   & 1.36   & 0.00  & \oom  \\
GPT-5                   & 3.17 & 41.36  & 32.27  & 32.73  & 34.55  & 30.45 & 29.09 \\
Gemma3-27B              & 8.00 & 23.18  & 14.55  & 20.45  & 17.27  & 13.18 & 9.55  \\
\bottomrule
\end{tabular}
\end{table*}
\begin{table*}[p]
\centering
\caption{\ours-Retrieval MQ-NIAH Results.}
\label{tab:appdx:ruler:mq}

\begin{tabular}{l|c|cccccc}
\toprule
Model          & $\compratio$  & 1k     & 2k     & 4k    & 8k    & 16k   & 32k   \\
\midrule
Qwen3-8B                & NA   & 100.00 & 98.18 & 100.00 & 98.18 & 100.00 & 95.45 \\
\midrule
InternVL3.5-8B         & 1.56 & 21.82  & 19.09 & 18.64  & 20.45 & 16.82  & 11.82 \\
InternVL3.5-38B        & 1.56 & 20.00  & 16.36 & 22.27  & 17.27 & 16.82  & 12.73 \\
Gemini-2.5-Pro          & 1.98 & 100.00 & 60.91 & 58.18  & 60.00 & 42.73  & 43.18 \\
Gemma3-27B              & 2.00 & 87.27  & 79.55 & 71.82  & 61.36 & \oom   & \oom  \\
GLM-4.1V-9B-Thinking    & 2.00 & 14.55  & 6.82  & 7.73   & 5.00  & 2.73   & 1.36  \\
Glyph                   & 2.00 & 84.09  & 74.55 & 79.55  & 71.82 & 67.73  & 63.64 \\
GPT-5                   & 2.00 & 81.82  & 75.00 & 72.27  & 63.18 & 58.18  & 57.73 \\
InternVL3.5-8B         & 2.00 & 15.91  & 13.18 & 10.91  & 13.18 & 6.82   & 5.00  \\
InternVL3.5-38B        & 2.00 & 15.45  & 11.82 & 10.00  & 8.18  & 7.73   & 6.82  \\
Kimi-VL-A3B-Instruct    & 2.00 & 87.73  & 53.18 & 48.64  & 43.18 & 35.91  & 32.27 \\
Qwen2.5-VL-7B-Instruct  & 2.00 & 82.27  & 75.45 & 64.09  & 67.27 & 70.45  & 53.18 \\
Qwen2.5-VL-72B-Instruct & 2.00 & 90.45  & 83.18 & 84.55  & 84.55 & 75.45  & 67.27 \\
Qwen3-VL-8B-Instruct    & 2.00 & 95.00  & 90.00 & 82.27  & 89.55 & 84.55  & 73.86 \\
Qwen3-VL-235B-A22B-Instruct& 2.00 & 95.00  & 92.27 & 95.00  & 90.91 & 96.82  & 88.96 \\
Qwen3-VL-8B-Instruct    & 2.55 & 95.00  & 75.91 & 65.45  & 55.91 & 47.27  & 49.09 \\
Qwen3-VL-235B-A22B-Instruct& 2.55 & 95.91  & 92.27 & 89.55  & 80.45 & 75.91  & 62.34 \\
Deepseek-OCR (Completion)& 3.12 & 21.82  & 18.18 & 19.09  & 6.82  & 0.00   & \oom  \\
Deepseek-OCR (QA)       & 3.12 & 7.73   & 11.36 & 6.82   & 0.00  & 0.00   & \oom  \\
GPT-5                   & 3.17 & 46.36  & 31.36 & 28.18  & 28.18 & 24.09  & 24.09 \\
Gemma3-27B              & 8.00 & 16.82  & 11.82 & 15.91  & 15.00 & 13.64  & 11.36 \\
\bottomrule
\end{tabular}
\end{table*}

\begin{figure*}[p]
\centering

\begin{subfigure}[b]{.32\linewidth}
    \includegraphics[width=\textwidth]{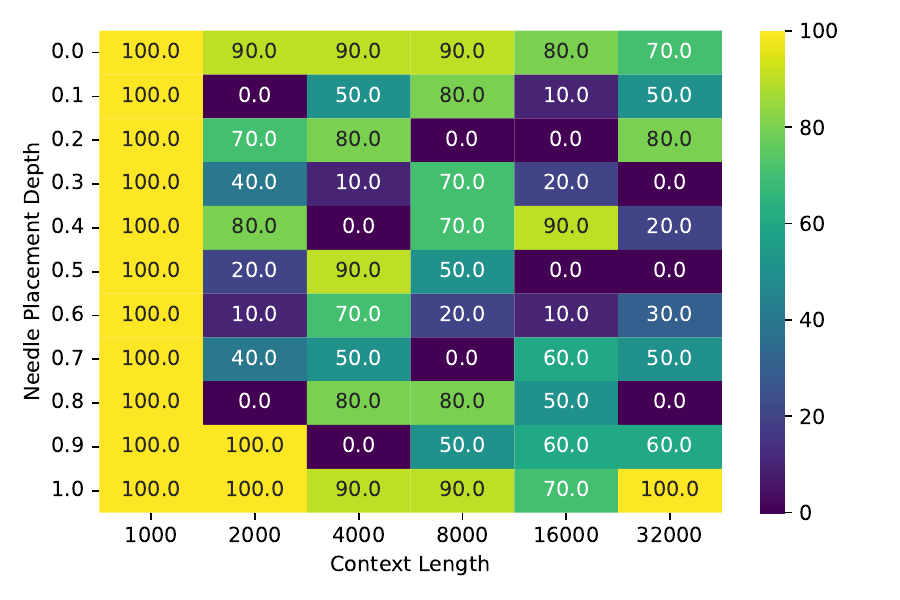}
    \caption{Gemini-2.5-Pro}
\end{subfigure}
\begin{subfigure}[b]{.32\linewidth}
    \includegraphics[width=\textwidth]{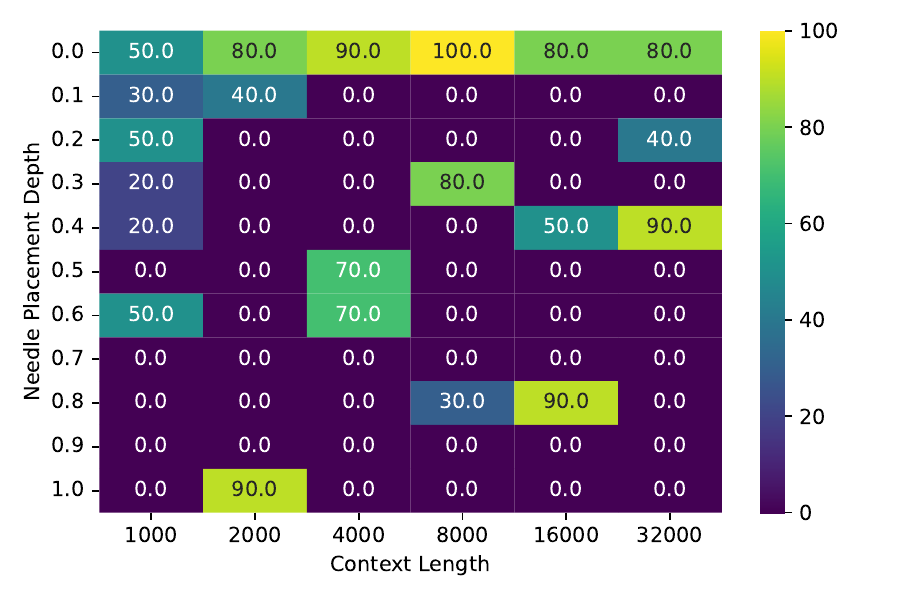}
    \caption{Gemma3-27B}
\end{subfigure}
\begin{subfigure}[b]{.32\linewidth}
    \includegraphics[width=\textwidth]{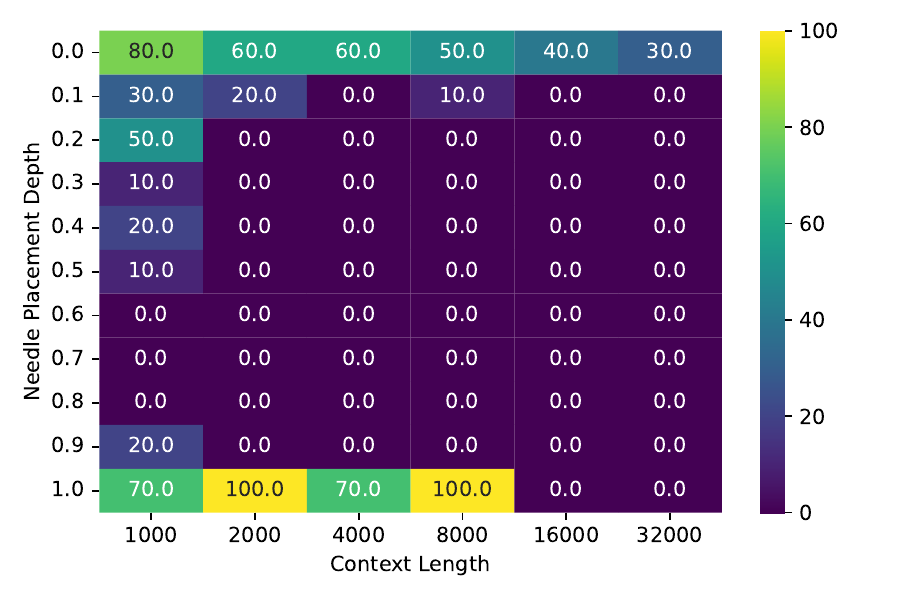}
    \caption{GLM-4.1V-9B-Thinking}
\end{subfigure}
\begin{subfigure}[b]{.32\linewidth}
    \includegraphics[width=\textwidth]{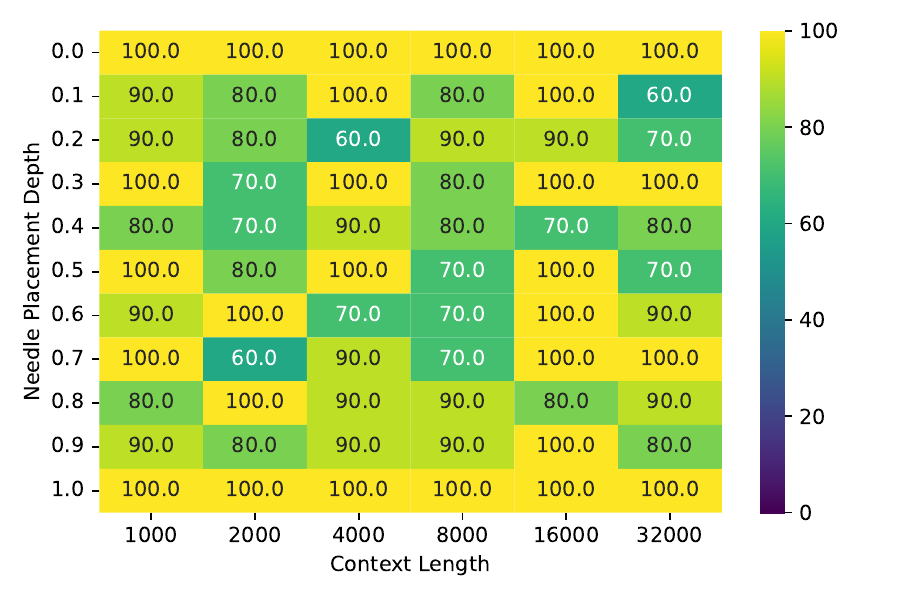}
    \caption{Glyph}\label{fig:gly_used}
\end{subfigure}
\begin{subfigure}[b]{.32\linewidth}
    \includegraphics[width=\textwidth]{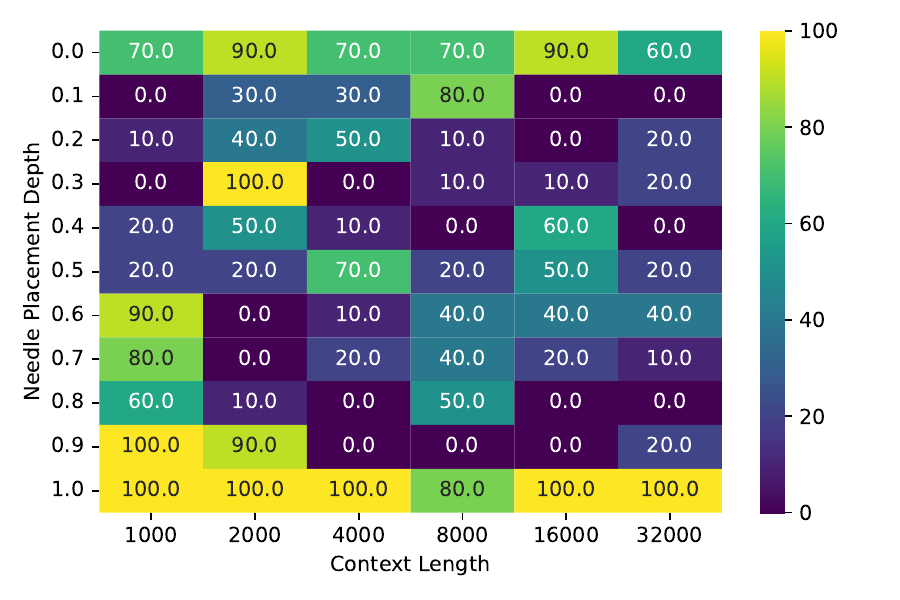}
    \caption{GPT-5}
\end{subfigure}
\begin{subfigure}[b]{.32\linewidth}
    \includegraphics[width=\textwidth]{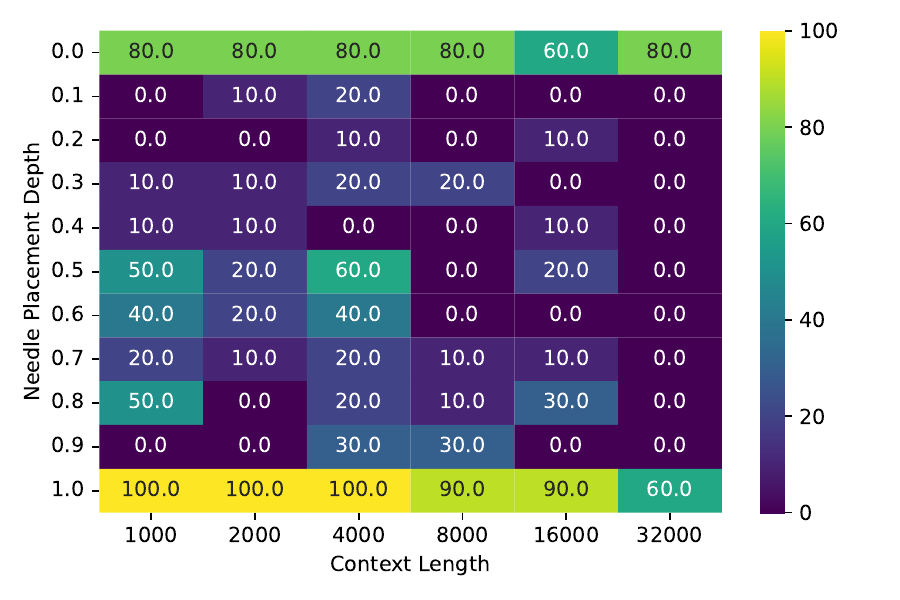}
    \caption{InternVL3.5-8B}
\end{subfigure}
\begin{subfigure}[b]{.32\linewidth}
    \includegraphics[width=\textwidth]{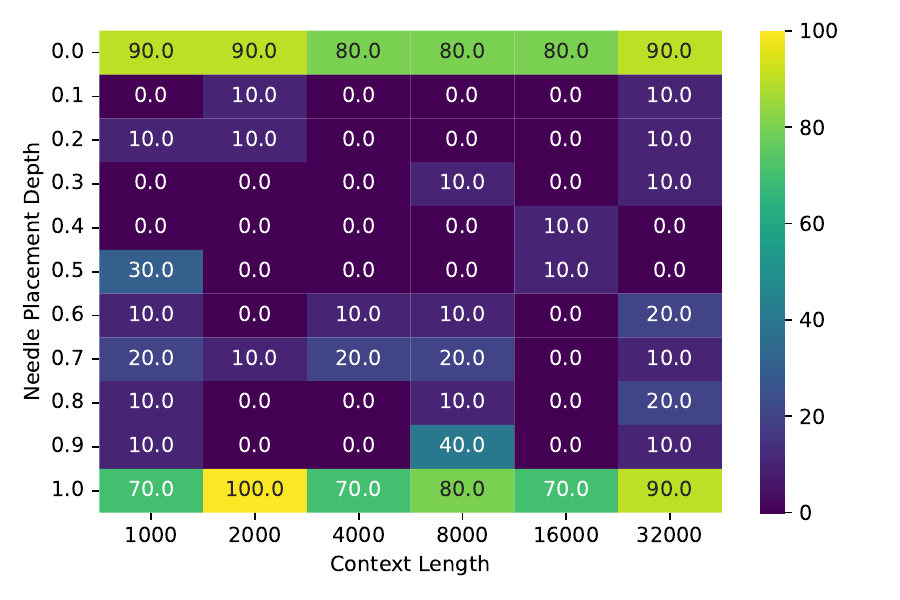}
    \caption{InternVL3.5-38B}
\end{subfigure}
\begin{subfigure}[b]{.32\linewidth}
    \includegraphics[width=\textwidth]{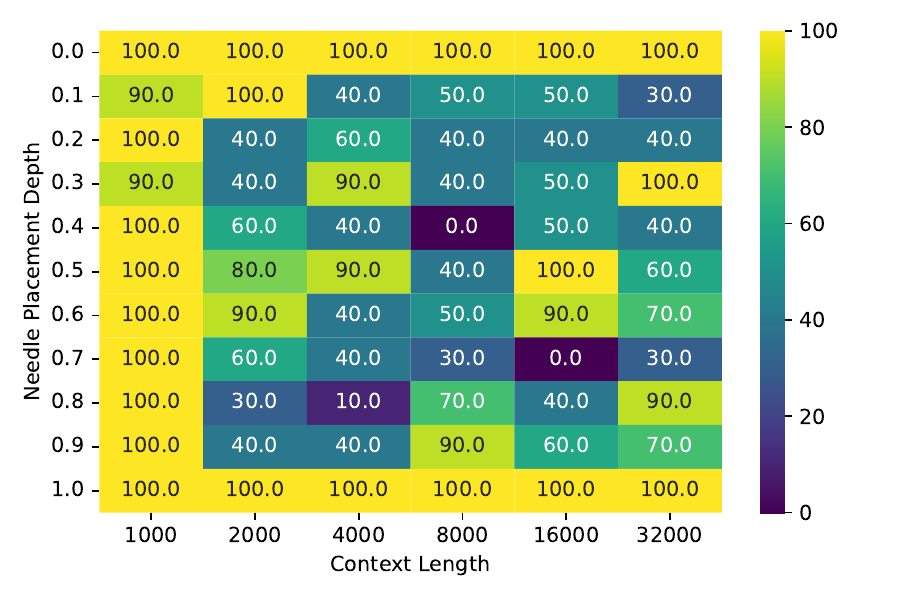}
    \caption{Kimi-VL-A3B-Instruct}
\end{subfigure}
\begin{subfigure}[b]{.32\linewidth}
    \includegraphics[width=\textwidth]{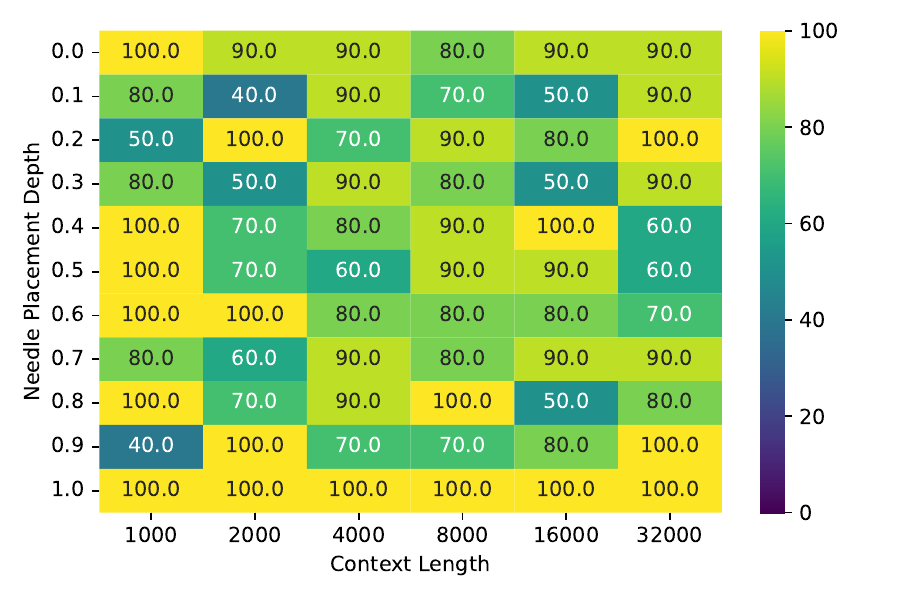}
    \caption{Qwen2.5-VL-7B-Instruct}
\end{subfigure}
\begin{subfigure}[b]{.32\linewidth}
    \includegraphics[width=\textwidth]{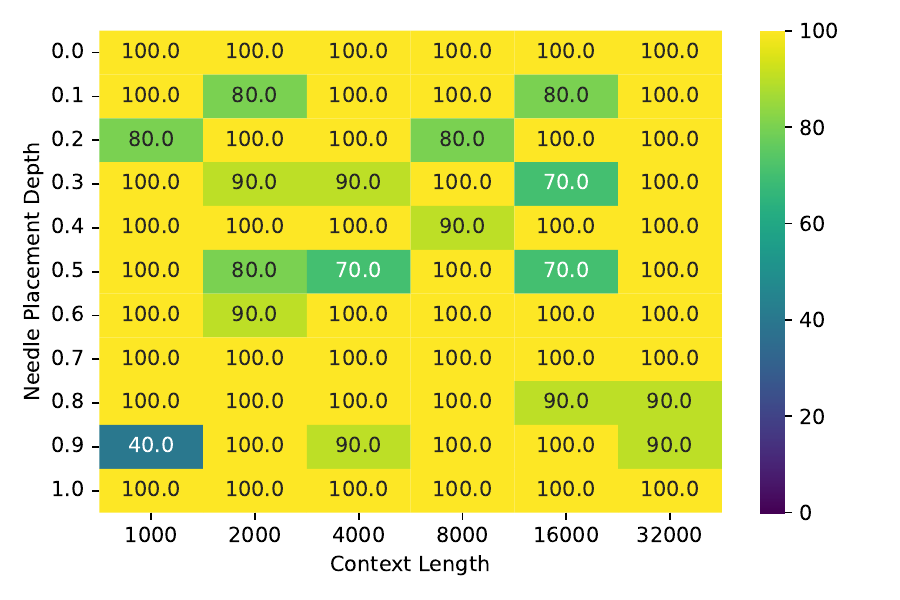}
    \caption{Qwen2.5-VL-72B-Instruct}\label{fig:qwen25_72}
\end{subfigure}
\begin{subfigure}[b]{.32\linewidth}
    \includegraphics[width=\textwidth]{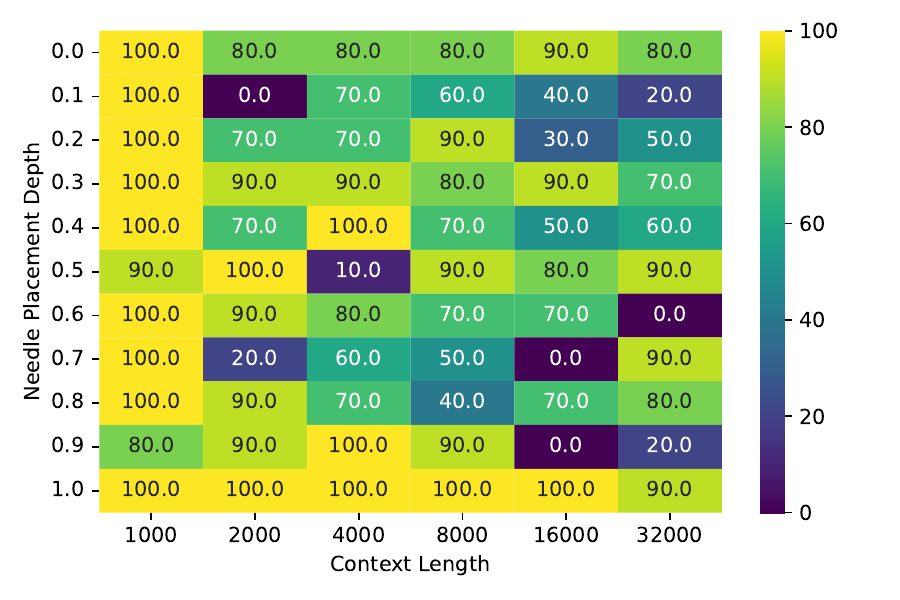}
    \caption{Qwen3-VL-8B-Instruct}
\end{subfigure}
\begin{subfigure}[b]{.32\linewidth}
    \includegraphics[width=\textwidth]{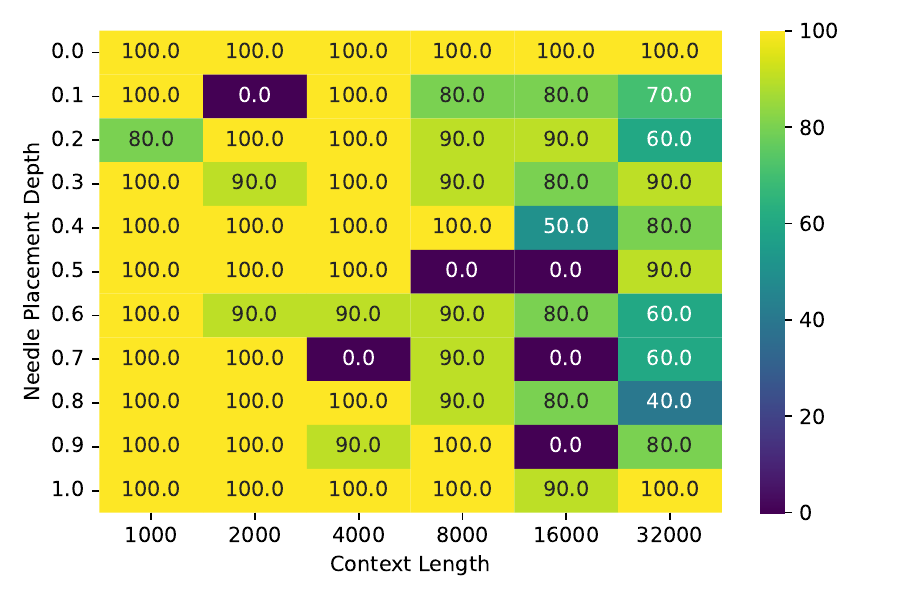}
    \caption{Qwen3-VL-235B-A22B-Instruct}
\end{subfigure}

\caption{\ours-Retrieval S-NIAH (preset $\render=\renderdefault$) accuracy w.r.t. needle placement depth. Blank cells indicate \oom.}
\label{fig:appdx:ruler:s}
\end{figure*}
\begin{figure*}[p]
\centering

\begin{subfigure}[b]{.32\linewidth}
    \includegraphics[width=\textwidth]{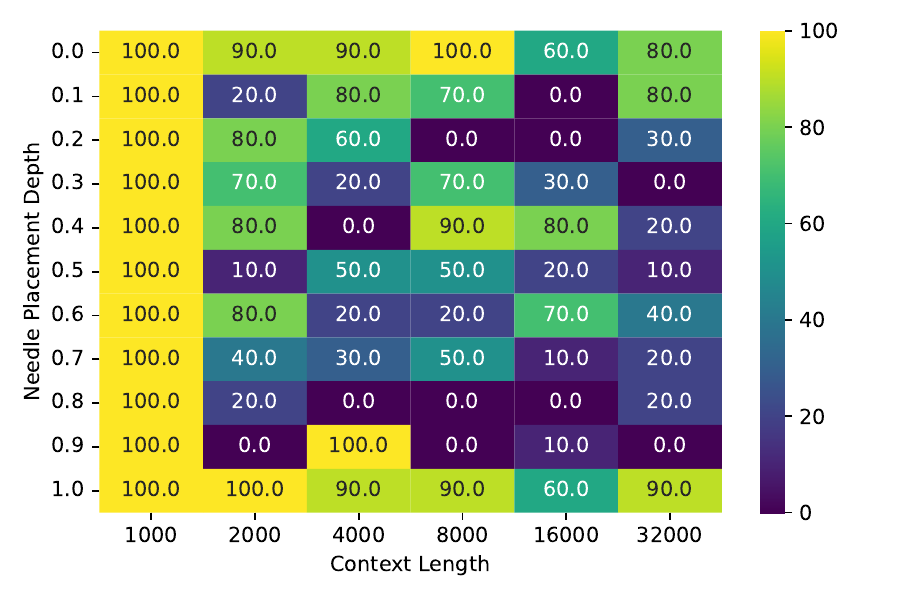}
    \caption{Gemini-2.5-Pro}
\end{subfigure}
\begin{subfigure}[b]{.32\linewidth}
    \includegraphics[width=\textwidth]{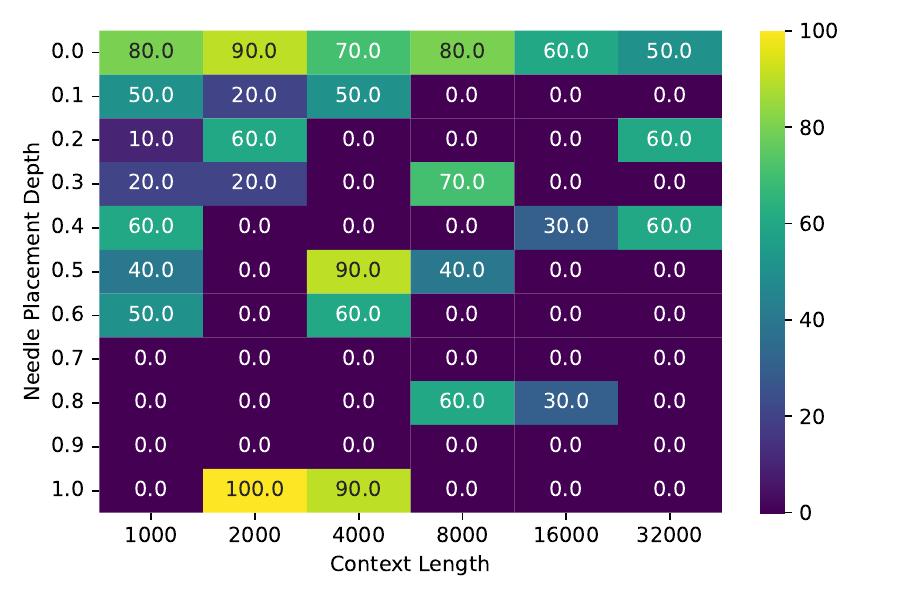}
    \caption{Gemma3-27B}
\end{subfigure}
\begin{subfigure}[b]{.32\linewidth}
    \includegraphics[width=\textwidth]{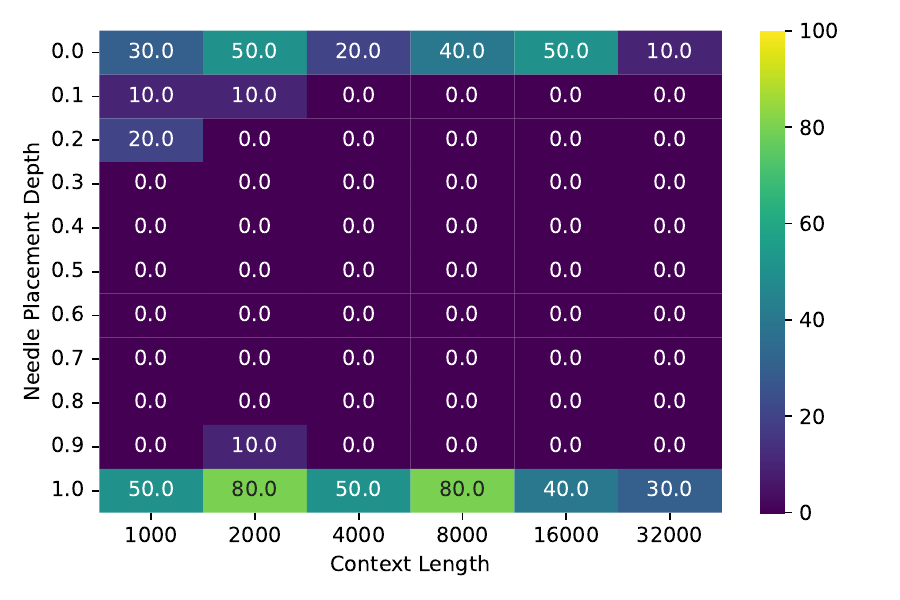}
    \caption{GLM-4.1V-9B-Thinking}
\end{subfigure}
\begin{subfigure}[b]{.32\linewidth}
    \includegraphics[width=\textwidth]{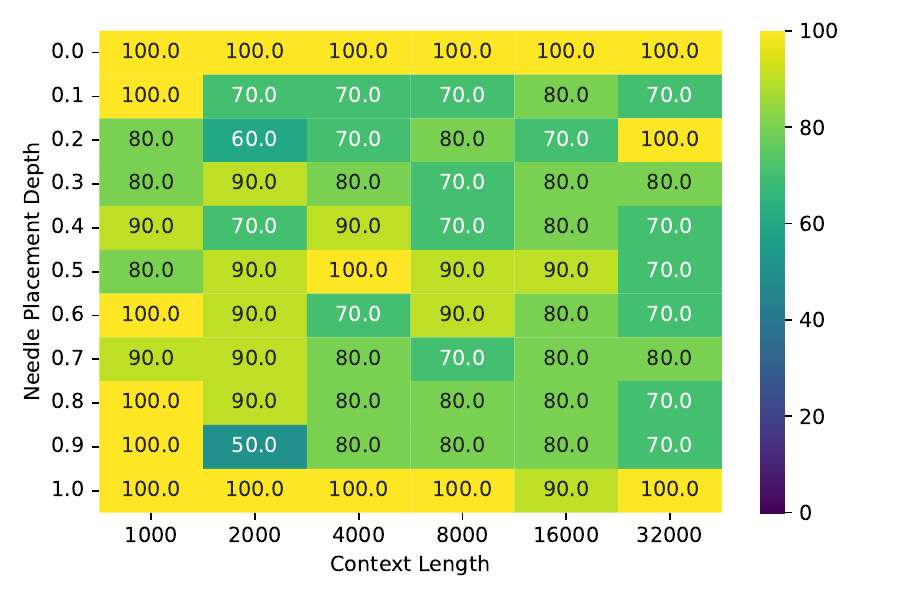}
    \caption{Glyph}
\end{subfigure}
\begin{subfigure}[b]{.32\linewidth}
    \includegraphics[width=\textwidth]{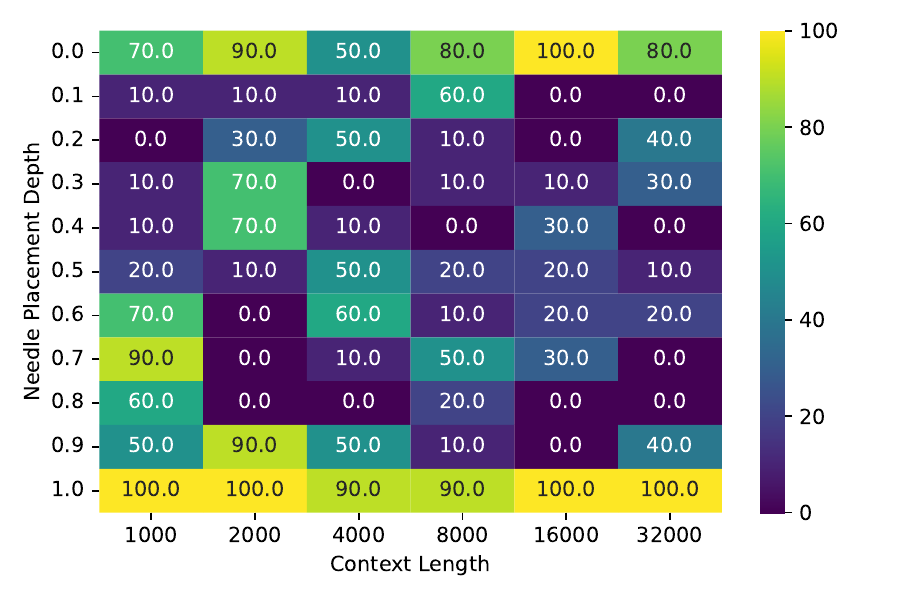}
    \caption{GPT-5}
\end{subfigure}
\begin{subfigure}[b]{.32\linewidth}
    \includegraphics[width=\textwidth]{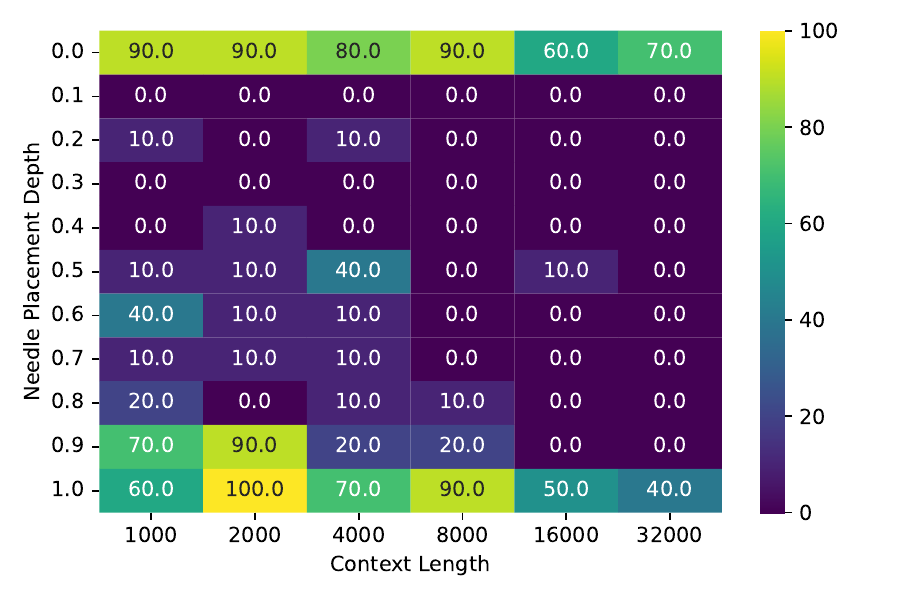}
    \caption{InternVL3.5-8B}
\end{subfigure}
\begin{subfigure}[b]{.32\linewidth}
    \includegraphics[width=\textwidth]{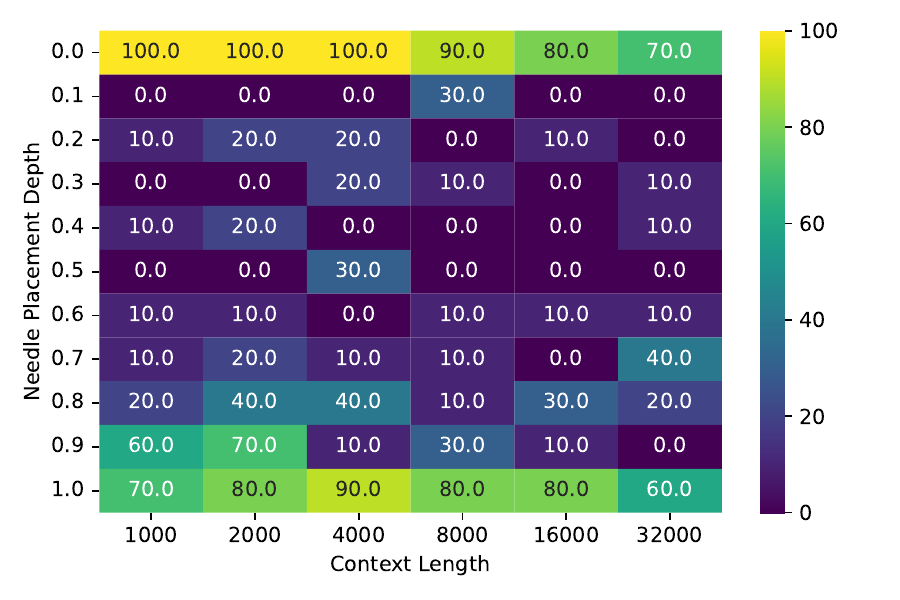}
    \caption{InternVL3.5-38B}
\end{subfigure}
\begin{subfigure}[b]{.32\linewidth}
    \includegraphics[width=\textwidth]{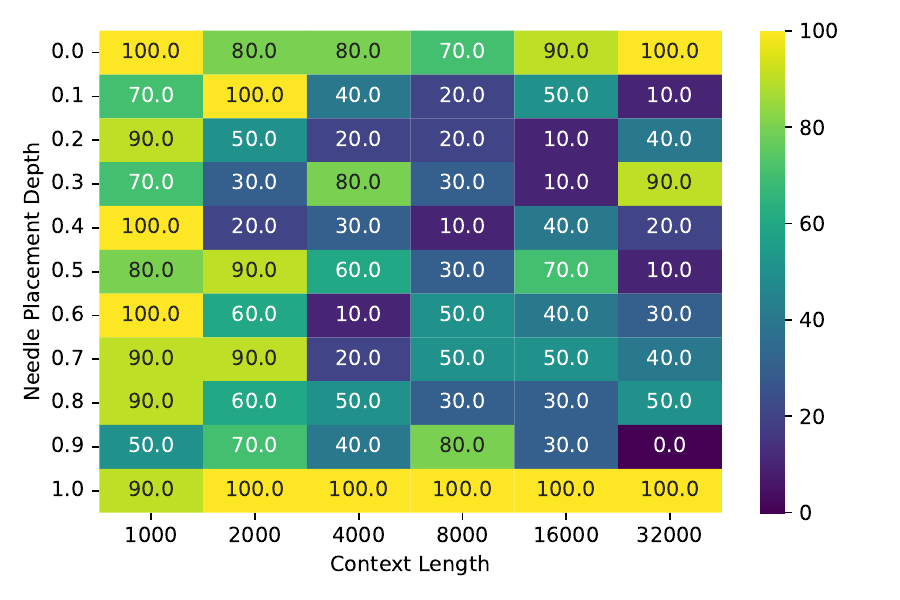}
    \caption{Kimi-VL-A3B-Instruct}
\end{subfigure}
\begin{subfigure}[b]{.32\linewidth}
    \includegraphics[width=\textwidth]{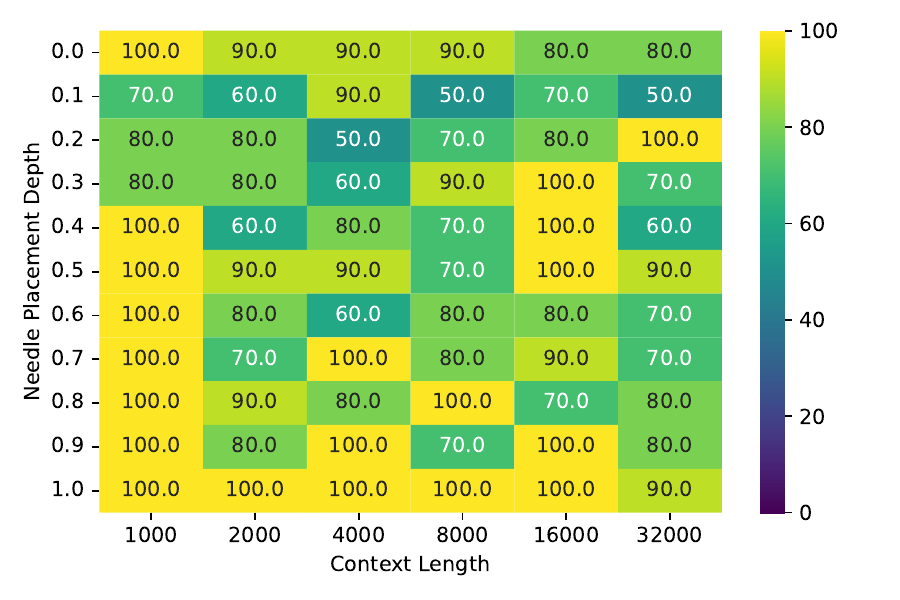}
    \caption{Qwen2.5-VL-7B-Instruct}
\end{subfigure}
\begin{subfigure}[b]{.32\linewidth}
    \includegraphics[width=\textwidth]{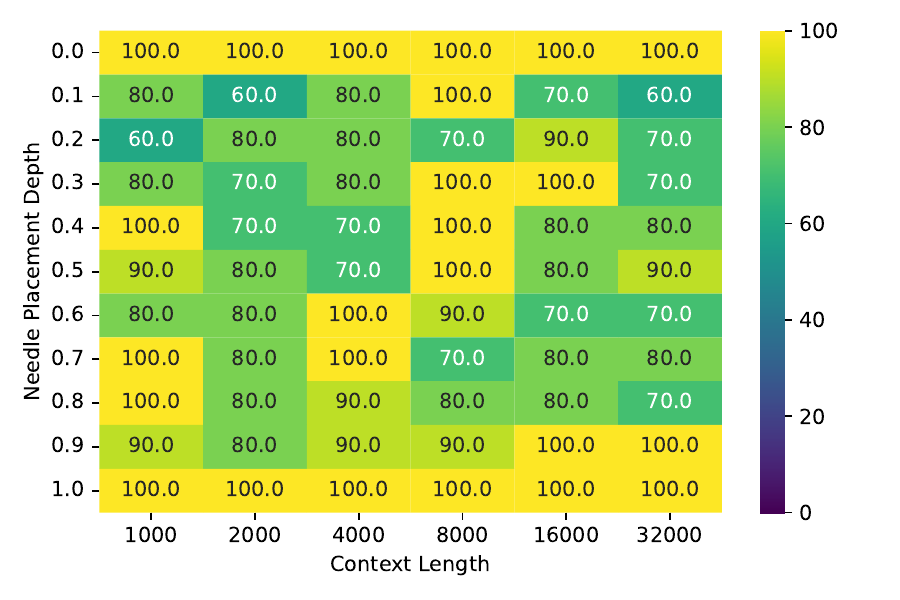}
    \caption{Qwen2.5-VL-72B-Instruct}
\end{subfigure}
\begin{subfigure}[b]{.32\linewidth}
    \includegraphics[width=\textwidth]{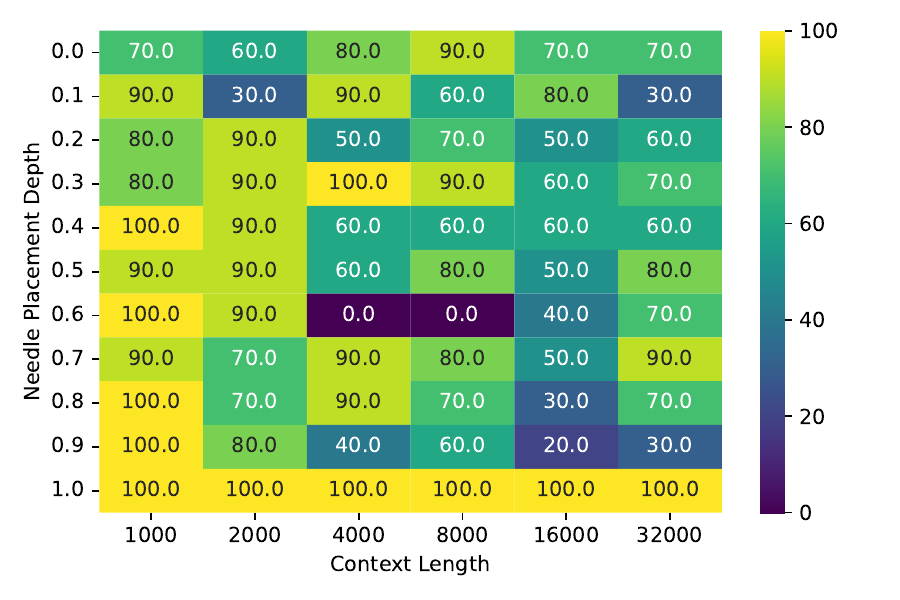}
    \caption{Qwen3-VL-8B-Instruct}
\end{subfigure}
\begin{subfigure}[b]{.32\linewidth}
    \includegraphics[width=\textwidth]{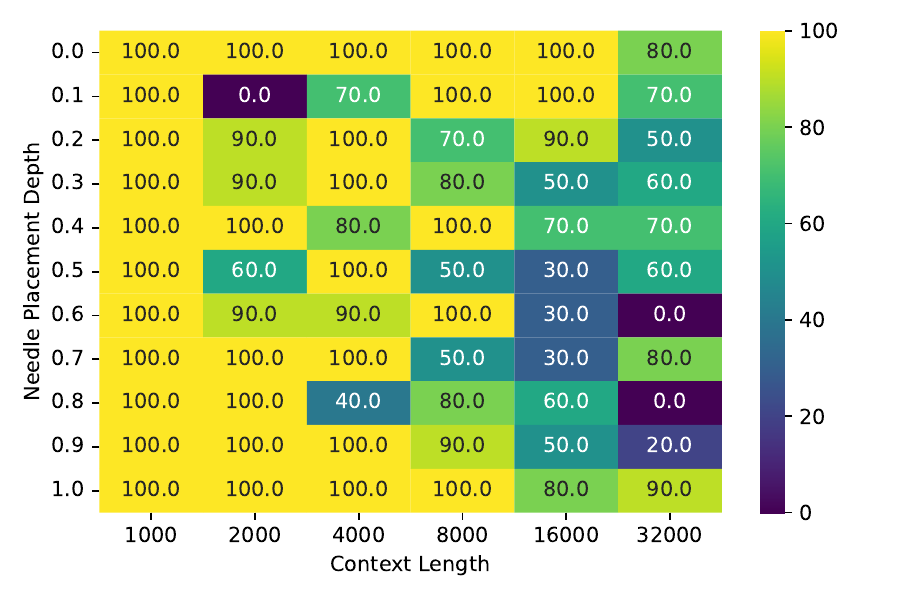}
    \caption{Qwen3-VL-235B-A22B-Instruct}
\end{subfigure}

\caption{\ours-Retrieval MK-NIAH (preset $\render=\renderdefault$) accuracy w.r.t. needle placement depth. Blank cells indicate \oom.}
\label{fig:appdx:ruler:mk}
\end{figure*}
\begin{figure*}[p]
\centering

\begin{subfigure}[b]{.32\linewidth}
    \includegraphics[width=\textwidth]{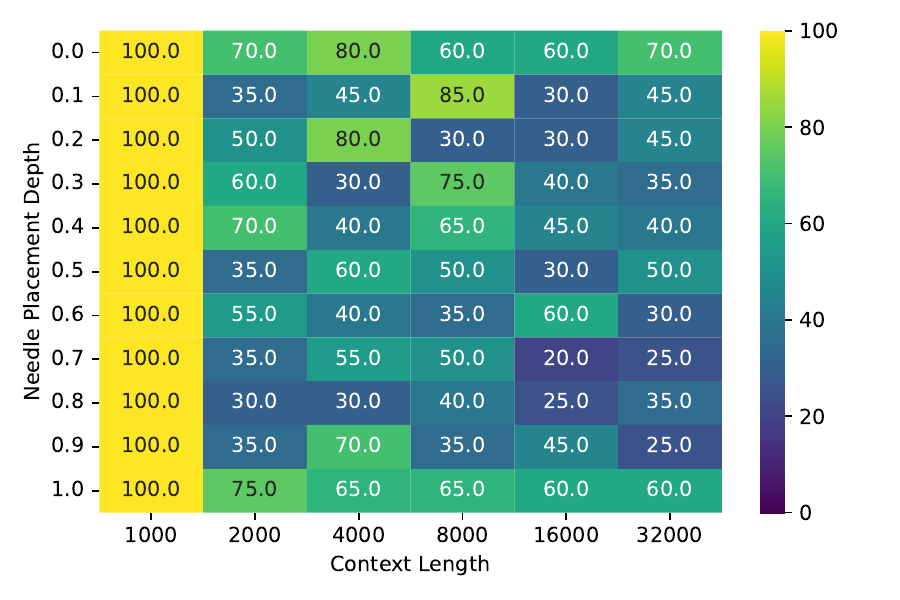}
    \caption{Gemini-2.5-Pro}
\end{subfigure}
\begin{subfigure}[b]{.32\linewidth}
    \includegraphics[width=\textwidth]{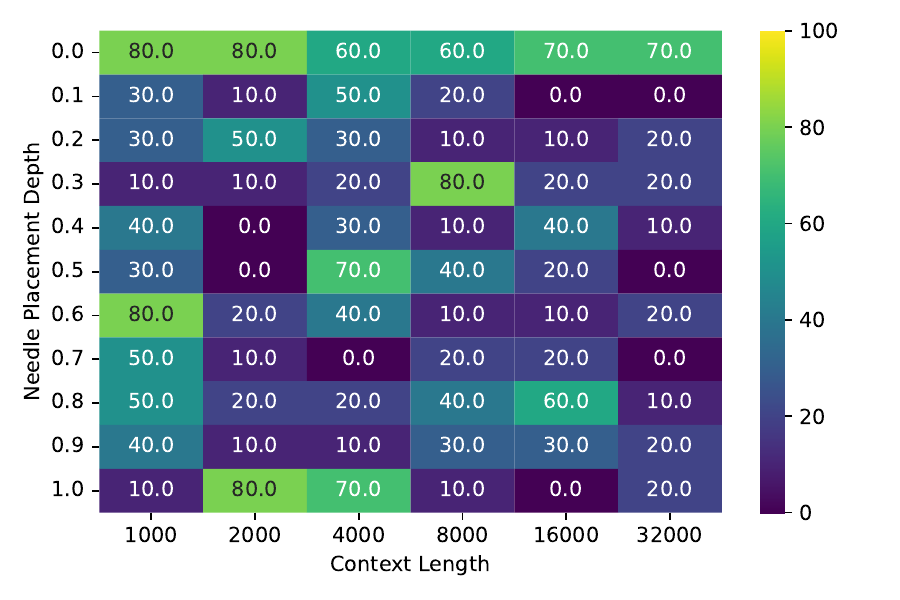}
    \caption{Gemma3-27B}
\end{subfigure}
\begin{subfigure}[b]{.32\linewidth}
    \includegraphics[width=\textwidth]{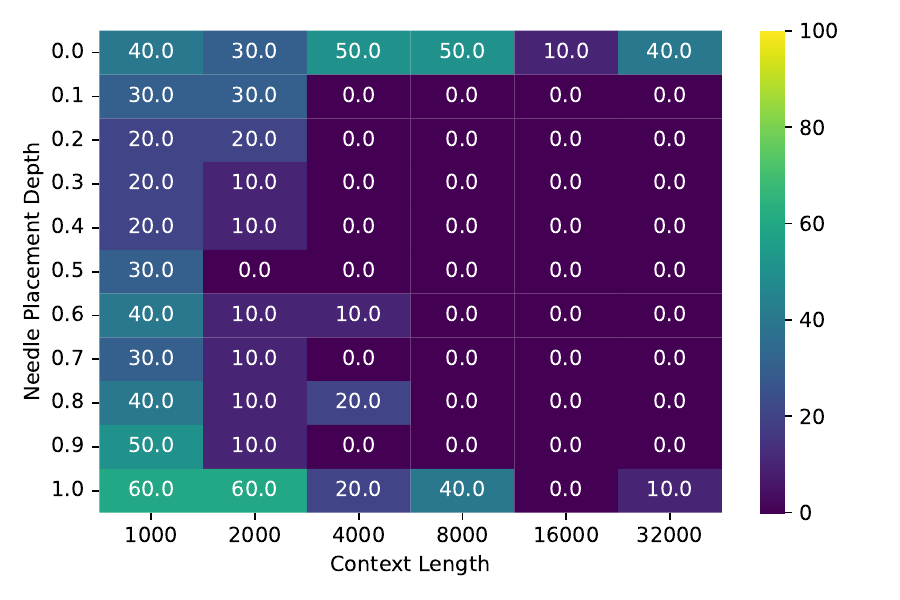}
    \caption{GLM-4.1V-9B-Thinking}
\end{subfigure}
\begin{subfigure}[b]{.32\linewidth}
    \includegraphics[width=\textwidth]{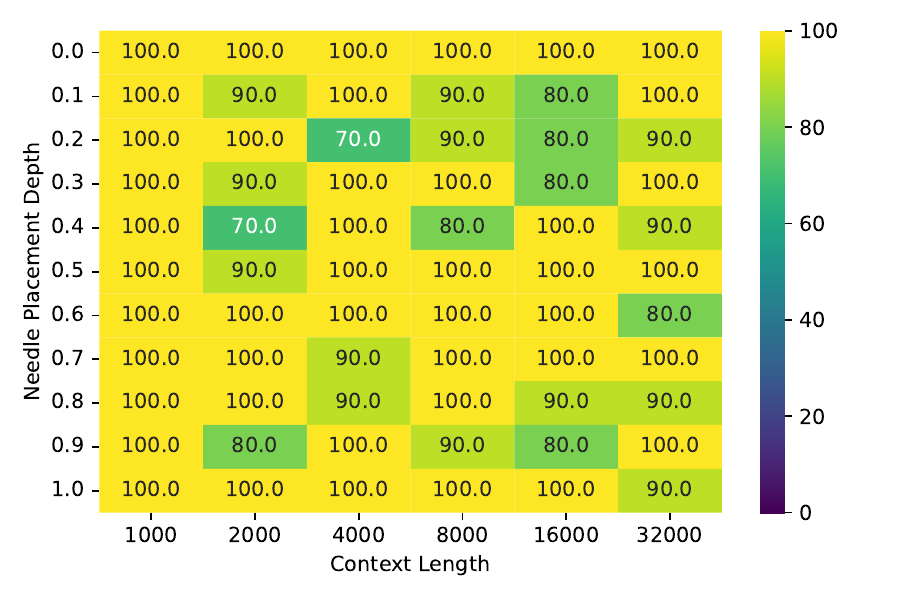}
    \caption{Glyph}
\end{subfigure}
\begin{subfigure}[b]{.32\linewidth}
    \includegraphics[width=\textwidth]{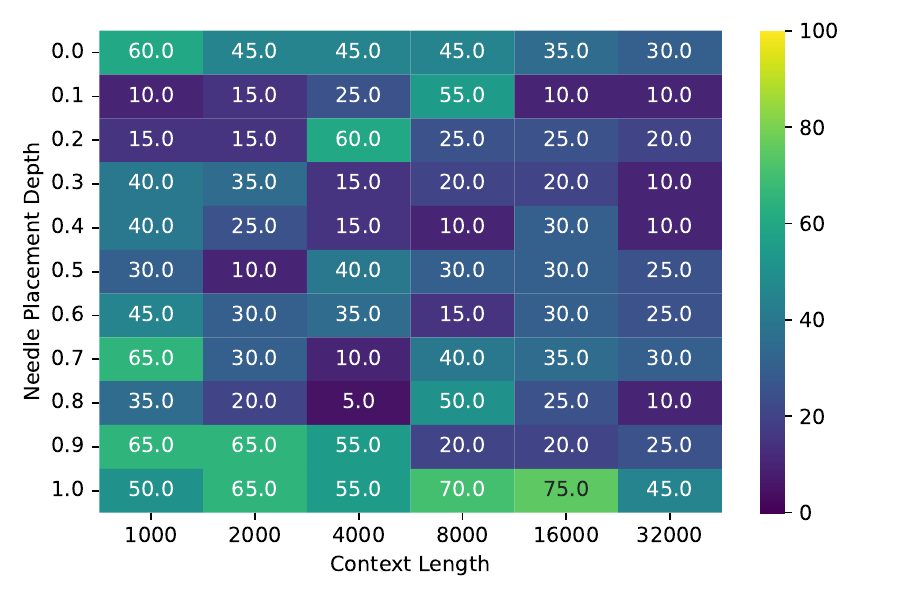}
    \caption{GPT-5}
\end{subfigure}
\begin{subfigure}[b]{.32\linewidth}
    \includegraphics[width=\textwidth]{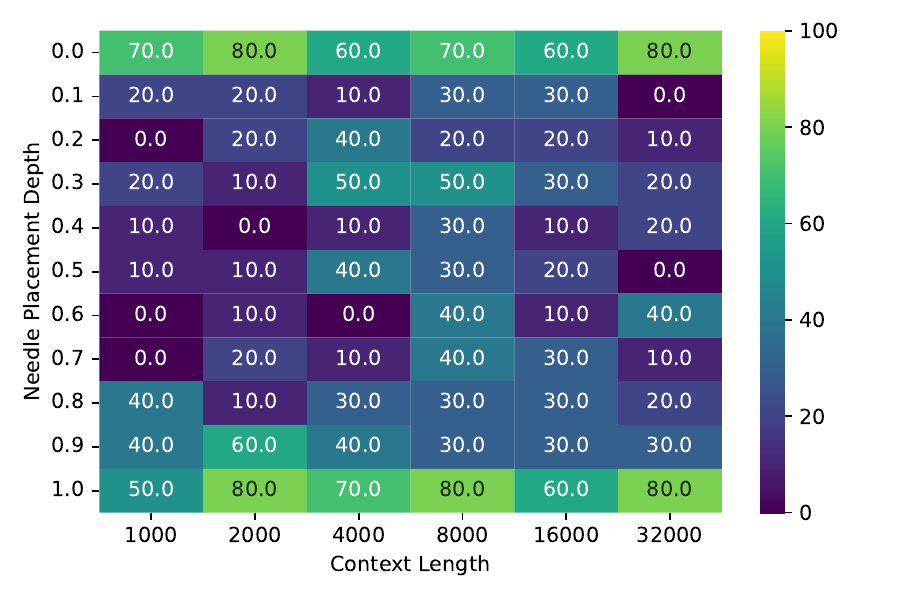}
    \caption{InternVL3.5-8B}
\end{subfigure}
\begin{subfigure}[b]{.32\linewidth}
    \includegraphics[width=\textwidth]{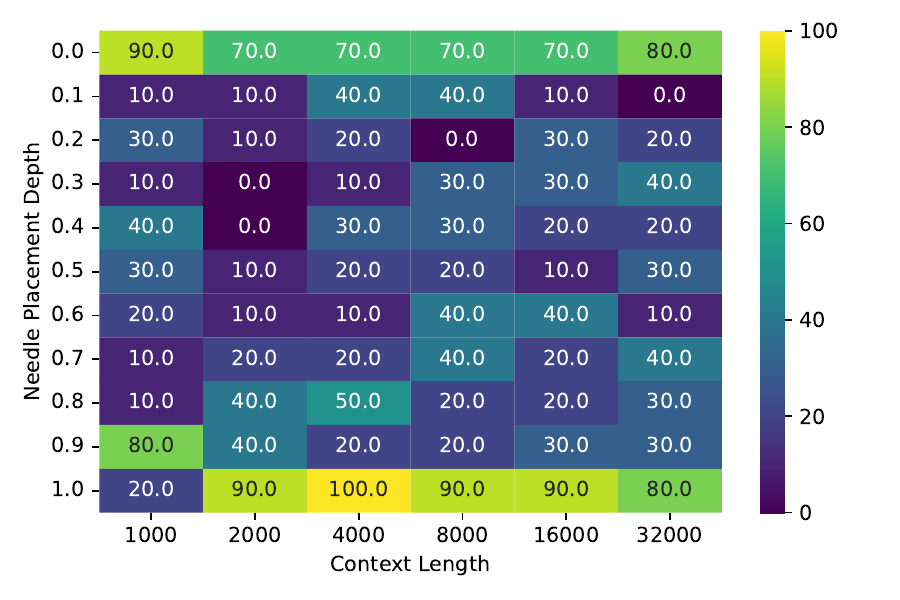}
    \caption{InternVL3.5-38B}
\end{subfigure}
\begin{subfigure}[b]{.32\linewidth}
    \includegraphics[width=\textwidth]{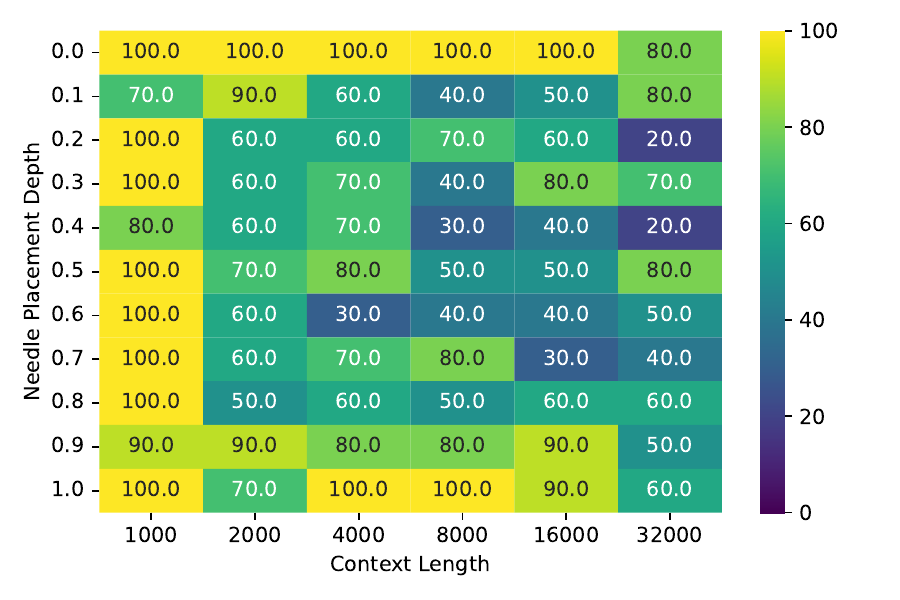}
    \caption{Kimi-VL-A3B-Instruct}
\end{subfigure}
\begin{subfigure}[b]{.32\linewidth}
    \includegraphics[width=\textwidth]{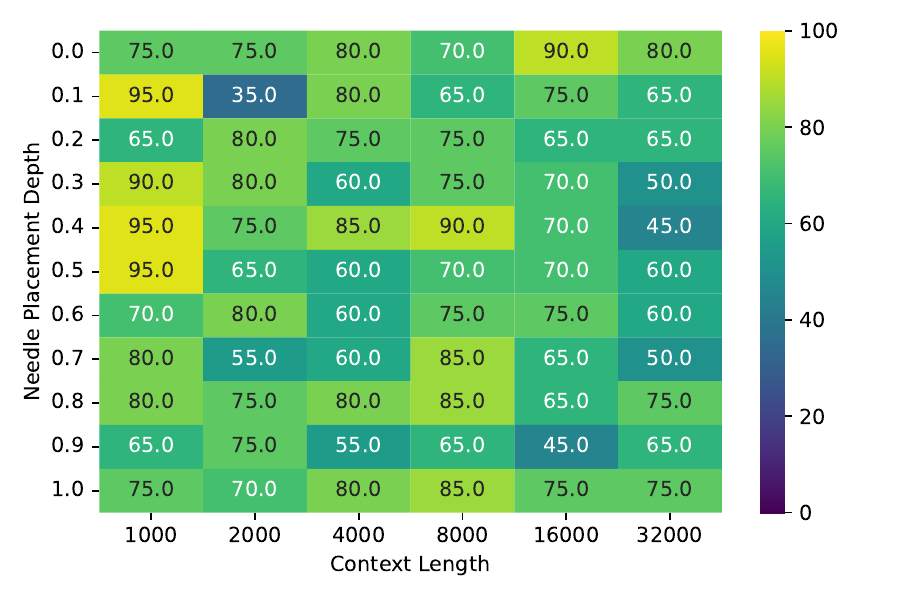}
    \caption{Qwen2.5-VL-7B-Instruct}
\end{subfigure}
\begin{subfigure}[b]{.32\linewidth}
    \includegraphics[width=\textwidth]{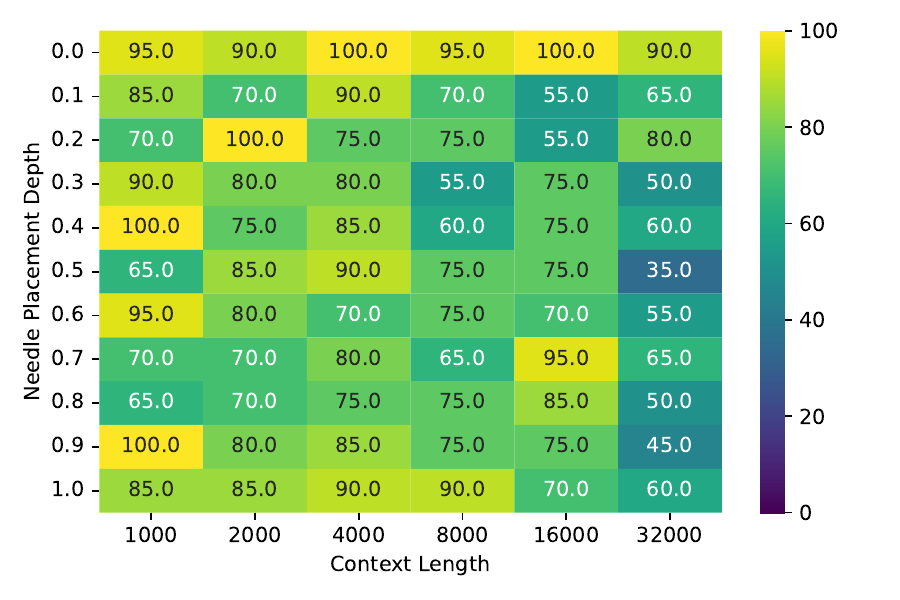}
    \caption{Qwen2.5-VL-72B-Instruct}
\end{subfigure}
\begin{subfigure}[b]{.32\linewidth}
    \includegraphics[width=\textwidth]{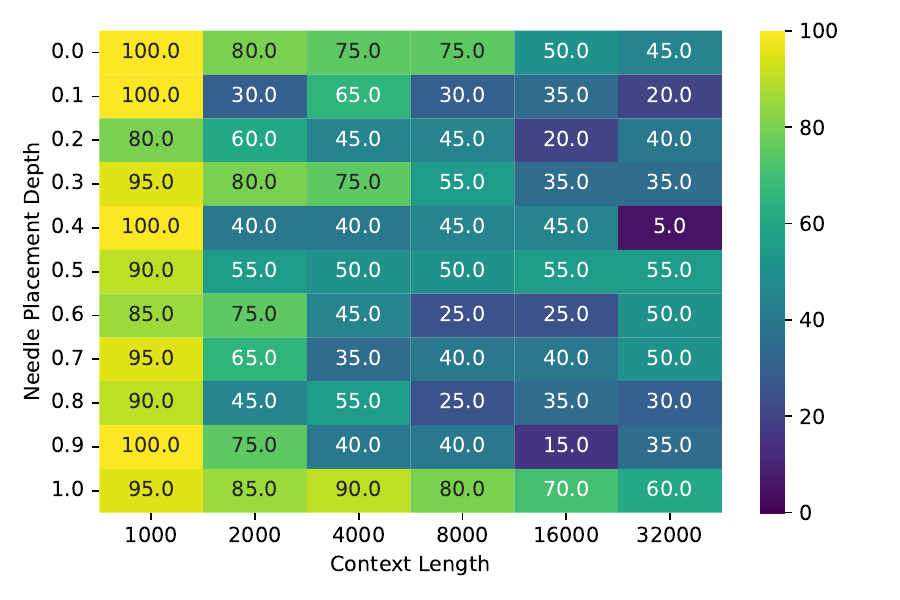}
    \caption{Qwen3-VL-8B-Instruct}
\end{subfigure}
\begin{subfigure}[b]{.32\linewidth}
    \includegraphics[width=\textwidth]{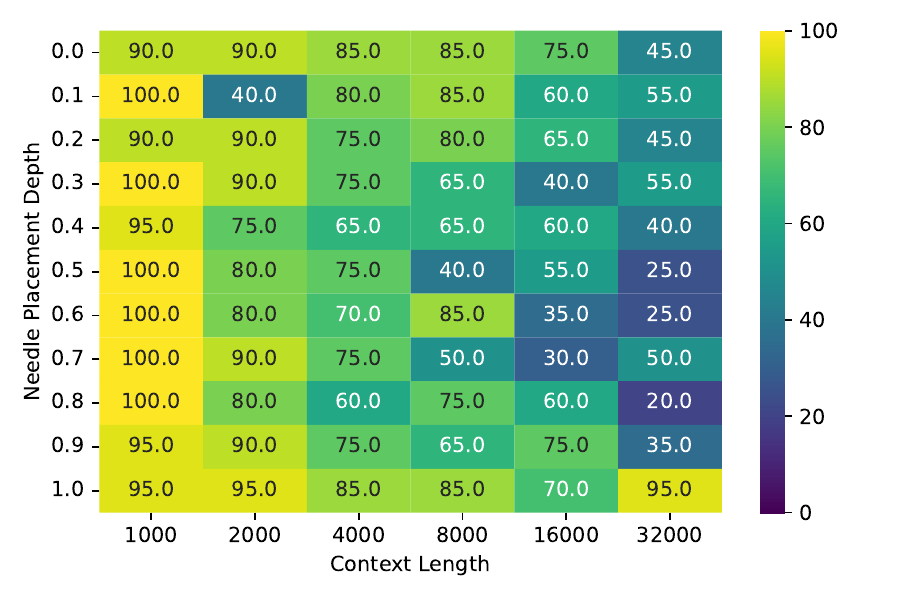}
    \caption{Qwen3-VL-235B-A22B-Instruct}
\end{subfigure}

\caption{\ours-Retrieval MV-NIAH (preset $\render=\renderdefault$) accuracy w.r.t. needle placement depth. Blank cells indicate \oom.}
\label{fig:appdx:ruler:mv}
\end{figure*}
\begin{figure*}[p]
\centering

\begin{subfigure}[b]{.32\linewidth}
    \includegraphics[width=\textwidth]{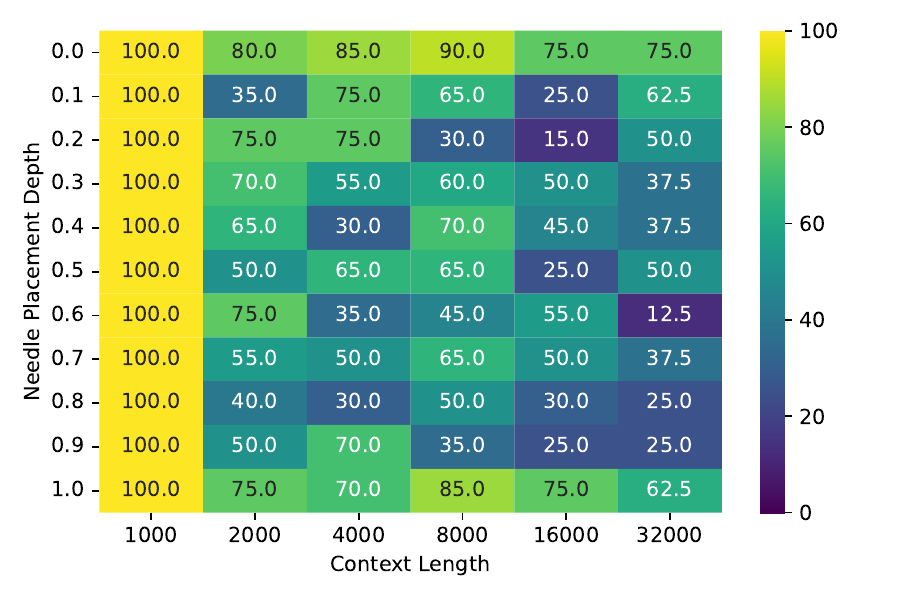}
    \caption{Gemini-2.5-Pro}
\end{subfigure}
\begin{subfigure}[b]{.32\linewidth}
    \includegraphics[width=\textwidth]{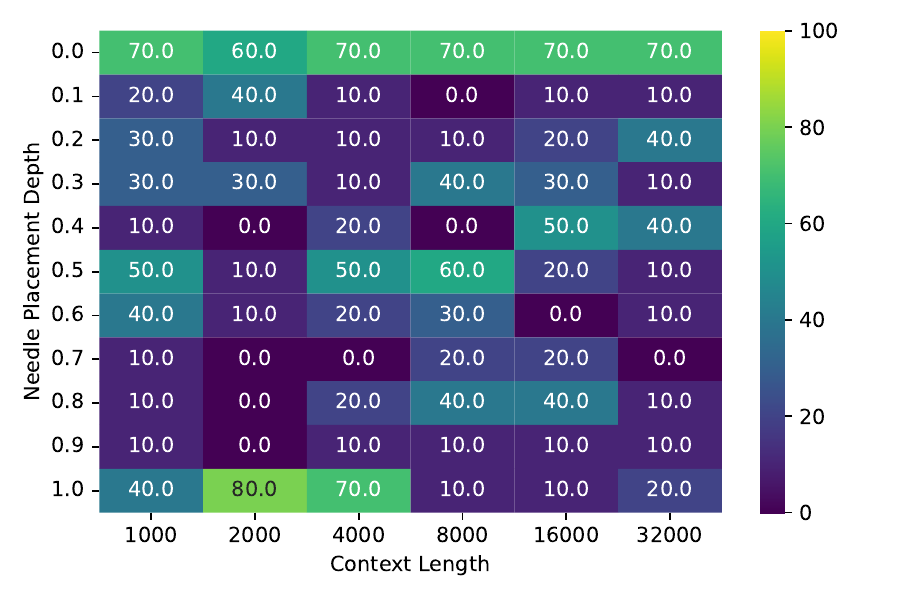}
    \caption{Gemma3-27B}
\end{subfigure}
\begin{subfigure}[b]{.32\linewidth}
    \includegraphics[width=\textwidth]{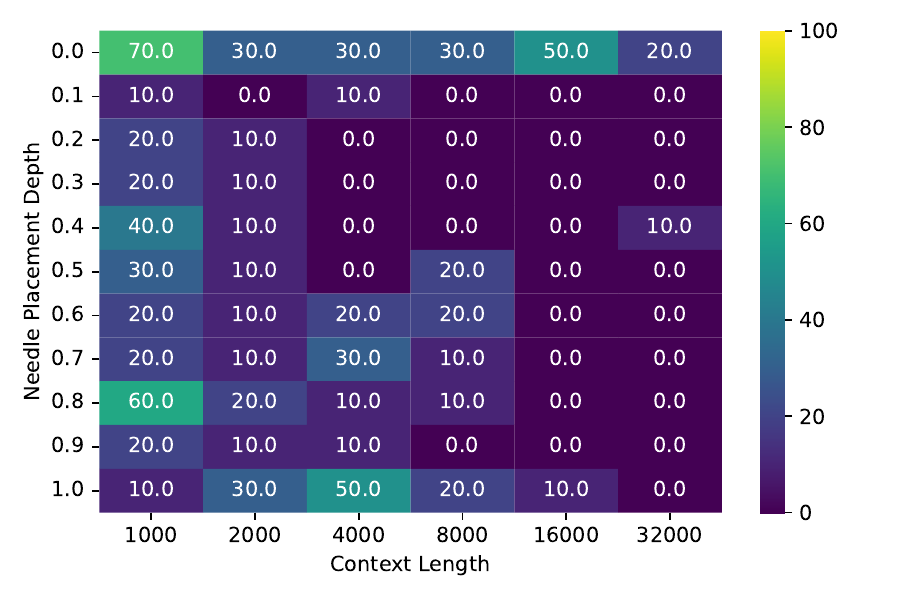}
    \caption{GLM-4.1V-9B-Thinking}
\end{subfigure}
\begin{subfigure}[b]{.32\linewidth}
    \includegraphics[width=\textwidth]{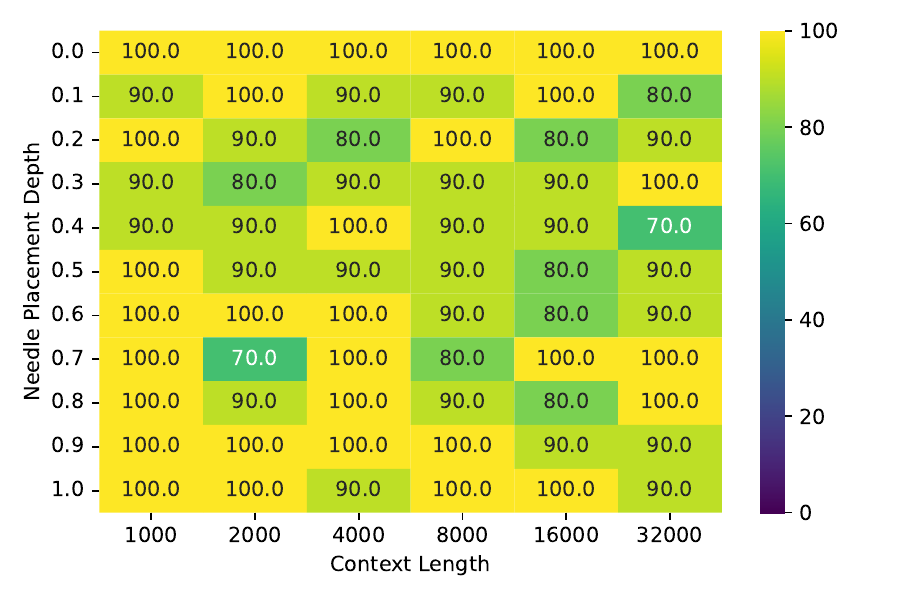}
    \caption{Glyph}
\end{subfigure}
\begin{subfigure}[b]{.32\linewidth}
    \includegraphics[width=\textwidth]{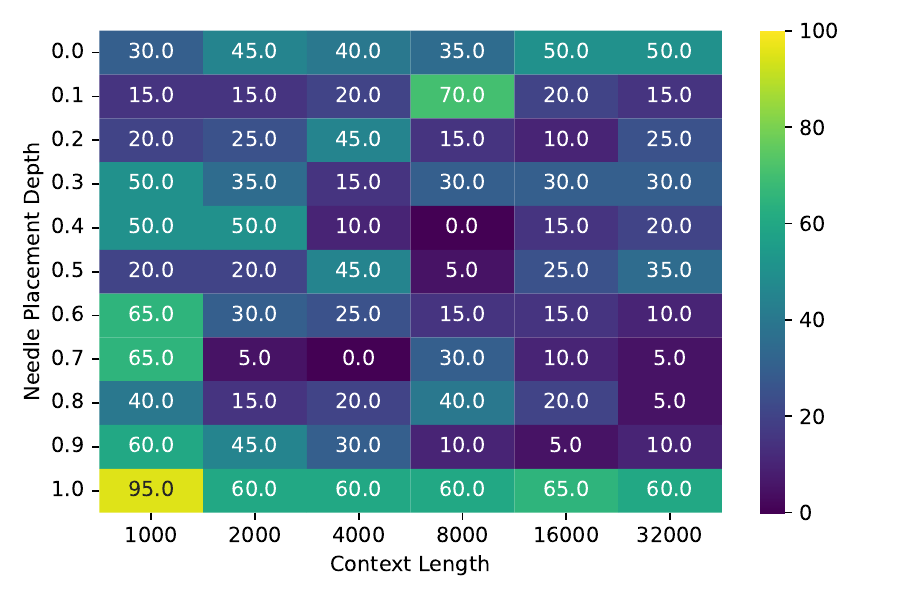}
    \caption{GPT-5}
\end{subfigure}
\begin{subfigure}[b]{.32\linewidth}
    \includegraphics[width=\textwidth]{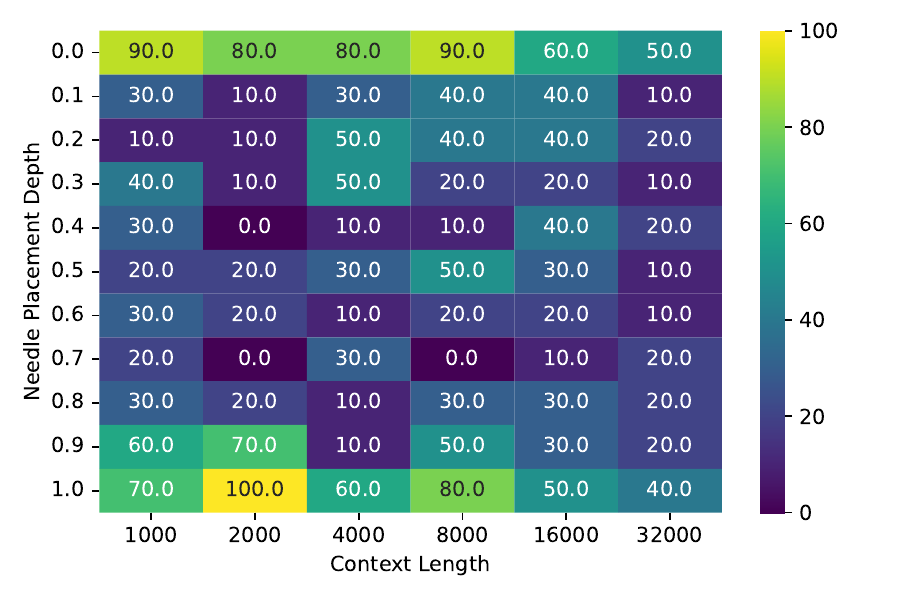}
    \caption{InternVL3.5-8B}
\end{subfigure}
\begin{subfigure}[b]{.32\linewidth}
    \includegraphics[width=\textwidth]{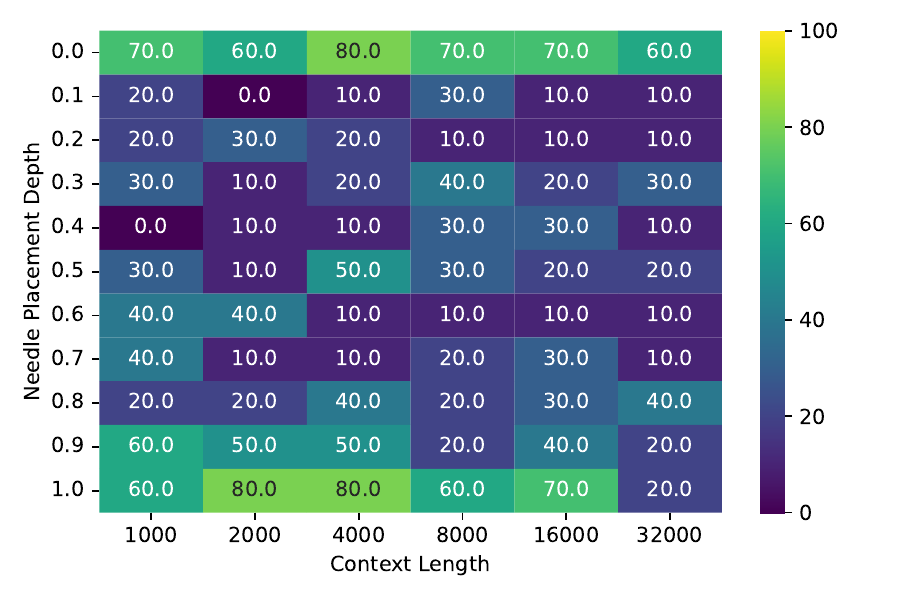}
    \caption{InternVL3.5-38B}
\end{subfigure}
\begin{subfigure}[b]{.32\linewidth}
    \includegraphics[width=\textwidth]{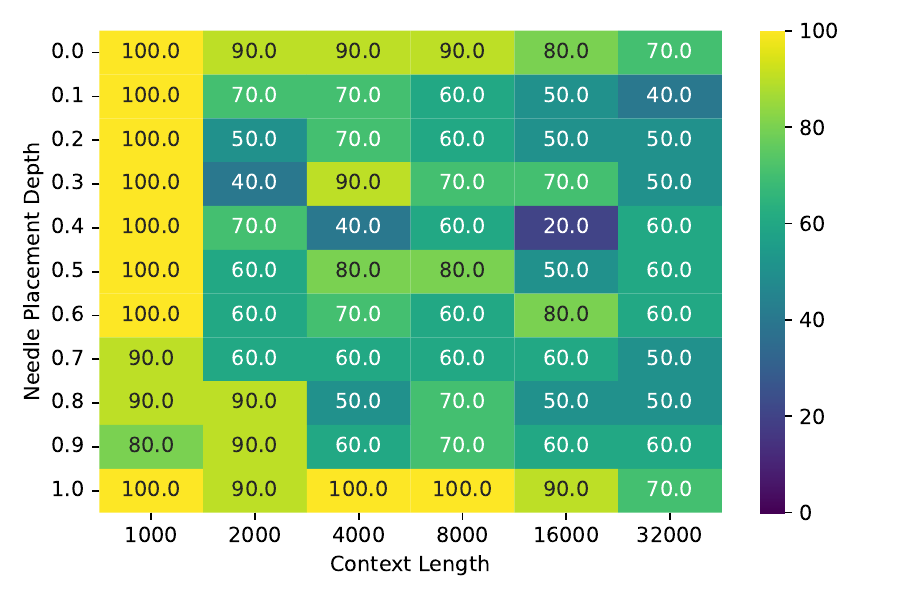}
    \caption{Kimi-VL-A3B-Instruct}
\end{subfigure}
\begin{subfigure}[b]{.32\linewidth}
    \includegraphics[width=\textwidth]{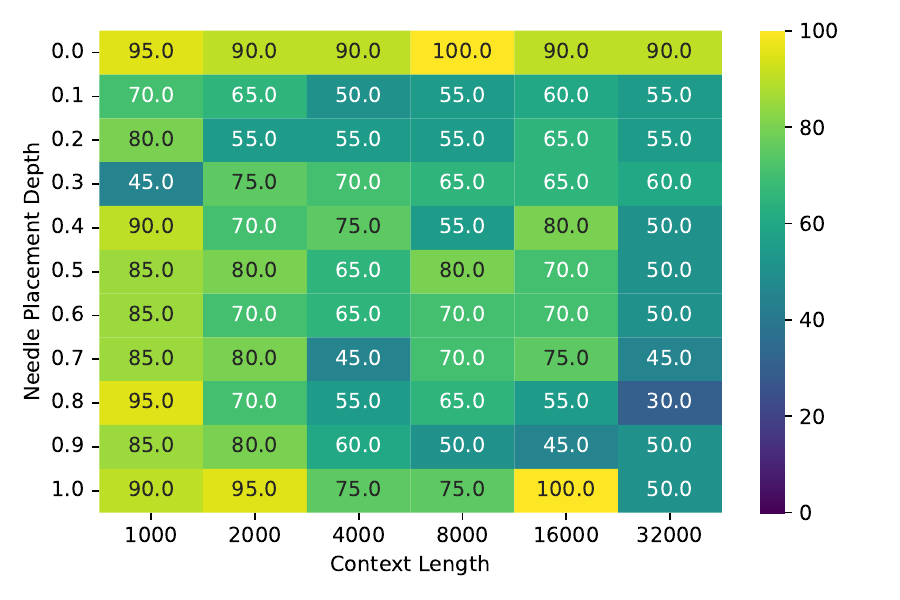}
    \caption{Qwen2.5-VL-7B-Instruct}
\end{subfigure}
\begin{subfigure}[b]{.32\linewidth}
    \includegraphics[width=\textwidth]{fig/heatmap/mq_niah/Qwen2.5-VL-72B-Instruct.pdf}
    \caption{Qwen2.5-VL-72B-Instruct}
\end{subfigure}
\begin{subfigure}[b]{.32\linewidth}
    \includegraphics[width=\textwidth]{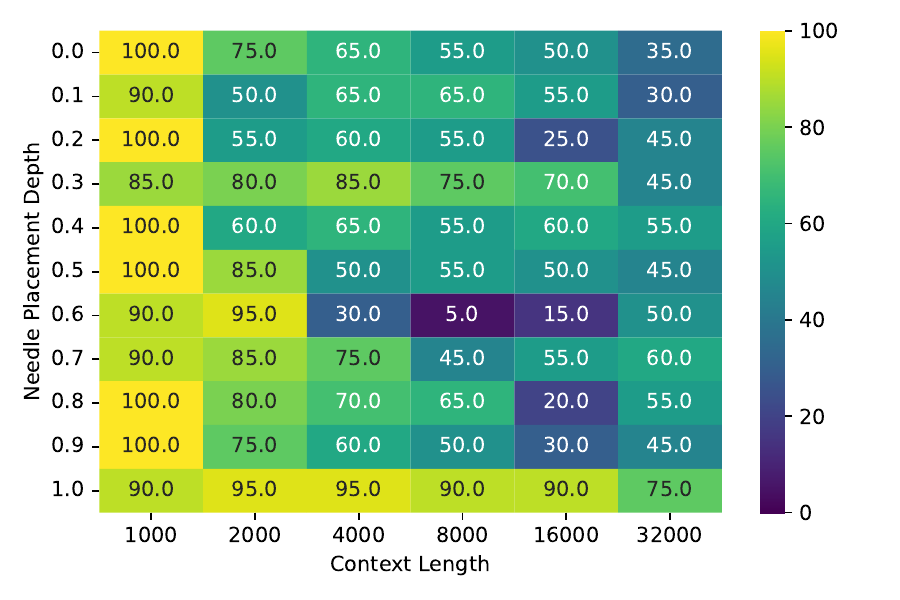}
    \caption{Qwen3-VL-8B-Instruct}
\end{subfigure}
\begin{subfigure}[b]{.32\linewidth}
    \includegraphics[width=\textwidth]{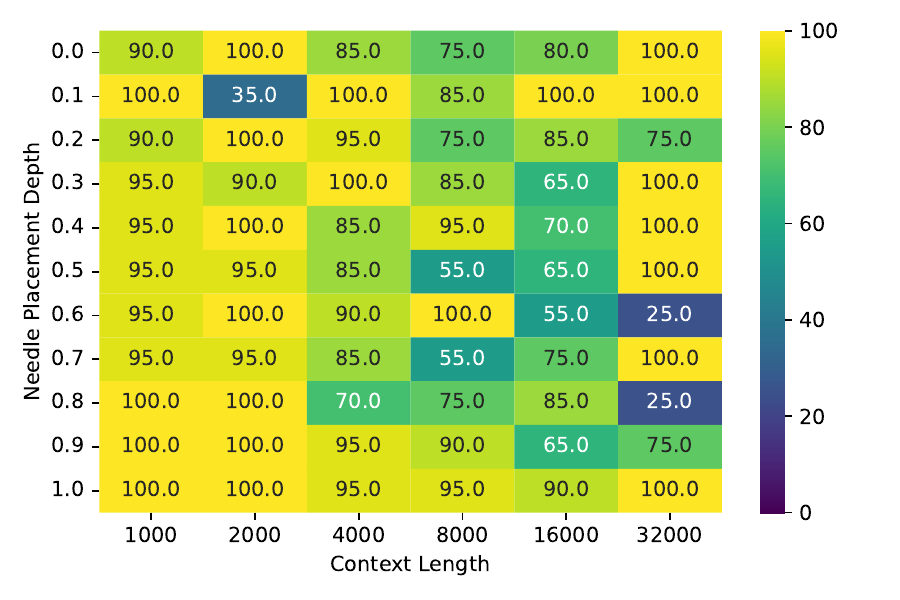}
    \caption{Qwen3-VL-235B-A22B-Instruct}
\end{subfigure}

\caption{\ours-Retrieval MQ-NIAH (preset $\render=\renderdefault$) accuracy w.r.t. needle placement depth. Blank cells indicate \oom.}
\label{fig:appdx:ruler:mq}
\end{figure*}

\begin{figure*}[p]
\centering

\begin{subfigure}[b]{.32\linewidth}
    \includegraphics[width=\textwidth]{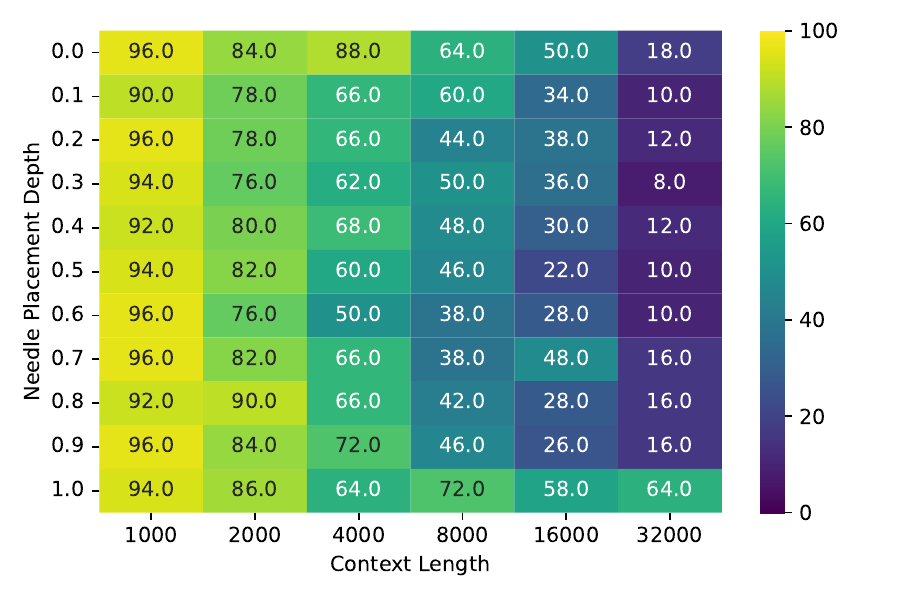}
    \caption{Qwen3-8B (LLM baseline)}
\end{subfigure}
\begin{subfigure}[b]{.32\linewidth}
    \includegraphics[width=\textwidth]{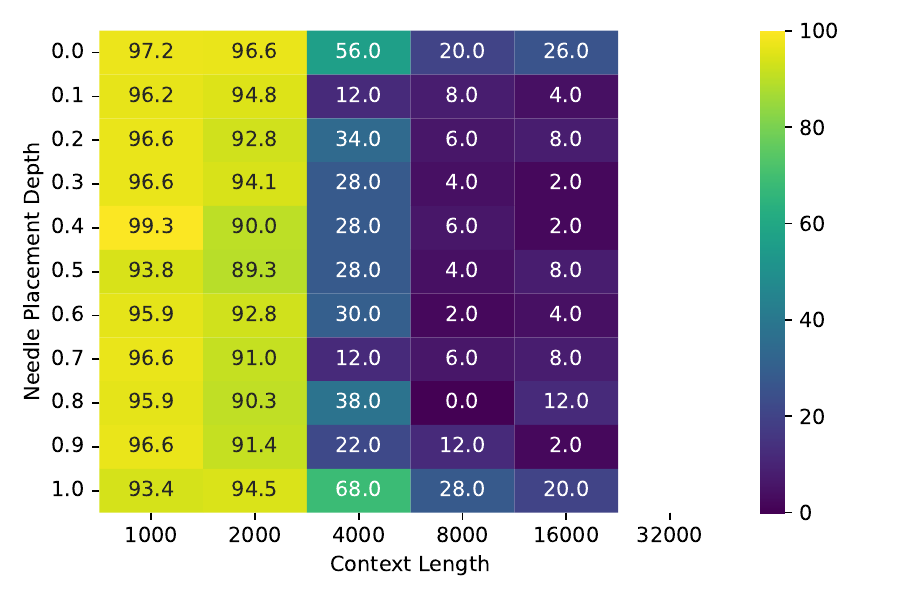}
    \caption{Gemini-2.5-Pro}
\end{subfigure}
\begin{subfigure}[b]{.32\linewidth}
    \includegraphics[width=\textwidth]{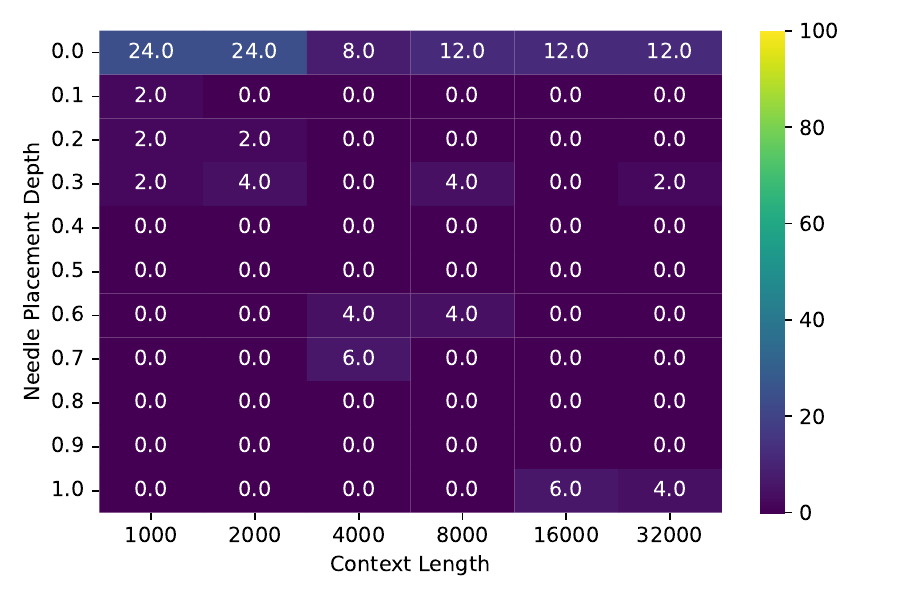}
    \caption{Gemma3-27B}
\end{subfigure}
\begin{subfigure}[b]{.32\linewidth}
    \includegraphics[width=\textwidth]{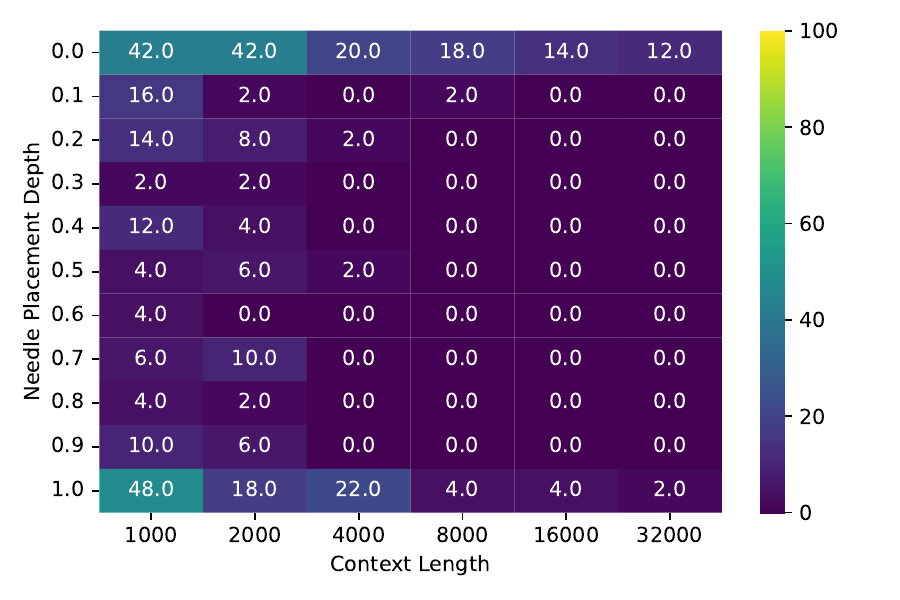}
    \caption{GLM-4.1V-9B-Thinking}
\end{subfigure}
\begin{subfigure}[b]{.32\linewidth}
    \includegraphics[width=\textwidth]{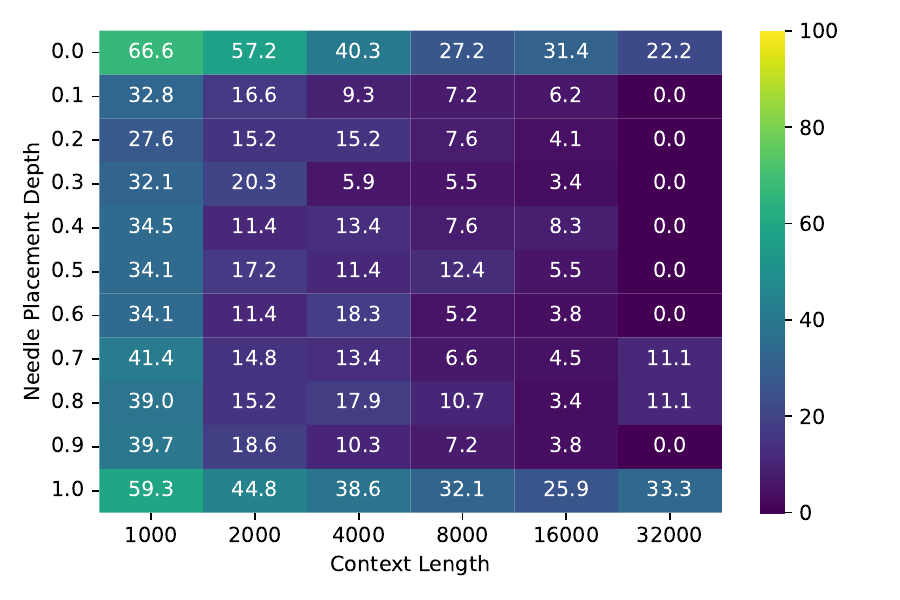}
    \caption{Glyph}
\end{subfigure}
\begin{subfigure}[b]{.32\linewidth}
    \includegraphics[width=\textwidth]{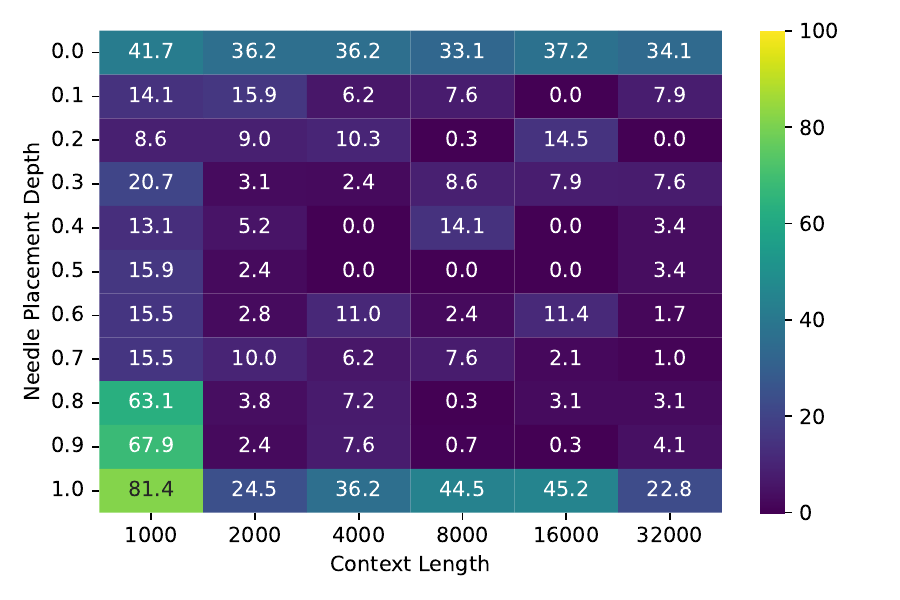}
    \caption{GPT-5}
\end{subfigure}
\begin{subfigure}[b]{.32\linewidth}
    \includegraphics[width=\textwidth]{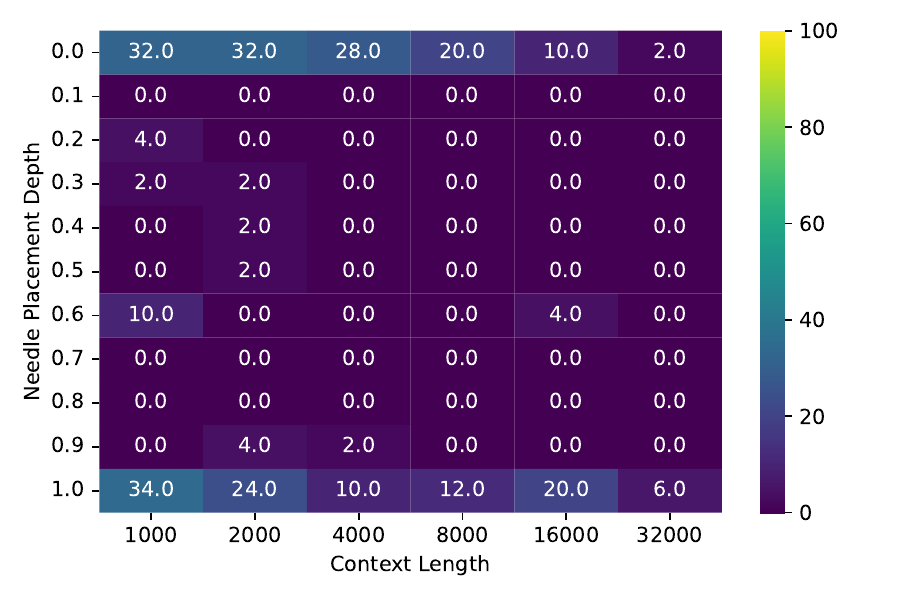}
    \caption{InternVL3.5-8B}
\end{subfigure}
\begin{subfigure}[b]{.32\linewidth}
    \includegraphics[width=\textwidth]{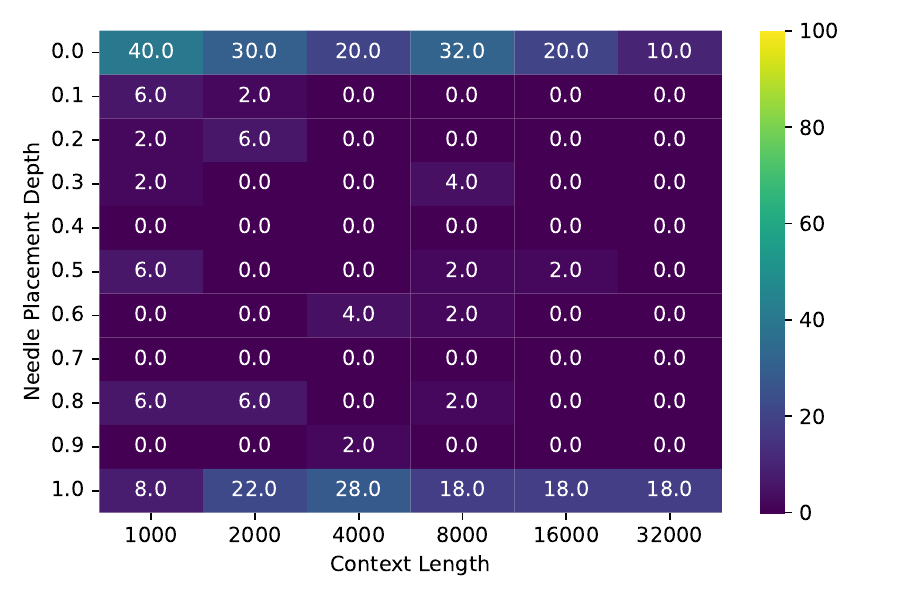}
    \caption{InternVL3.5-38B}
\end{subfigure}
\begin{subfigure}[b]{.32\linewidth}
    \includegraphics[width=\textwidth]{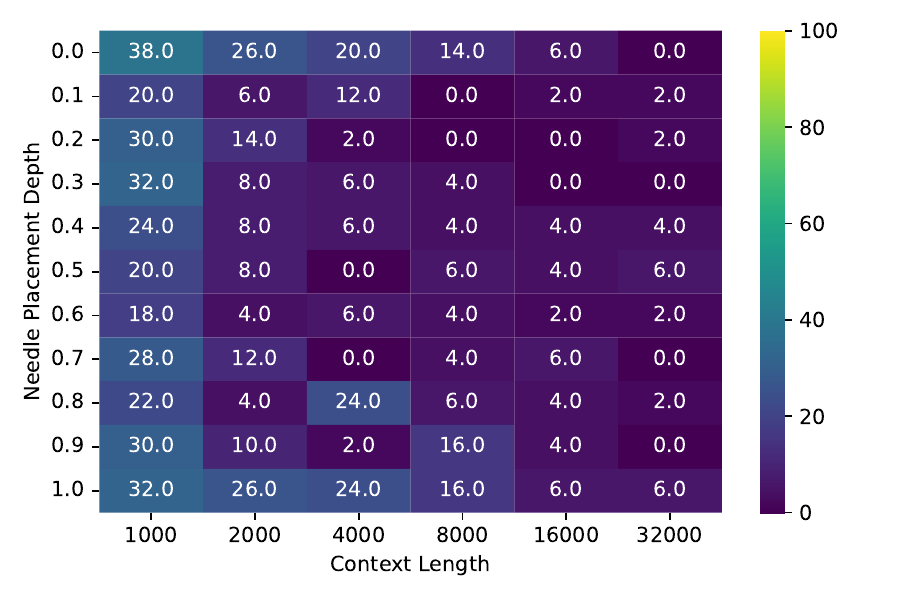}
    \caption{Kimi-VL-A3B-Instruct}
\end{subfigure}
\begin{subfigure}[b]{.32\linewidth}
    \includegraphics[width=\textwidth]{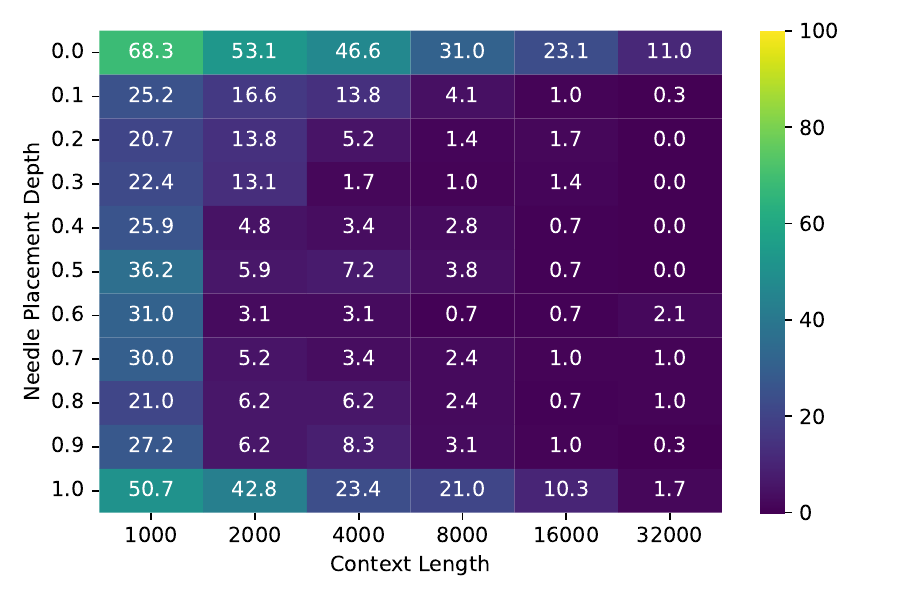}
    \caption{Qwen2.5-VL-7B-Instruct}
\end{subfigure}
\begin{subfigure}[b]{.32\linewidth}
    \includegraphics[width=\textwidth]{fig/heatmap/NoLiMa/Qwen2.5-VL-72B-Instruct.pdf}
    \caption{Qwen2.5-VL-72B-Instruct}
\end{subfigure}

\caption{\ours-Reasoning (preset $\render=\renderdefault$) accuracy w.r.t. needle placement depth. Blank cells indicate \oom.}
\label{fig:appdx:nolima}
\end{figure*}

\clearpage
\newpage
{
    \small
    \bibliographystyle{ieeenat_fullname}
    \bibliography{main}
}

\end{document}